\documentclass[]{informs3} % current default for manuscript submission
\OneAndAHalfSpacedXII % current default line spacing
\newenvironment{proof}{\emph{Proof: }}{\hfill$\square$}
\usepackage{natbib}
\usepackage[ruled]{algorithm}
\usepackage[noend]{algorithmic}
\usepackage{multirow}
\newcommand{\tabincell}[2]{\begin{tabular}{@{}#1@{}}#2\end{tabular}}

 \bibpunct[, ]{(}{)}{,}{a}{}{,}%
 \def\bibfont{\small}%
\allowdisplaybreaks %%% this make pages look tight.

\newcommand{\CUT}[1]{{}}

\renewcommand{\paragraph}[1]{\smallskip \noindent {\textbf{#1}}}

\input epsf

\newtheorem{theorem}{Theorem}

\newtheorem{corollary}[theorem]{Corollary}
\newtheorem{definition}{Definition}
\newtheorem{lemma}{Lemma}
\newtheorem{assumption}{Assumption}

\begin{document}

%\setcounter{page}{1}
%\DOI{10.1287/xxxx.0000.0000}%


% \RUNTITLE{Towards Optimal Ads Sequencing for Markovian Users}

 \TITLE{Cascade Submodular Maximization: Question Selection and Sequencing in Online Personality Quiz}

\ARTICLEAUTHORS{%
\AUTHOR{Shaojie Tang}
\AFF{Naveen Jindal School of Management, University of Texas at Dallas,  \\\EMAIL{800 W Campbell Road
Richardson, TX 75080, USA, shaojie.tang@utdallas.edu}}
\AUTHOR{Jing Yuan}
\AFF{Department of Computer Science, University of Texas at Dallas,
\\\EMAIL{ 800 W Campbell Road
Richardson, TX 75080, USA, csyuanjing@gmail.com}
}
} % end of the block

\ABSTRACT{%
%E-Commerce personalization aims to provide individualized offers, product recommendations, and other content to customers based on their interests.  The foundation of any personalization effort is customer segmentation. The idea of customer segmentation is to group customers together according to identifiable segmentation attributes including geolocation, gender, age, and interests.
Personality quiz is a powerful tool that enables costumer segmentation  by actively asking them questions, and marketers are using it as an effective method of generating leads and increasing e-commerce sales. In this paper, we study the problem of how to select and sequence a group of quiz questions so as to optimize the quality of customer segmentation. We assume that the customer will
sequentially scan the list of questions. After reading a question, the customer makes two, possibly correlated, random decisions: 1) she first decides whether to answer this question or not, and then 2) decides  whether to continue reading the next question or not.  We further assume that the utility of questions that have been answered can be captured by a monotone and submodular function. In general, our problem falls into the category of non-adaptive active learning based customer profiling. Note that under the our model, the probability of a question being answered depends on the location of that question, as well as  the set of other questions placed ahead of that question, this makes our problem fundamentally different from existing studies on submodular optimization. %We formulate our problem as a submodular selection and sequencing problem under cascade browse model. %In particular, we use conditional entropy to measure the utility of a given group of quiz questions. Our objective is to compute a sequence of quiz questions that lead to the maximum utility.
%We model the customer behavior when interacting with a sequence of quiz questions as a Markov process called cascade browse model.
We develop a series of question selection and sequencing strategies with provable performance bound.
Although we focus on the application of  quiz design in this paper, our results apply to a broad range of applications, including assortment optimization with position bias effect.}
%https://www.instagram.com/p/BJlFMxMBjSR/
%\KEYWORDS{deterministic inventory theory; infinite linear programming duality;
%  existence of optimal policies; semi-Markov decision process; cyclic schedule}

%}
%\HISTORY{Received: Feb 2019; Accepted: Dec 2020 by Subodha Kumar, after two rounds of revision}

\maketitle
%%%%%%%%%%%%%%%%%%%%%%%%%%%%%%%%%%%%%%%%%%%%%%%%%%%%%%%%%%%%%%%%%%%%%%

% Samples of sectioning (and labeling) in ISRE
% NOTE: (1) \section and \subsection do NOT end with a period
%       (2) \subsubsection and lower need end punctuation
%       (3) capitalization is as shown (title style).
%
%\section{Introduction.}\label{intro} %%1.
%\subsection{Duality and the Classical EOQ Problem.}\label{class-EOQ} %% 1.1.
%\subsection{Outline.}\label{outline1} %% 1.2.
%\subsubsection{Cyclic Schedules for the General Deterministic SMDP.}
%  \label{cyclic-schedules} %% 1.2.1
%\section{Problem Description.}\label{problemdescription} %% 2.

%%%%%%%%%%%%%%%%%%%%%%%%%%%%%%%%%%%%%%%%%%%%%
\section{Introduction}
\label{sec:intro}
E-Commerce personalization has been recognized as one of the most effective methods in increasing sales \citep{yang2005evaluation,ansari2003customization}. For example, Amazon and other retail giants provide personalized product recommendations based on their customers' interest. Gartner predicts that by 2020,  those who successfully handle personalization in E-Commerce will increase their profits by up to 15\%.  The starting point of any personalization effort is to obtain a clearer picture of individual customers,  this is often done by customer segmentation, e.g., partition a customer base into groups of individuals with similar characteristics that are relevant to marketing, such as a geographic location, interests, time of visit, etc.  Since customer segmentation relies on both the quality and quantity of data collected from customers, it is critical to decide what data will be collected and how it will be collected. For returning customers, we can use their past behavior such as browsing history to perform customer segmentation and further personalize her current experience. But how to decide the segment a new customer belongs to without knowing her browsing history? One popular approach to gather data from first-time customers is  personality quiz. The purpose of personality quiz is to segment every potential and current customer by actively asking them questions. After answering a few questions, customers are matched with the type of recommendations or product that best suit their responses. Marketers are starting to use it as an effective method of generating leads and increasing e-commerce sales. For example, several websites  such as Warby Parker \footnote{\url{https://www.warbyparker.com/}} and ipsy\footnote{\url{https://www.ipsy.com/}}  are using personality quizzes to determine customers' interest profiles and make recommendations.
As compared with other preference elicitation methods, personality quiz-based system requires significantly less effort from customers and they expressed stronger intention to reuse this system and introduce it to others \citep{hu2009acceptance,hu2009comparative}.  Although the benefit of personality quiz has been well recognized, it is not clear, in general, how to optimize the quiz design so as to maximize this benefit. In this paper, we formulate the quiz design problem as a combinatorial optimization problem. We aim at selecting and further sequencing a group of questions from a given pool  so as to maximize the quality of customer segmentation. While the design of each individual question such as  formatting, coloring, the way the question is asked can be complex and important \citep{couper2001web}, that topic is out of scope of this paper. We exclusively focus on  the question selection and sequencing problem by assuming that all candidate questions are pre-given\footnote{In one of our extensions, we are allowed to optimize the question design by treating the ``Prefer Not to Answer'' (PNA) option as a decision variable for each question.}.

The input of our problem is a set of attributes and a pool of candidate questions. We say a question covers an attribute if the answer to that question reveals the value of that attribute. For example, question $\mbox{``Where are you living?"}$ covers  attribute ``location''.  Intuitively, a group of ``good'' questions should cover as many important attributes as possible so as to minimize the uncertainty about the customer.  Given answers to a group of questions, we measure its uncertainty using the conditional entropy of the uncovered attributes of the customer. Our ultimate goal is to select and sequence a group of questions so as to minimize the uncertainty  subject to a set of practical constraints. {We notice that there are two types of quiz design, namely, paging design and scrolling design. The one question per page quiz is called paging design while the multiple questions per page quiz is called scrolling design. In this paper, we focus on paging design where only one question is displayed to the customer per page. It requires the customer to click the ``Next'' button once they finish a question, thus the customer must follow the sequence specified by the quiz designer to answer (or PNA) the questions.}

In general, our problem falls into the category of non-adaptive active learning based customer profiling. The idea of most existing studies is to actively select a group of items, e.g., movies or cars, and asking for customers' feedback on them. This feedback, in turn, can help to enhance the performance of recommendation in the future. However, they often assume that the customers are willing to provide feedback on all selected items, irrespective of which items are selected and in what sequence. As a consequence, their problem is reduced to a subset selection problem. We argue that this assumption may not hold in our problem, e.g., it has been shown that not all customers are willing to share their personal information with a site. According to the survey conducted by \cite{culnan2001culnan}, two of three customers abandon sites that asks for personal information and one of five customers has provided false information to a site. This motivates us to consider a realistic but significantly more complicated customer behavior model. Our model captures the externality of a question by allowing the customer to  ``opt-out'' of answering a question or even quit the quiz prematurely after answering some questions. A more detailed comparison between our work and related work is presented in Section \ref{sec:related}. We next give a brief overview to some important constraints considered in this paper.

\paragraph{Cardinality Constraint} We can select up to $b$ questions to include in the quiz where $b$ is some positive integer. For example, it has been shown that 6-8 questions per quiz could be an appropriate setting since it maximizes completions and leads generated (https://socialmediaexplorer.com/content-sections/tools-and-tips/how-to-make-a-personalized-quiz-to-drive-sales/).

\paragraph{User Behavior} Our setting considers that the user behavior during a personality quiz can be described as a Markov process (a detailed description of this model is presented in Section \ref{sec:behavior}). The customer interacts with a sequence of questions in order, after reading a question, she decides probabilistically whether or not to answer it with some question specific probability, called \emph{answer-through-rate}. In principle, this probability could depend on many factors including the cognitive efforts required for understanding and answering the question, and the sensitivity of the question, etc. Our model also allows the customer to select ``Prefer Not to Answer'' (PNA) option, if any, to skip a particular question. A more detailed discussion on PNA option is provided in the next subsection.  In addition, each question has a \emph{continuation probability}, representing the likelihood that the customer is willing to continue the quiz after interacting with the current one. This continuation probability captures the \emph{externality} of a question, e.g., a very sensitive or lengthy question could cause the customer to exit the quiz prematurely. The existence of such externality makes our problem even more complicated, e.g., the ordering of selected questions matters. For example, Typeform\footnote{\url{https://www.typeform.com/surveys/question-types/}}, an online software as a service (SaaS) company that specializes in online form building and online surveys, suggests that it is better to put sensitive and demographic questions at the end of a quiz or survey.%: ``\emph{Starting a survey with intimidating or demographic questions like age and income can put people off. Your first survey question should be interesting, light, and easy to answer. Once they've started, they're more likely to finish and answer more sensitive questions.}''

\paragraph{PNA option} Regarding the role played by PNA option in a quiz, there exist two contradicting arguments. On the one hand, several studies \citep{schuman1996questions,hawkins1981uninformed} empirically demonstrate that the data and subsequent analyses will better off by including a PNA option due to it decreases the proportion of uninformed responses. On the other hand, opponents believe that providing a PNA option could negatively impact the quality of the answer because some customers tend not to answer the question so as to minimize the effort required to complete the quiz \citep{poe1988don,sanchez1992probing}. Since both arguments are empirically validated by previous studies, we decide to cover both cases in this work.

{

\subsection{Summary of Contributions}

We next summarize the contributions made in this paper. We first show (in Section \ref{sec:2.1}) that our problem  cannot be approximated in polynomial time within a ratio of $(1-1/e+\epsilon)$, unless $P=NP$. Then we develop a series of effective solutions with provable performance bound. For the case where PNA is not an option (in Section \ref{sec:3}), our algorithm achieves an approximation ratio that is arbitrarily close to $\frac{1-1/e-\epsilon}{4}$ where $e$ is a constant whose value is arbitrarily close to 2.718. For the case where PNA is available (in Section \ref{sec:4}), we achieve the same approximation ratio. We validate the performance of our solution through a comparison to the benchmark solutions and to the optimal solution. Our solution outperforms the benchmarks under all test settings. We show that the actual performance of our solution achieves the expected utility exceeding $87\%$ of the optimum.

We subsequently consider three extensions of the basic model by 1. taking into account the slot-dependent decay factor,  2. incorporating the PNA option as a decision variable, and 3. extending this study to scrolling design (in Section \ref{sec:5}). In the first extension, we assume that the answer-through-rate of a question does not only depend on its intrinsic quality, but also depends on its position. For the question selection and sequencing problem under this model, our algorithm achieves an approximation ratio that is arbitrarily close to  $\frac{1-1/e}{4}$ (resp. $\frac{0.38}{4}$) when PNA is not an option (resp. is an option). In the second extension, we treat the PNA option as a decision variable for each question. In addition to identifying a sequence of questions, we must also decide whether or not to offer the PNA option for each question. We propose a nearly $\frac{0.38}{4}$ approximate solution to this extended problem. In the last extension, we extend our results to the  scrolling design. From a technical point of view, we develop several  approximation algorithms of independent interest, and make fundamental contributions to the field of  submodular maximization and submodular sequencing. We also unravel hidden relationship between our study and the assortment optimization problems.

One nice property of our solutions is that they are all non-adaptive, making them easy to implement. That is, once a sequence of questions are returned from our solution, it does not change with time or with the answers being collected from the customer. As compared with adaptive solutions which often require to update the list of questions dynamically,  our solution, which is obtained by solving a deterministic problem, is easy to compute and implement.}

Most of notations are listed in Table \ref{table:1}. All missing proofs are provided in the online appendix.
\begingroup
\renewcommand*{\arraystretch}{0.5}
\begin{table}[h!]
\centering
\caption{Main Notation}
\label{T1}
\begin{tabular}{|l|l|}
	\hline
	Notation & Description                     \\ \hline
	$\mathcal{Q}$ & A set of questions.            \\ \hline
	$Q$ & A sorted sequence of $Q$.					  \\ \hline
    $g(\mathcal{Q})$ & The utility of answers to $\mathcal{Q}$ \\ \hline
    $f(Q)$& The expected utility of displaying $Q$ to the customer. \\ \hline
	$p^+_q$ (resp. $p^-_q$) & Probability of answering (resp. PNA) $q$ after reading it.	\\ \hline
	$c_q^+$ (resp.  $c_q^-$)  & Probability of continuing to read the next question after \\
                              &answering (resp. PNA)  $q$. \\ \hline
    $c_q$ &  Aggregated continuation probability of $q$: $p^+_{q}c_{q}^+ +  p^-_{q} c_{q}^-$ \\ \hline
    	$C_{Q[i]}$ &  Reachability of the $i$-th question of $Q$.\\ \hline
	$Q \oplus Q'$ & Concatenation  of two sequences $Q$ and $Q'$.\\ \hline
	$Q_{\leq i}$ (resp. $Q_{< i}$, $Q_{> i}$, $Q_{\geq i}$) &  The subsequence of $Q$ which is scheduled no later than \\
&(resp. before, after, no earlier than) slot $i$. \\ \hline
$\mathcal{J}(Q)$ & A random set of answered questions given $Q$.\\ \hline
 $\mathcal{R}(Q)$ or $\mathcal{R}(\mathcal{Q})$ & A random set obtained by including each question $q\in Q$ \\
 &independently  with probability $p^+_q$.\\ \hline
\end{tabular}
\label{table:1}
\end{table}
\endgroup
\section{Literature Review}
\label{sec:related}
Our paper falls into the general category of non-adaptive active learning supported personalization. This section reviews the literature on four topics that are closely related to our research.

\subsection{Active Learning based Recommender System} Active learning, as a subfield of machine learning \citep{bishop2006pattern}, has been widely used in the design of effective recommender systems. For the purpose of acquiring training data, active learning based recommender system often  actively  solicits customer feedback on a group of carefully selected items \citep{kohrs2001improving,rubens2007influence,golbandi2011adaptive,chang2015using}. Existing systems can be further classified into two categories: \emph{adaptive} learning and \emph{non-adaptive} learning. Non-adaptive learning refers to those learning strategies who require all customers to rate the same set of items while adaptive learning \citep{boutilier2002active,golbandi2011adaptive,rubens2007influence} could propose different items to different customers to rate. Since our paper belongs to the category of non-adaptive learning, we next give a detailed review to the state-of-art of non-adaptive learning based recommender system. Depending on the item selection rule, there are three types of strategies: uncertainty-reduction, error reduction and attention based.  The goal of uncertainty-reduction based systems is to reduce the uncertainty about the customers' opinion about new items, they achieve this by selecting items with highest variance \citep{kohrs2001improving,teixeira2002activecp} or highest entropy \citep{rashid2002getting} or highest entropy0 \citep{rashid2008learning}. The goal of error reduction  based systems is to minimize the system error \citep{golbandi2010bootstrapping,liu2011wisdom,cremonesi2010performance}. The idea of attention-based strategy is to select the items that are most popular  among the customers \citep{golbandi2010bootstrapping,cremonesi2010performance}. Our problem is largely different from all aforementioned studies in terms of both application context  and problem formulation: (1) Instead of investigating a particular recommender system, we study a general costumer segmentation problem whose solution  serves as the foundation of any personalized service; (2) We are dealing with a significantly more complicated customer behavior model where the customer is allowed to pick PNA option or terminate the quiz prematurely. All aforementioned studies assume that the customers are guaranteed to rate all selected items, regardless of the sequence of those items, thus their problem is reduced to a subset selection problem; (3) Most of existing studies are developing heuristics without provable performance bound, we develop the first series of algorithms that achieve bounded approximation ratios.

\subsection{Learning Offer Set and Consideration Set} The other two related topics are ``offer set'' \citep{atahan2011accelerated}  and ``consideration sets'' \citep{roberts1991development}.  Our problem differs from both ``offer set'' and ``consideration sets'' in fundamental ways. The focus of offer set is to investigate how the profile learning process can be accelerated by carefully selecting the links to display to the customer. In \citep{atahan2011accelerated}, customers implicitly compare alternative links and reveal their preferences based on the set of links offered, this is different from our model where customers are explicitly asked to answer questions. The literature on consideration sets aims at determining the subset of brands/products that a customer may evaluate when making purchase decision. Their model did not capture the externality of a question e.g., the customers are forced to answer all questions, thus the sequence of questions did not play a role in their non-adaptive solution.  In addition, most of aforementioned studies did not provide any theoretical bounds on their proposed solutions. We consider a joint question selection and sequencing problem which is proved to be NP-hard (in Section \ref{sec:2.1}), and theoretically bound the gap between our solution and the optimal solution.

{\subsection{Assortment Optimization and Product Ranking} Our work is also closely related to the assortment optimization problem \citep{doi:10.1111/poms.13111,davis2014assortment,blanchet2016markov,farias2013nonparametric,rusmevichientong2014assortment,doi:10.1111/j.1937-5956.2007.tb00265.x,feldman2015bounding,doi:10.1111/poms.12685}. Assortment optimization is a core operations management problem which arises in many domains such as retailing \citep{doi:10.1287/msom.1050.0088,mahajan2001stocking,tayur2012quantitative}, revenue management for airlines and hotels \citep{gallego2015general,li2015d}, as well as online advertising \citep{saure2013optimal,rusmevichientong2010dynamic}. In assortment optimization, a firm offers a set of products to customers and its objective is to find the best assortment of products that maximizes the expected revenue. Although the problem setting of our problem is different from assortment optimization, these two problems share some similarities. For example, in both problems, the firm aimed at selecting a group of items (e.g., a sequence of quiz questions in our problem v.s. a group of products in assortment optimization) to display to a customer in order to  maximize some utility function (e.g., maximizing  the quality of customer segmentation in our problem v.s. maximizing  the expected revenue in assortment optimization).  Moreover, both problems require some customer's behavior model as an input to capture the customer's behavior upon viewing a list of items displayed to her. We next briefly discuss the relationship of our work and the contributions of our work to the assortment optimization literature.

Despite a rich body of literature on assortment optimization, most of them assume that the position of a product does not affect the expected utility gained from that product. In other words, most of existing studies assortment optimization did not incorporate  the position bias effect in their customer's behavior model. However, in our customer's behavior model, the position bias effect plays an important role in evaluating the quality of a given solution. \cite{davis2013assortment} and  \cite{abeliuk2016assortment} were the first to investigate the assortment optimization problem  with position bias.  They assume that the the chance of an item being purchased is jointly decided by its own location and the locations of the rest of the displayed items.  Their model is different from ours, e.g., we adopt a sequential browse model where the chance of an item (e.g., quiz question) being viewed is only affected by those items placed in earlier positions. %preference weight, which is a term used in their customer choice model to measure the , of an item (e.g., product)  is decided by both its value and its location. However, they did not explicitly dec. Under their model, given a sequence of items (e.g., products) that are selected to display to a customer, the chance of an item being bought is decided by its value and its own location (without depending on the locations of the other displayed items).  However, under our setting, the chance of answering an item (e.g., quiz question) is not only dependent on its own location but is also affected by the locations of the other items (e.g., those questions which are placed ahead of it).
Recently, \cite{aouad2015display,ferreira2019learning,asadpour2020ranking} consider  a setting similar to ours. They assume that there are a limited number of  vertically differentiated display positions on a webpage, and the firm's objective is to assign a group of products to those positions to maximize the expected revenue. They adopt a consider-then-choose model to capture the customer's purchase behavior, e.g., they assume that the customer first browses a random number of products, and then makes a choice within browsed products according to some customer choice model such as multinomial logit models \citep{rusmevichientong2014assortment}. One limitation of their browse model is that it can not capture the externality among displayed items, e.g., they assume that the chance of an item being browsed is independent of the other displayed items. The closest study to ours is that of \citep{tang2020product,tang2020assortment} who consider the assortment optimization problem under the cascade browse model. Similar to our setting, they assume that each item is associated with a continuation probability which measures the probability that the customer continues  browsing  the next item after browsing this item. However, their utility functions are neither monotone nor submodular. For a given sequence of displayed items, \cite{najafi2019multi}  aimed at finding the best price for each displayed item to maximize the expected revenue under the cascade browse model. % However, their focus is not on finding the best sequence of items, in fact, they assume that the sequence of displayed items is already given.
Although we put our focus on the application of quiz design in this paper, our result is general enough to apply to a broad range of optimization problems, including assortment optimization problems, that involve cascade model and submodular utility functions. In particular, we can apply our results to get a constant-factor approximation  for any assortment optimization problem that satisfies the following two conditions: 1) The consideration set of a customer is formed based on the  cascade  model, and 2) the revenue function is monotone and submodular in terms of the set of products included in the consideration set.  Notably, \cite{han2019assortment} and \cite{asadpour2020ranking} relate the assortment optimization and submodularity, and \cite{han2019assortment} characterizes the conditions under which the expected revenue obtained from a single customer class under some classic customer choice model is submodular. One important observation is that  if all products have the same revenue under the multinomial logit models, then the expected revenue function  is shown to be monotone and submodular.}

\subsection{Submodular Optimization} We later show that our problem is a submodular maximization problem. Although submodular maximization has been extensively studied in the literature  \citep{nemhauser1978best,nemhauser1978analysis,kawahara2009submodularity,calinescu2011maximizing}, most of them focus on subset selection problem where the ordering of selected elements does not affect its utility. Our work differs from theirs in that we consider a joint selection and sequencing problem. Recently, \cite{tschiatschek2017selecting,alaei2010maximizing,zhang2015string} consider the submodular sequencing problem, however their model and problem formulation are largely different from ours. \cite{tschiatschek2017selecting} use a directed graph to model the ordered preferences and their objective is to find a sequence of nodes that covers as many edges in the directed graph as possible. Their objective function is not always submodular, and their formulation does not involve any subset selection, because, by default, they can select all elements. \cite{alaei2010maximizing} develop approximation algorithms for maximizing  a utility function
that is sequence submodular and sequence non-decreasing. However, our utility function is not sequence non-decreasing. \cite{zhang2015string} develops a greedy algorithm whose performance bound is depending on the total backward curvature of the utility function with respect to the optimal solution. The worst-case performance of their solution is arbitrarily bad under our setting due to the unbounded total backward curvature of our utility function with respect to the optimal solution. Very recently, \cite{tang2021aamas} studied the cascade submodular maximization problem under the adaptive setting. They aimed at selecting a group of items sequentially based on the feedback from previously selected items. However, they assume that the continuation probability of an item is independent of its state (i.e., in the context of quiz design, the state of a question is determined by whether it  receives an answer or not). Our study generalizes their setting by considering a state-dependent continuation probability. As mentioned earlier, although we restrict our attention to the question selection and sequencing  problem in this paper, our research contributes fundamentally to the field of submodular subset selection and sequencing maximization.

\section{Preliminaries and Problem Formulation}
\subsection{Preliminaries}
\label{sec:behavior}

\subsubsection{Utility of Answered Questions}
 \label{sec:utility}
 Consider any group of answered questions $\mathcal{S}\subseteq  \Omega$, we use  $g(\mathcal{S})$ to represent the utility of $\mathcal{S}$. Intuitively, obtaining answers to a group of ``good'' questions should reduce the uncertainty and provide better insights on the customer.
\begin{assumption}
In this work, we assume that  $g$ is non-decreasing and submodular, and $g(\emptyset)=0$. That is,  for every $\mathcal{Y}_1, \mathcal{Y}_2 \subseteq \Omega$ with $\mathcal{Y}_1 \subseteq \mathcal{Y}_2$ and every $y \in \Omega \backslash \mathcal{Y}_2$, we have that $g(\mathcal{Y}_1\cup \{v\})-g(\mathcal{Y}_1)\geq g(\mathcal{Y}_2\cup \{y\})-g(\mathcal{Y}_2)$ (Submodularity). Moreover,  $g(\mathcal{Y}_1) \leq g(\mathcal{Y}_2)$  (Monotonicity).
\end{assumption}

We next give a concrete example to show that an entropy-like utility function $g(\mathcal{S})$ is indeed non-decreasing and submodular.

\underline{An Example of Entropy-like Utility Function.} Assume there are $m$ attributes $\Phi$ and $n$ questions $\Omega$. We say question $q\in \Omega$ covers attribute  $a \in \Phi$ if the answer to $q$ reveals the value of $a$. %For example, question $\mbox{``What is your gender?"}$ covers  attribute ``gender''.
We say a group of questions $\mathcal{S}\subseteq \Omega$ covers $a$ if $a$ can be covered by at least question from $\mathcal{S}$. We use  $\mathcal{A}(\mathcal{S})$ to denote the set of all attributes that can be covered by $\mathcal{S}$. One common notation of uncertainty is the conditional entropy of the unobserved attributes of a customer after answering $\mathcal{S}$.
\begin{equation}
\label{eq:0000}
H(X_{\mathcal{A}\setminus \mathcal{A}(\mathcal{S})}|X_{\mathcal{A}(\mathcal{S})})=-\sum_{\substack{\mathbf{x}_{\mathcal{A}\setminus \mathcal{A}(\mathcal{S})}\in \mathrm{dom } \,X_{\mathcal{A}\setminus \mathcal{A}(\mathcal{S})}\\ \mathbf{x}_{\mathcal{A}(\mathcal{S})}\in \mathrm{dom} \,X_{\mathcal{A}(\mathcal{S})}}} P(\mathbf{x}_{\mathcal{A}\setminus \mathcal{A}(\mathcal{S})}, \mathbf{x}_{\mathcal{A}(\mathcal{S})})\log P(\mathbf{x}_{\mathcal{A}\setminus \mathcal{A}(\mathcal{S})}| \mathbf{x}_{\mathcal{A}(\mathcal{S})})
\end{equation}
where we use $X_{\mathcal{A}\setminus \mathcal{A}(\mathcal{S})}$ and $X_{\mathcal{A}(\mathcal{S})}$ to denote sets of random variables associated with attributes in $\mathcal{A}\setminus \mathcal{A}(\mathcal{S})$ and $\mathcal{A}(\mathcal{S})$. Intuitively, a group of ``good'' questions $\mathcal{S}$ would  minimize Eq. (\ref{eq:0000}). Based on the chain-rule of entropies, we have $H(X_{\mathcal{A}\setminus \mathcal{A}(\mathcal{S})}|X_{\mathcal{A}(\mathcal{S})})=H(X_{\mathcal{A}})-H(X_{\mathcal{A}(\mathcal{S})})$. Due to $H(X_{\mathcal{A}})$ is fixed,  minimizing Eq. (\ref{eq:0000}) is reduced to maximizing $H(X_{\mathcal{A}(\mathcal{S})})$.  Therefore, it is reasonable to define the utility of $\mathcal{S}$ as $g(\mathcal{S})=H(X_{\mathcal{A}(\mathcal{S})})$ and we next show that $H(X_{\mathcal{A}(\mathcal{S})})$ is non-decreasing and submodular.
%All proofs in this section are stated in the Appendix.
\begin{lemma}
\label{lem:submodular}
$H(X_{\mathcal{A}(\mathcal{S})})$ is  non-decreasing and submodular.
\end{lemma}
%The proof is provided in the appendix.
%\proof $g(\mathcal{S})$ is clearly non-decreasing according to \emph{information never hurts principle} \citep{krause2005note}.
%Moreover, consider any $\mathcal{S}_1 \subseteq \mathcal{S}_2\subseteq \Omega$, we have $\mathcal{A}(\mathcal{S}_1) \subseteq \mathcal{A}(\mathcal{S}_2)$, it follows that $g(\mathcal{S}_1\cup\{q\})-g(\mathcal{S}_1\cup\{q\})=H(X_{\mathcal{A}(\mathcal{S}_1\cup\{q\})})-H(X_{\mathcal{A}(\mathcal{S}_1)})=H(X_{\mathcal{A}(\mathcal{S}_1)\cup\mathcal{Z}(\{q\})})-H(X_{\mathcal{Z}(\mathcal{A})})\geq H(X_{\mathcal{A}(\mathcal{S}_2)\cup\mathcal{A}(\{q\})})-H(X_{\mathcal{A}(\mathcal{S}_2)})=g(\mathcal{S}_2\cup\{q\})-g(\mathcal{S}_2\cup\{q\})$. The inequality is due to $H(X_{\mathcal{A}})$ as a function of $\mathcal{A}$ is submodular \citep{krause2005note}. Therefore, $g(\mathcal{S})$ is also non-decreasing and submodular. \endproof
%
%We remark that although we assume an entropy-like utility function in the above discussion,  all results derived in this paper apply to any utility function that is non-decreasing and submodular.

%$\mathcal{S}\subseteq \mathcal{Q}$ from a customer, $f(\mathcal{S})$ represents the amount of uncertainty that can be reduced about that customer's segment. The only requirement we make for $f$ is that it is monotone non-decreasing and submodular. We next give an example illustrating how the quality of questions is captured by a monotone and submodular function.

\subsubsection{Question Scanning Process} We use a Markov process to model the customer's browsing behavior when interacting with a sequence of quiz questions. Our model, which is called \emph{Cascade Browse Model},  is similar to the Cascade Model developed in \citep{craswell2008experimental}, which provides the best explanation for position bias of organic search results. We define the \emph{answer-through-rate}
$p^+_q\in[0,1]$ of a question $q\in \Omega$ as the probability that the customer chooses to answer $q$ after reading it. In principle, the value of this question-specific probability is decided by many factors, including the cognitive efforts required for understanding and answering the question, question sensitivity, etc. Instead of answering $q$, the customer may also (1) select ``Prefer Not to Answer'' (PNA) option, if any, with probability  $p^-_q$ to skip $q$, or (2) simply exit the quiz with probability $1-(p^+_{q}+p^-_{q})$. Each question $q$ has two continuation probabilities: $c_q^+$ and $c_q^-$. $c_q^+$ (resp. $c_q^-$) represents the probability that the customer continues reading the next question after answering (resp. PNA) $q$.

\emph{Some Basics:} Throughout this paper, we use capital letter to denote \emph{sequence} and calligraphy letter to denote \emph{set}. For example, $\mathcal{Q}$ denotes a set of questions and $Q$ denotes a sorted sequence of $\mathcal{Q}$. For a given sequence of questions $Q$, let $Q[i]$ denote the question scheduled at slot $i$, we use $Q_{\leq i}$ (resp. $Q_{< i}$, $Q_{> i}$, $Q_{\geq i}$) to denote the subsequence of $Q$ which is scheduled no later than (resp. before, after, no earlier than) slot $i$. Given two sequences $Q$ and $Q'$, we define $Q \oplus Q'$ as a new sequence by first displaying $Q$ and then displaying $Q'$.  For notational simplicity, we define $c_q=p^+_{q}c_{q}^+ +  p^-_{q} c_{q}^-$ as the aggregated continuation probability of $q$. We use $|\mathcal{S}|$ (resp. $|S|$) to denote the size of a set  $\mathcal{S}$ (resp. a sequence $S$). We summarize the question scanning process under the cascade browse model as follows.

  \begin{center}
\framebox[0.9\textwidth][c]{
\enspace
\begin{minipage}[t]{0.85\textwidth}
\small
\begin{itemize}
\item Starting with the first slot $i=1$.
\item After reading $Q[i]$, the customer chooses one of the following five actions to take:
 \begin{enumerate}
 \item Answer $Q[i]$  and  \begin{enumerate}
  \item continue to read the next question with probability   $p^+_{Q[i]}c_{Q[i]}^+$;
  \item exit the quiz with probability $p^+_{Q[i]}(1-c_{Q[i]}^+)$.
  \end{enumerate}
  \item PNA $Q[i]$ and
   \begin{enumerate}
   \item continue to  read the next question with probability   $p^-_{Q[i]} c_{Q[i]}^-$;
   \item exit the quiz with probability $p^-_{Q[i]} (1-c_{Q[i]}^-)$.
    \end{enumerate}
  \item Exit the quiz with probability $1-(p^+_{Q[i]}+p^-_{Q[i]})$.
  \end{enumerate}
% \item After answering (resp. skipping) the $Q[i]$, the customer continues to read the next question with probability  $c_{Q[i]}^+$ (resp.  $c_{Q[i]}^-$ ); otherwise, exit the quiz.
 \item The above process repeats until the customer exits the quiz or no more questions remain.
\end{itemize}
\end{minipage}
}
\end{center}
\vspace{0.1in}

We next introduce an important definition.
\begin{definition}[Reachability of a Question]
Given  a sequence of questions $Q$, for each $i\in\{1, 2, \cdots, |Q|\}$, we define the reachability $C_{Q[i]}$ of the $i$-th question $Q[i]$   as the probability of $Q[i]$ being read. Hence, $C_{Q[i]}$  can be written as:
\[C_{Q[i]}= \prod_{q\in Q_{<i}}c_{q}\]
%where $Q_{\overleftarrow{q}}$ denotes the subsequence of questions in $Q$ which is displayed before $q$.
\end{definition}

%In Section \ref{sec:5}, we also consider an advanced model by assuming slot-dependent answer-through-rate, e.g., the answer-through-rate of question $Q[i]$ is $\lambda_i p^+_{Q[i]}$ where  $\lambda_i\leq 1$ is a slot-dependent decay factor.

\subsection{Problem Formulation}
\label{sec:2.1}
Given any sequence of questions $Q$, we define its expected utility as
\[f(Q)=\sum_{\mathcal{S}\subseteq \mathcal{Q}}\Pr[\mathcal{S}|Q] g(\mathcal{S})\]
where $\Pr[\mathcal{S}|Q]$ denotes the probability of $\mathcal{S}$ being answered. Our objective is to identify a sequence of questions $Q$ that maximizes $f(Q)$ subject to a cardinality constraint $b$. We next present the formal definition of our problem $\textbf{P1}$.
 \begin{center}
\framebox[0.45\textwidth][c]{
\enspace
\begin{minipage}[t]{0.45\textwidth}
\small
$\textbf{P1}$
$\max f(Q)$
\textbf{subject to:} $|Q|\leq b$;
\end{minipage}
}
\end{center}
\vspace{0.1in}
The following theorem states that this problem is intractable in general.
\begin{theorem}
\label{thm:hard}
Problem $\textbf{P1}$ is NP-hard.
\end{theorem}
%The proof is provided in the appendix.

{
In the following, we show that there is no hope of finding any polynomial time algorithm achieving expected utility larger than $(1-1/e)$ of the optimal solution.
\begin{theorem}
\label{thm:harder}
For any $\epsilon>0$, problem $\textbf{P1}$ cannot be approximated in polynomial time within a ratio of $(1-1/e+\epsilon)$, unless $P=NP$.
\end{theorem}
}

{\subsection{Connection to Assortment Optimization}
 \label{sec:connection}
 Despite our focus on quiz design, the general framework of this study is well suited for many operational optimization problems, including assortment optimization. Assortment optimization is a core research problem in operations and revenue management. We next identify a class of assortment optimization problems which fit into our model. As a result, all results developed in this paper are applicable to those assortment optimization problems to obtain an approximate solution.

By abuse of notations, assume the platform has a set $\Omega$ of products  and $b$ vertically differentiated display positions \citep{aouad2015display}. In assortment optimization, we aimed at  allocating up to $b$ products from $\Omega$ to those display locations to maximize the expected revenue. Suppose  the customers' purchasing behavior is captured by a classic ``consider-then-choose'' model, where the customer first forms a \emph{consideration set} by  ``considering'' some subset of the original group of displayed products, then makes a purchasing decision from among her consideration set. We next describe each step in more detail.
\begin{enumerate}
\item \emph{Forming a consideration set.} Each product $q\in \Omega$ has a \emph{consider-through-rate} $p^+_q\in[0,1]$ which is the probability that the customer adds $q$ to her consideration set after browsing it. In practice, the  consider-through-rate  of a product is dependent on the type of the product and the costumer's preference. Let $p^-_q=1-p^+_q$ denote the probability that the customer skips $q$ after browsing it. Each product $q \in \Omega$ is also associated with two continuation probabilities $c^+_q$ and $c^-_q$: $c^+_q$ represents the probability that the customer continues browsing the next product after adding  $q$ to her consideration set, and $c^-_q$ represents the continuation probability after skipping $q$. Given an assortment $Q$ of products, let $\mathcal{S}$ denote  the (random) consideration set induced by the assortment $Q$.
\item \emph{Picking a product from the consideration set.} For any realized consideration set $\mathcal{S}$, the customer purchases a product $q\in \mathcal{S}$  with probability $\pi(q, \mathcal{S})$ according to a standard choice model. One widely adopted choice model is the  multinomial logit model \citep{rusmevichientong2014assortment}, where each product $q\in \Omega$ is associated with a weight $w_q$. The probability that $q \in \mathcal{S}$ is picked under the multinomial logit model is  $\pi(q, \mathcal{S}) = \frac{w_q}{1+W(\mathcal{S})}$ where $W(\mathcal{S})=\sum_{q\in \mathcal{S}} w_q$.
\end{enumerate}

 Assume each product $q\in \mathcal{S}$ has revenue $\gamma_q$, then the expected revenue $r(\mathcal{S})$ of a realized consideration set $\mathcal{S}$ can be computed as follows:
 \begin{equation}
 \label{eq:revenue}
 r(\mathcal{S})=\sum_{q\in \mathcal{S}} \pi(q, \mathcal{S}) \gamma_q
 \end{equation}
% Assume that each product $q\in \Omega$ has revenue $v_q$, then the expected revenue $g(\mathcal{S})$ gained from $\mathcal{S}$ is
%\begin{equation}
%g(\mathcal{S}) = \sum_{q\in \mathcal{S} } \frac{v_q w_q}{1+W(\mathcal{S})}
%\end{equation}

For any assortment $Q$, let $\Pr[\mathcal{S}|Q]$ denote the probability that $\mathcal{S}$ is the realized consideration set given that $Q$ is displayed to the customer. Then our objective in assortment optimization is to find an assortment $Q$ that maximizes the the expected revenue $f(Q) = \sum_{\mathcal{S} \subseteq \mathcal{Q}}\Pr[\mathcal{S}|Q] r(\mathcal{S})$ subject to a cardinality constraint $|Q| \leq b$. It is easy to verify that the framework of $\textbf{P1}$ is general enough to subsume the above assortment optimization problem when $r$ is monotone and submodular. The hardness result developed in Theorem \ref{thm:harder} also applies to the above assortment optimization problem. Since the above framework is not restricted to any particular underlying choice model, our results may provide insights into a broad category of assortment optimization problems \citep{davis2015assortment} whose revenue function $r$ is monotone and submodular. Recently, \cite{han2019assortment} explores the relation between the assortment optimization and submodularity under the  multinomial logit model, and they successfully identify  the conditions under which the revenue function $r$  is submodular. They
show that submodularity is closely related to cannibalization and that it can be observed in many
important applications. Notably, when all products have the same revenue under the  multinomial logit model, i.e., $\forall q\in \Omega, \forall q'\in \Omega: \gamma_q=\gamma_{q'}$, the revenue function $r$ is monotone and submodular. The same observation also holds under a more general markov chain based choice model \citep{blanchet2016markov}, which provides a good approximation to a large class of existing choice models.}

\section{Warming Up: Question Selection and Sequencing with No PNA Option}
\label{sec:3}
We first study the case where ``PNA'' is not an option. In other words, the customer is left with two options after reading a question: either answer it or exit the quiz. The reason for investigating this restricted case is twofold: (1) Although the benefit of including PNA option has been discussed in many existing work \citep{schuman1996questions,hawkins1981uninformed}, opponents believe that providing a PNA option could have negative impact on the quality of the answer because some customers tend not to answer the question so as to minimize the effort required to complete the quiz by simply ticking PNA option \citep{poe1988don,sanchez1992probing}. Since both arguments are empirically validated by previous studies, we decide to study both cases in this work. (2) Technically speaking, the case with no PNA option is a special case of the original problem (by setting $p^-_q=0$ for every $q$ in the original problem), starting with this simplified case makes it easier to explain our approach used to solve the general case. We first present a simplified question scanning process under this restricted case as follows:
  \begin{center}
\framebox[0.9\textwidth][c]{
\enspace
\begin{minipage}[t]{0.85\textwidth}
\small
\begin{itemize}
\item Starting with the first slot $i=1$.
\item After reading $Q[i]$, the customer chooses one of the following three actions to take:
\begin{enumerate}
 \item Answer $Q[i]$ and
 \begin{enumerate}
 \item continue to read the next question  with probability   $p^+_{Q[i]}c_{Q[i]}^+$;
 \item exit the quiz with probability $p^+_{Q[i]}(1-c_{Q[i]}^+)$.
 \end{enumerate}
  %\item PNA $Q[i]$ with probability $p^-_{Q[i]}$ and continue to read the next question with probability   $c_{Q[i]}^-$.
  \item Exit the quiz with probability $1-p^+_{Q[i]}$.
  \end{enumerate}
 \item The above process repeats until the customer exits the quiz or no more questions remain.
\end{itemize}
%\[C_q= \prod_{q'\in Q_{\overleftarrow{q}}}(p^+_{q'}c_{q'}^+ +  p^-_{q'} c_{q'}^-)\]
\end{minipage}
}
\end{center}
\vspace{0.1in}

\subsection{Algorithm Design}
 Before presenting our algorithm,  we first introduce a useful property of any optimal solution. In particular, given an optimal solution $Q^*$, we show that little is lost by discarding those questions whose reachability is sufficiently small. Notably, the following lemma holds even under the general model where PNA is offered as an option for each question.
\begin{lemma}
\label{lem:1}
For any $\rho\in[0,1]$, there is a solution $Q$ of value at least $(1-\rho)f(Q^*)$ such that %$Q_\rho \overset{\mathrm{subsequence}}{\preceq} Q^*$ and
$|Q|\leq b$ and $\forall i\in \{1, 2, \cdots, |Q|\}: C_{Q[i]} \geq \rho$.
\end{lemma}

 Lemma \ref{lem:1} allows us to  ignore those questions whose reachability is small, at the expense of a bounded decrease in utility.  A similar observation was made in \citep{kempe2008cascade} who considers a linear utility function. We generalize their results to monotone and submodular functions. This motivates us to introduce a new problem $\textbf{P2}$ by only considering those questions whose reachability is sufficiently high. The  objective function of $\textbf{P2}$ is \[u(q, \mathcal{S})= p_q^+ g(\mathcal{S}\cup\{q\})+(1-p_q^+)g(\mathcal{S})\] The goal of  $\textbf{P2}$ is to find a solution $(q, \mathcal{S})$ that maximizes $u(q, \mathcal{S})$ subject to three constraints. After solving $\textbf{P2}$ (approximately) and obtaining a solution $(q', \mathcal{S}')$, we build the final solution to the original problem based on  $(q', \mathcal{S}')$.

  \begin{center}
\framebox[0.45\textwidth][c]{
\enspace
\begin{minipage}[t]{0.45\textwidth}
\small
$\textbf{P2}$
\emph{Maximize$_{q, \mathcal{S}}$ $u(q, \mathcal{S})$}\\
\textbf{subject to:}
\begin{equation*}
\begin{cases}
%$\forall \theta_{\mathcal{S}}> 0: |\mathcal{S}|\leq 1 $\\
-\sum_{l\in \mathcal{S}} \log (p^+_{l}c_{l}^+) \leq -\log \rho\quad\mbox{(C1)}\\
% \sum_{v\in V}\sum_{d\in D} y_{vd}\leq K \quad(C2)\\
% |\mathcal{S}|\leq b-1 \quad\mbox{(C2)}\\
|\mathcal{S}| < b \quad \mbox{(C2)}\\
\mathcal{S} \subseteq \Omega \setminus \{q\}
\end{cases}
\end{equation*}
\end{minipage}
}
\end{center}
\vspace{0.1in}

 We next take a closer look at $\textbf{P2}$. Intuitively, the solution to $\textbf{P2}$ is composed of two parts: $\mathcal{S}$ and $q$ where  $q$ is scheduled after $\mathcal{S}$. The reason we separate $q$ from other questions in $\mathcal{S}$ is that $q$ is scheduled at the last slot, thus there is no restriction on $q$'s aggregated continuation probability $c_q$. Constraint (C1) ensures that the reachability of every question in our solution is sufficiently high, and constraint (C2) ensures the feasibility of the final solution, e.g., the size of our solution is upper bounded by $b$.

 In the rest of this section, we focus on solving $\textbf{P2}$. We first show that $u(q, \mathcal{S})$ is a monotone and  submodular function of $\mathcal{S}$.
\begin{lemma}
\label{lem:9}
For any fixed $q\in \Omega$, $u(q, \mathcal{S})$ is a monotone and  submodular function of $\mathcal{S}$.
\end{lemma}

As a consequence of Lemma \ref{lem:9}, for any fixed $q$,  $\textbf{P2}$ is a monotone submodular maximization problem subject to two linear constraints (constraints (C1) and (C2)), and there exists a $(1-1/e-\epsilon)$ approximate algorithm to this problem \citep{kulik2009maximizing} where $e$ is a constant whose value is arbitrarily close to 2.718. In order to solve $\textbf{P2}$, we exhaustively
try all possibilities of $q$ which will be scheduled at the last slot, for each $q$, we run a $(1-1/e-\epsilon)$ approximate algorithm to obtain a candidate solution $\mathcal{S}$. Among all candidate solutions, assume $(q', \mathcal{S}')$ has the largest utility, $S'\oplus q'$ is returned as the final solution to the original problem where $S'$ is an arbitrary sequence of $\mathcal{S}'$. We present the detailed description of our solution in Algorithm \ref{alg:greedy-peak}.
\begin{algorithm}[h]
{\small
\caption{Question Selection and Sequencing with No PNA option}
\label{alg:greedy-peak}
\textbf{Input:} $\rho, b, \Omega$.\\
\textbf{Output:} $Q^{\mathrm{Alg1}}$.
\begin{algorithmic}[1]
\STATE Set $\mathcal{S}'=\emptyset, q'=\emptyset$.
\FOR{$q\in \Omega$}
\STATE Fix $q$, apply a $(1-1/e-\epsilon)$ approximate algorithm \citep{kulik2009maximizing} to solve $\textbf{P2}$  and obtain $\mathcal{S}$ \label{line:2}
\IF {$u(q, \mathcal{\mathcal{S}})> u(q', \mathcal{S}')$}
\STATE $\mathcal{S}' \leftarrow \mathcal{S}, q'\leftarrow q$
\ENDIF
\ENDFOR
%\STATE $\mathcal{Q}_{\mathrm{Alg1}}\leftarrow \mathcal{S}'\cup\{q'\}$
\STATE $Q^{\mathrm{Alg1}}\leftarrow S' \oplus \{q'\}$ where $ S'$ is an arbitrary sequence of $\mathcal{S}'$ \label{line:11}
\RETURN $Q^{\mathrm{Alg1}}$
\end{algorithmic}
}
\end{algorithm}

\subsection{Performance Analysis}
We next analyze the performance bound of Algorithm \ref{alg:greedy-peak}. We first present some preparatory lemmas.
Since for each $q\in \Omega$, we find a $(1-1/e-\epsilon)$ approximate solution and $(q', \mathcal{S}')$ has the maximum utility among all returned solutions, the following lemma holds.
\begin{lemma}
\label{lem:90}
$(q', \mathcal{S}')$ is a $(1-1/e-\epsilon)$ approximate solution to $\textbf{P2}$.
\end{lemma}

Now we are ready to provide a performance bound on the final solution $Q^{\mathrm{Alg1}}$. We first show that for any  $\rho\in[0,1]$, $f(Q^{\mathrm{Alg1}})\geq \rho u(q', \mathcal{S}')$, i.e., the utility of $Q^{\mathrm{Alg1}}$ is close to the value of $u(q', \mathcal{S}')$.
\begin{lemma}
\label{lem:3}
For any $\rho\in[0,1]$, $f(Q^{\mathrm{Alg1}})\geq \rho u(q', \mathcal{S}')$, where $Q^{\mathrm{Alg1}}=  S' \oplus \{q'\}$ where $S'$ is an arbitrary sequence of $\mathcal{S}'$. (Refer to Line \ref{line:11} of Algorithm \ref{alg:greedy-peak} for details.)
\end{lemma}

We present the main theorem as follows.
\begin{theorem}
\label{thm:1}
For any $\rho\in[0,1]$, $f(Q^{\mathrm{Alg1}})\geq \rho (1-\rho)(1-1/e-\epsilon) f(Q^*)$.
\end{theorem}

 \begin{corollary}
By choosing $\rho=1/2$, we have $f(Q^{\mathrm{Alg1}})\geq \frac{1-1/e-\epsilon}{4} f(Q^*)$.
\end{corollary}
\section{Question Selection and Sequencing under General Model}
\label{sec:4}
We now add  PNA option to our model. The workflow of our solution is similar in structure to Algorithm \ref{alg:greedy-peak}: we first introduce a new problem, then build the final solution based on the solution to that new problem. However, the way we define the new problem as well as the analysis of our solution are largely different from the one used in the previous model.
\subsection{Algorithm Design}
For any given sequence of questions $Q$, we use  $\mathcal{R}(Q)$ and $\mathcal{R}(\mathcal{Q})$ interchangeably to denote a random set obtained by including each question $q\in Q$ independently with probability $p^+_q$. We first introduce a new problem $\mathbf{P3}$ whose objective function is $v(q, \mathcal{S})= \mathbb{E}[g(\mathcal{R} (\mathcal{S}\cup\{q\}))]$. The goal of $\mathbf{P3}$ is to find a solution $(q, \mathcal{S})$ that maximizes function $v$. Similar to constraints (C1) and (C2) used in $\mathbf{P2}$, we use  constraint (C3) (resp. constraint (C4)) to ensure that all selected questions can be reached with high probability (resp. the size of the solution is upper bounded by $b$).

 \begin{center}
\framebox[0.45\textwidth][c]{
\enspace
\begin{minipage}[t]{0.45\textwidth}
\small
$\textbf{P3}$
\emph{Maximize$_{q, \mathcal{S}}$ $v(q, \mathcal{S})$}\\
\textbf{subject to:}
\begin{equation*}
\begin{cases}
%$\forall \theta_{\mathcal{S}}> 0: |\mathcal{S}|\leq 1 $\\
-\sum_{l\in \mathcal{S}} \log c_{l} \leq -\log \rho\quad\mbox{(C3)}\\
% \sum_{v\in V}\sum_{d\in D} y_{vd}\leq K \quad(C2)\\
% |\mathcal{S}|\leq b-1 \quad\mbox{(C2)}\\
|\mathcal{S}| < b \quad \mbox{(C4)}\\
\mathcal{S} \subseteq \Omega\setminus \{q\}
\end{cases}
\end{equation*}
\end{minipage}
}
\end{center}
\vspace{0.1in}

In the following lemma we show that if $q$ is fixed, then $v(q, \mathcal{S})$,  as a function of $\mathcal{S}$, is monotone and submodular.
 \begin{lemma}
 \label{lem:67}
 For any fixed $q\in \Omega$, $v(q, \mathcal{S})$ is a monotone and submodular function of $\mathcal{S}$.
 \end{lemma}

The above lemma implies that for any fixed $q$, $\textbf{P3}$ is a monotone submodular maximization problem subject to two linear constraints (constraints (C3) and (C4)), and there exists a $(1-1/e-\epsilon)$ approximate algorithm to this problem \citep{kulik2009maximizing}. The rest of the design is similar to the approach developed in the previous section, we enumerate all possibilities of $q$, for each $q$, we run a $(1-1/e-\epsilon)$ approximate algorithm to obtain a candidate solution $\mathcal{S}$. Among all candidate solutions, assume $(q', \mathcal{S}')$ has the largest utility, we choose $S'\oplus q'$  as the final solution to the original problem where $S'$ is an arbitrary sequence of $\mathcal{S}'$. We present the detailed description of our solution in Algorithm \ref{alg:greedy-peak1}.

 %For a given $Q$, define $Q_L=\{Q[i]: i \leq L\}$ to be the set of questions whose reachability is no less than $\rho$.

%\begin{lemma}
%For any $\rho>0$, there is a solution $Q_\rho$ of value at least $(1-\rho)f(Q^*)$ such that $Q_\rho \overset{\mathrm{subsequence}}{\preceq} Q^*$ and $\forall q\in Q_\rho: C_q^{Q^*} \geq \rho$.
%\end{lemma}
%Assume $Q^*[k]$ is the last question in $Q^*$ whose rechability is no smaller than $\rho$, e.g.,  $k=\arg\max_{i} C_{Q^*[k]}^{Q^*}\geq \rho$, and we use $Q^*_{> k}$ (resp. $Q^*_{\leq k}$) to denote the sequence of all questions scheduled after (resp. before) $Q^*[k]$. For ease of presentation, we assume $Q^*[k]$ belongs to $Q^*_{\leq k}$. Then we have $C_{Q^*[k]}^{Q^*} c_{Q^*[k]} f(Q^*_{> k})\geq  f(Q^*)-f(Q^*_{\leq k})$, this is  because $Q^*_{> k}$ can be reached with probability $C_{Q^*[k]}^{Q^*} c_{Q^*[k]}$ and $f$ is a submodular function. It follows that $f(Q^*_{\leq k})\geq f(Q^*)-C_{Q^*[k]} c_{Q^*[k]} f(Q^*_{> k})$. Since $C_{Q^*[k]} c_{Q^*[k]} < \rho$ due to the definition of $k$, and $f(Q^*_{> k})\leq f(Q^*)$ due to $Q^*$ is the optimal solution, we have $f(Q^*_{\leq k})\geq (1-\rho)f(Q^*)$. Thus, $Q^*_{\leq k}$ is a valid $Q^*_L$.

\begin{algorithm}[h]
{\small
\caption{Question Selection and Sequencing with PNA option}
\label{alg:greedy-peak1}
\textbf{Input:} $\rho, b, \Omega$.\\
\textbf{Output:} $Q^{\mathrm{Alg2}}$.
\begin{algorithmic}[1]
\STATE Set $\mathcal{S}'=\emptyset, q'=\emptyset$.
\FOR{$q\in \Omega$}
\STATE Fix $q$, apply a $(1-1/e-\epsilon)$ approximate algorithm \citep{kulik2009maximizing} to solve $\textbf{P3}$  and obtain $\mathcal{S}$ \label{line:3}
\IF {$v(q, \mathcal{\mathcal{S}})> v(q', \mathcal{S}')$}
\STATE $\mathcal{S}' \leftarrow \mathcal{S}, q'\leftarrow q$
\ENDIF
\ENDFOR
%\STATE $\mathcal{Q}_{\mathrm{Alg1}}\leftarrow \mathcal{S}'\cup\{q'\}$
\STATE $Q^{\mathrm{Alg2}}\leftarrow S' \oplus \{q'\}$ where $ S'$ is an arbitrary sequence of $\mathcal{S}'$ \label{line:1}
\RETURN $Q^{\mathrm{Alg2}}$
\end{algorithmic}
}
\end{algorithm}

\subsection{Performance Analysis}
We still use $Q^*$ to denote the optimal solution to the original problem under the general model, and use $Q^*_{\leq k}$ to denote the longest prefix of $Q^*$ such that all questions in $Q^*_{\leq k}$ can be reached with probability no smaller than $\rho$, e.g.,  $k=\arg\max_{i} (C_{Q^*[i]}\geq \rho)$. We first show that the expected utility of random set $\mathcal{R}(Q^*_{\leq k})$ is at least $f(Q^*_{\leq k})$.
{\begin{lemma}
\label{lem:7}
$f(Q^*_{\leq k}) \leq \mathbb{E}[g(\mathcal{R}(Q^*_{\leq k}))]$.
\end{lemma}
%For notational simplicity, we use  $c_q$ to denote $p_qc^+_q+(1- p_q)c^-_q$ in the rest of the proof. Consider any question $q\in Q^*_{\leq k}$, assume $q$ is scheduled at slot $i\leq k$, let $Q^*_{\leq i}$ denote the subsequence of questions scheduled before $q$.

}

 We next prove that the utility of $Q^{\mathrm{Alg2}}$ is close to the expected utility of a random set $\mathcal{R}(Q^{\mathrm{Alg2}})$.
\begin{lemma}
\label{lem:9999999}For any $\rho\in[0,1]$,
\[f(Q^{\mathrm{Alg2}})\geq \rho \mathbb{E}[g(\mathcal{R}(Q^{\mathrm{Alg2}}))]\]
\end{lemma}

Now we are ready to present the main theorem of this paper.
{
\begin{theorem}
\label{thm:@}
For any $\rho\in[0,1]$, $f(Q^{\mathrm{Alg2}})\geq \rho(1-\rho)(1-1/e-\epsilon)f(Q^*)$.
\end{theorem}

 \begin{corollary}
 \label{col:1}
By choosing $\rho=1/2$, we have $f(Q^{\mathrm{Alg2}})\geq \frac{1}{4}(1-1/e-\epsilon)f(Q^*)$.
\end{corollary} }

{\paragraph{Contribution to assortment optimization.} Recall that we identify a group of assortment optimization problems that fit into our model in Section \ref{sec:connection}. We can readily use Algorithm \ref{alg:greedy-peak1} to solve them and find a $\frac{1}{4}(1-1/e-\epsilon)$ approximate solution. In particular, we follow the same procedure listed in Algorithm \ref{alg:greedy-peak1}, using $v(q, \mathcal{S})= \mathbb{E}[r(\mathcal{R} (\mathcal{S}\cup\{q\}))]$ where $r$ is the revenue function defined in (\ref{eq:revenue}), to find a solution for the assortment optimization problem.  When the revenue function $r$ is monotone and submodular, Algorithm \ref{alg:greedy-peak1} achieves the same  approximation ratio provided in  Corollary 6. As this performance bound does not depend on any particular underlying choice model, we believe that our results may provide guidance to a broad class of assortment optimization problems.}

\section{Numerical Experiments}
\label{sec:exp}
We next evaluate the performance of our proposed algorithm for solving the question selection and sequencing problem. Here we abbreviate Algorithm \ref{alg:greedy-peak1} as {\it QSS} algorithm. %We first evaluate the quality of {\it QSS} with respect to changes in the answer-through-rate and the continue probability. Then we compare  {\it QSS} with two benchmark heuristics as well as the optimal policy. At last, we explore how the expected utility changes when the PNA option is not offered.
%In Section \ref{subsec:imp1}, we focus on evaluating the quality of {\it QSS} with respect to changes in the answer-through-rate and the continue probability. In Section \ref{subsec:imp2}, we compare the performance of {\it QSS} with two benchmark heuristics on a wide range of parameters. In Section \ref{subsec:imp3}, we evaluate the quality of {\it QSS} as compared to the optimal policy, measured by the average ratio of the expected utility generated by the two. In Section \ref{subsec:imp4}, we explore how the expected utility changes when the PNA option is not offered, and discuss the conditions under which PNA should be added or removed.

\subsection{Impact of the Answer-through-Rate and the Continue Probability}
\label{subsec:imp1}

\begin{figure*}[hptb]
\begin{center}
\includegraphics[scale=0.18]{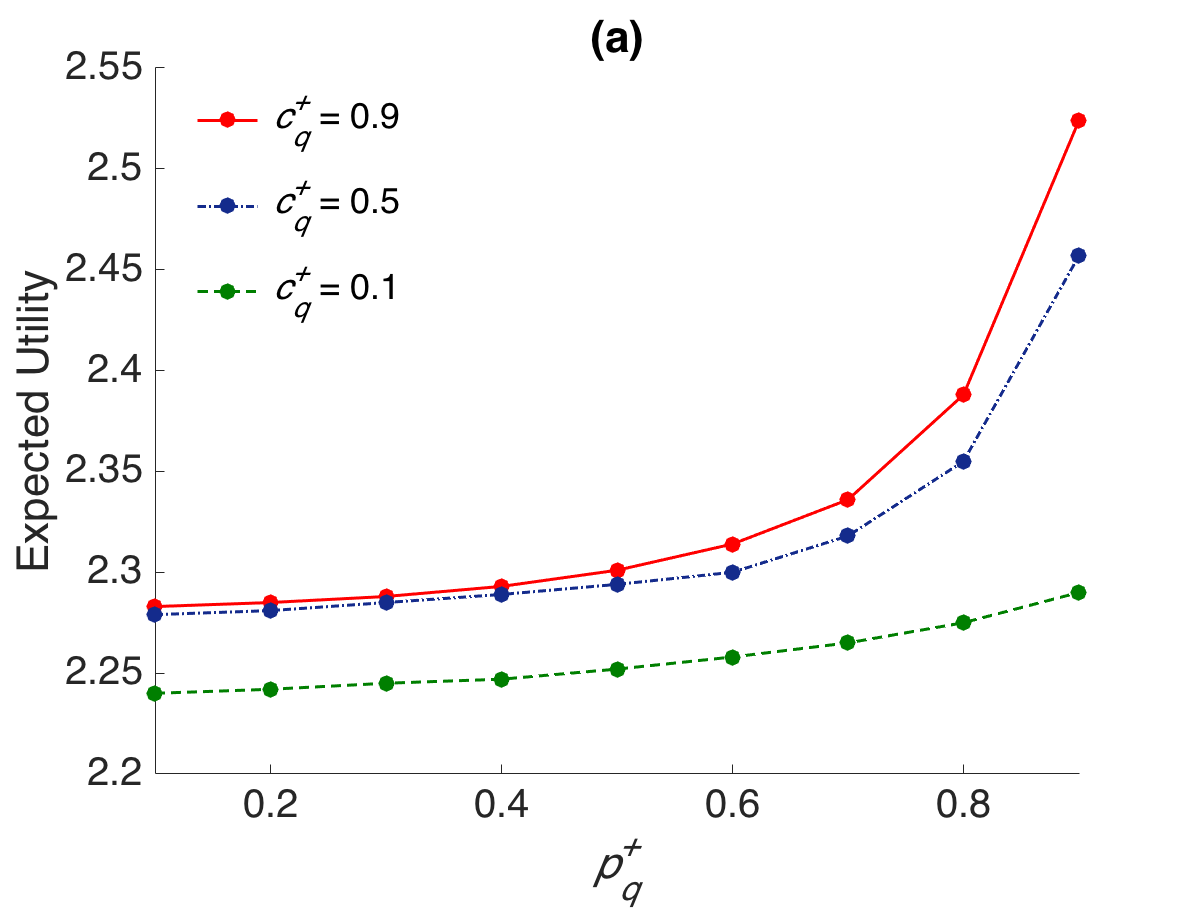}
%\hskip -0.1in
\includegraphics[scale=0.18]{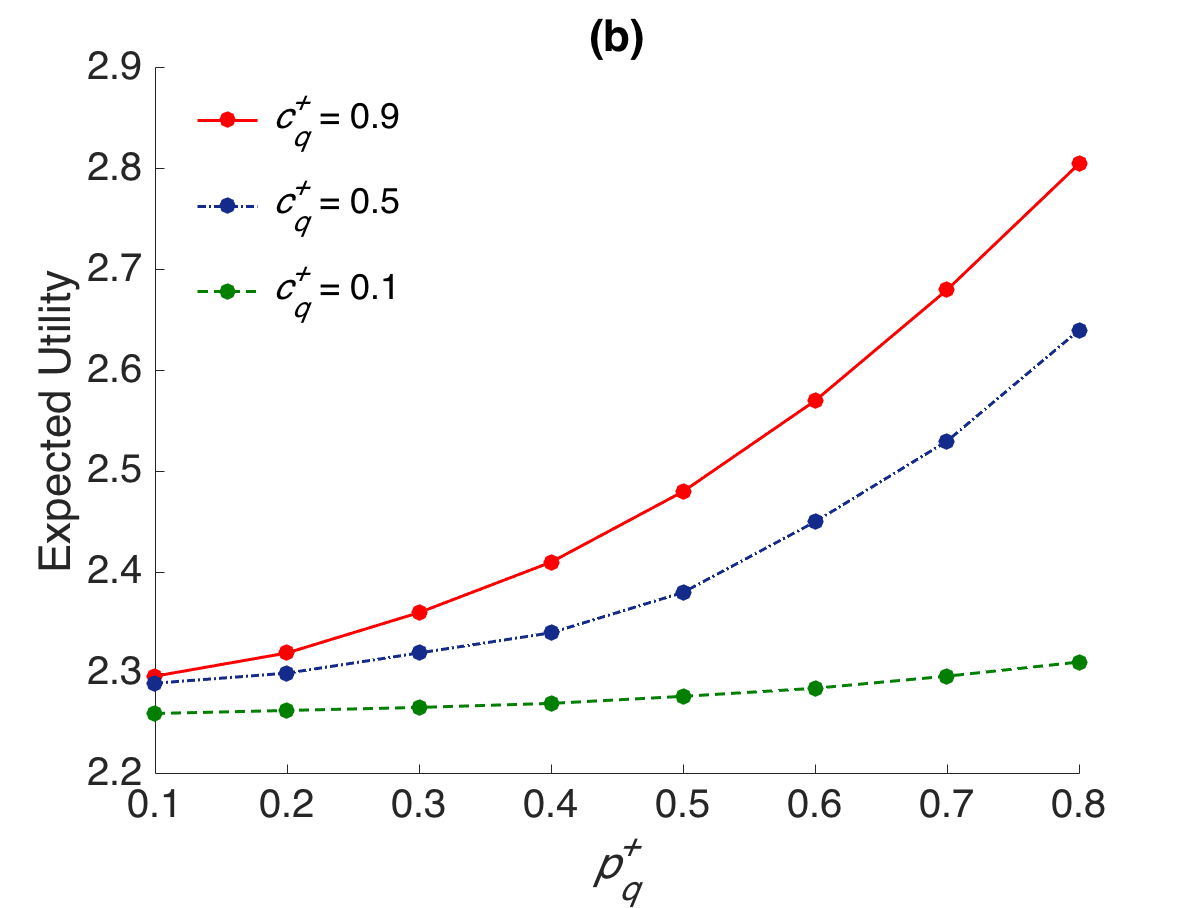}
\includegraphics[scale=0.18]{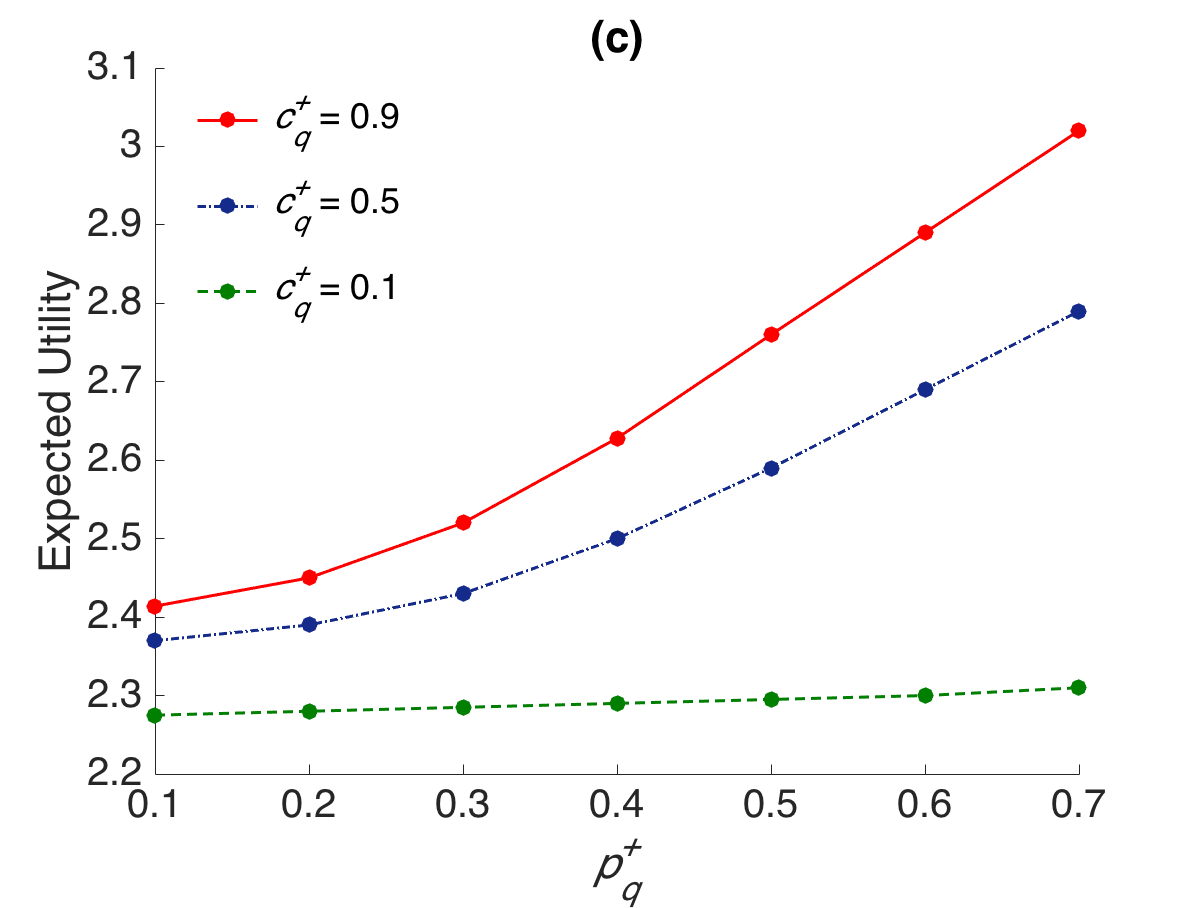}
%\hskip -0.1in
\includegraphics[scale=0.18]{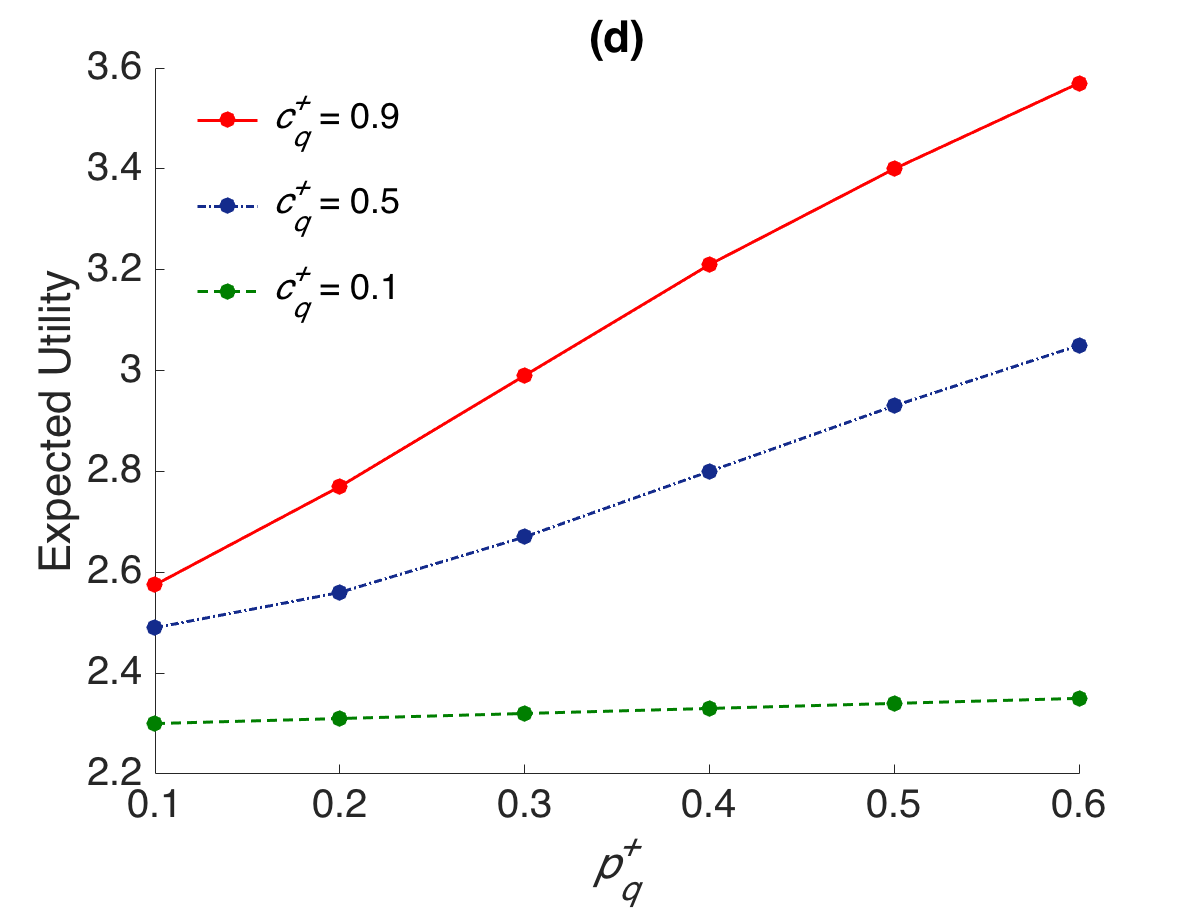}
\caption{Expected utility yielded by {\it QSS} with respect to changes in the answer-through-rate}
\label{fig:utility}
\end{center}
\end{figure*}

The experimental design to generate the problem instances is as follows. We consider $12$ quiz questions, each with $5$ choices and a PNA option. We assume each question covers one unique attribute of a customer, and the random variable associated with the attribute takes $5$ possible values. The frequentist probability of each value of the variable is randomly generated from $[0, 1]$ with unity-based normalization.  We set the cardinality constraint parameter as $b=6$, i.e., our task is to select $6$ out of $12$ quiz questions with the goal of maximizing the expected utility. To measure the expected utility of a given group of quiz questions, we calculate the expected entropy of the observed attributes covered by the questions.

We first evaluate how {\it QSS} performs under different settings of parameters: the answer-through-rate $p^+_q$, the continue probability $c^+_q$, and the click probability of the PNA option $p^-_q$. The results are illustrated in Figure \ref{fig:utility}. We set $p^-_q$ to $0.1$, $0.2$, $0.3$ and $0.4$ in Figure \ref{fig:utility}(a)-(d), respectively, with $c^-_q$ set to $0.5$. As shown in all four plots in Figure \ref{fig:utility}, the $x$-axis represents the value of $p^+_q$ and the $y$-axis represents the expected utility generated by {\it QSS}. Note that in each plot, $p^+_q$ only takes values that are feasible with the corresponding value of $p^-_q$, because the sum of these two parameters must be no greater than $1$. Each data point is the average value, over $1000$ problem instances, of the expected utility generated by {\it QSS}.

We observe that the expected utility yielded by {\it QSS} increases as $p^+_q$ increases. This is expected because when other parameters are fixed, a larger $p^+_q$ indicates a higher probability that the customer answers a question, and then results in a larger expected utility. As shown in each plot, for a fixed $p^+_q$, the expected utility is larger for a larger $c^+_q$. The underlying reason is straightforward, that when the customer continues to read the next question instead of exiting the quiz, with a higher probability, she can answer more questions, giving us a larger expected utility as a result. We also observe that when other parameters are fixed, a larger $p^-_q$ leads to a larger expected utility. Moreover, compare the four plots in Figure \ref{fig:utility}, we find that when $p^-_q$ is small (e.g., $p^-_q=0.1$ in Figure \ref{fig:utility}(a)), the expected utility grows rapidly when $p^+_q \geq 0.6$. As $p^-_q$ becomes larger (e.g., $p^-_q=0.4$ in Figure \ref{fig:utility}(d)), the expected utility grows rapidly even when $p^+_q$ is relatively small (e.g., $p^+_q=0.2$). The reason is that when $p^+_q$ is small, a small $p^-_q$ indicates a higher chance of customer quitting the quiz; when $p^-_q$ becomes larger, the customer has a lower chance of quitting thus even a slight change in $p^+_q$ can make a big difference in the expected utility.

\subsection{Comparison with the Heuristics}
\label{subsec:imp2}

\begin{figure*}[hptb]
\begin{center}
\includegraphics[scale=0.18]{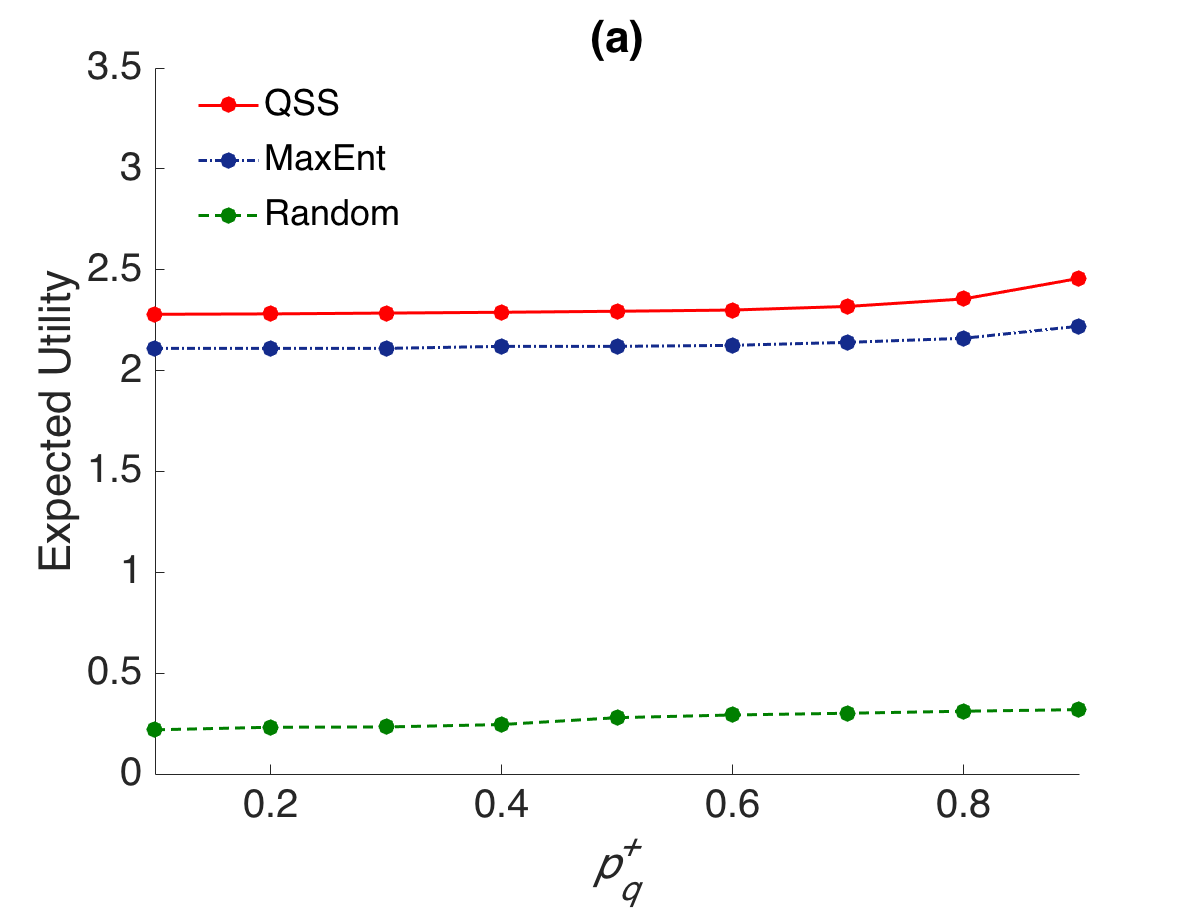}
%\hskip -0.1in
\includegraphics[scale=0.18]{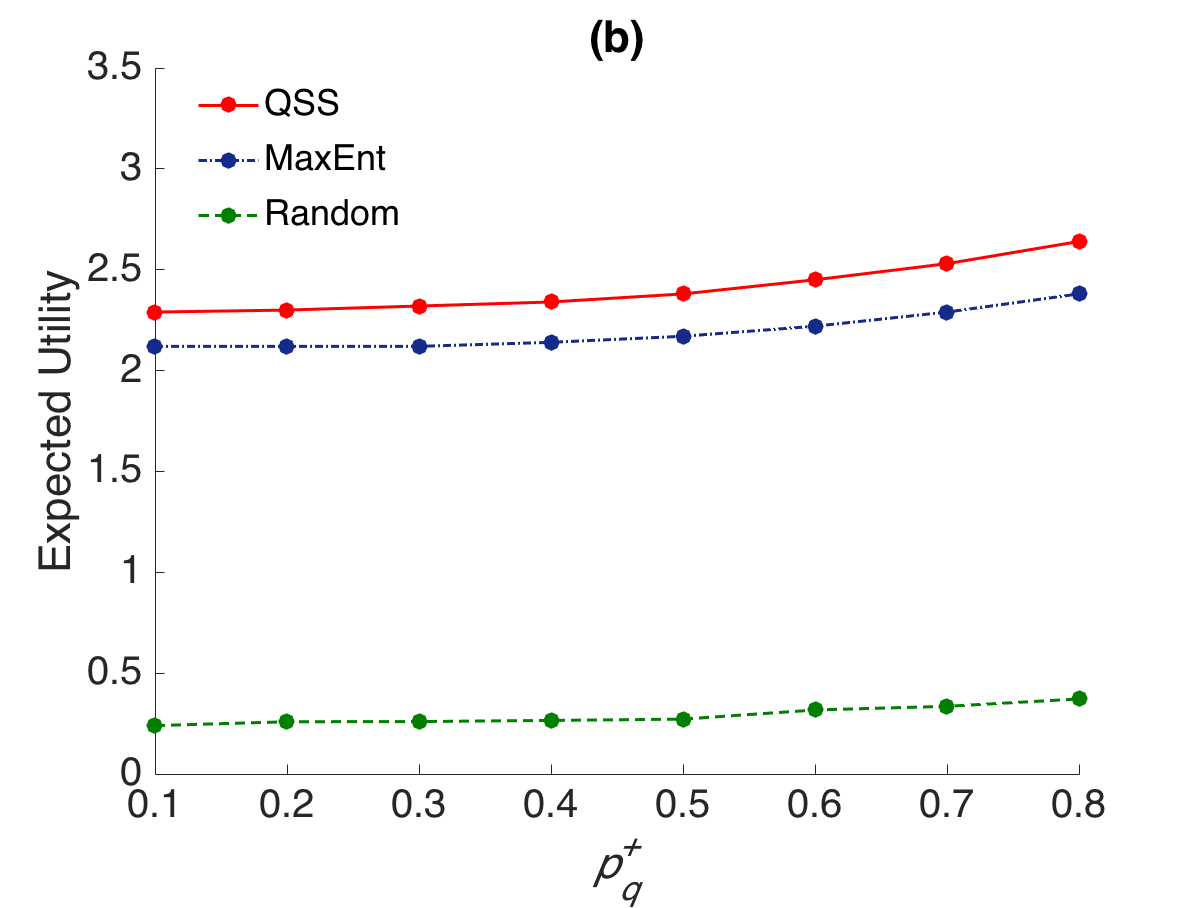}
\includegraphics[scale=0.18]{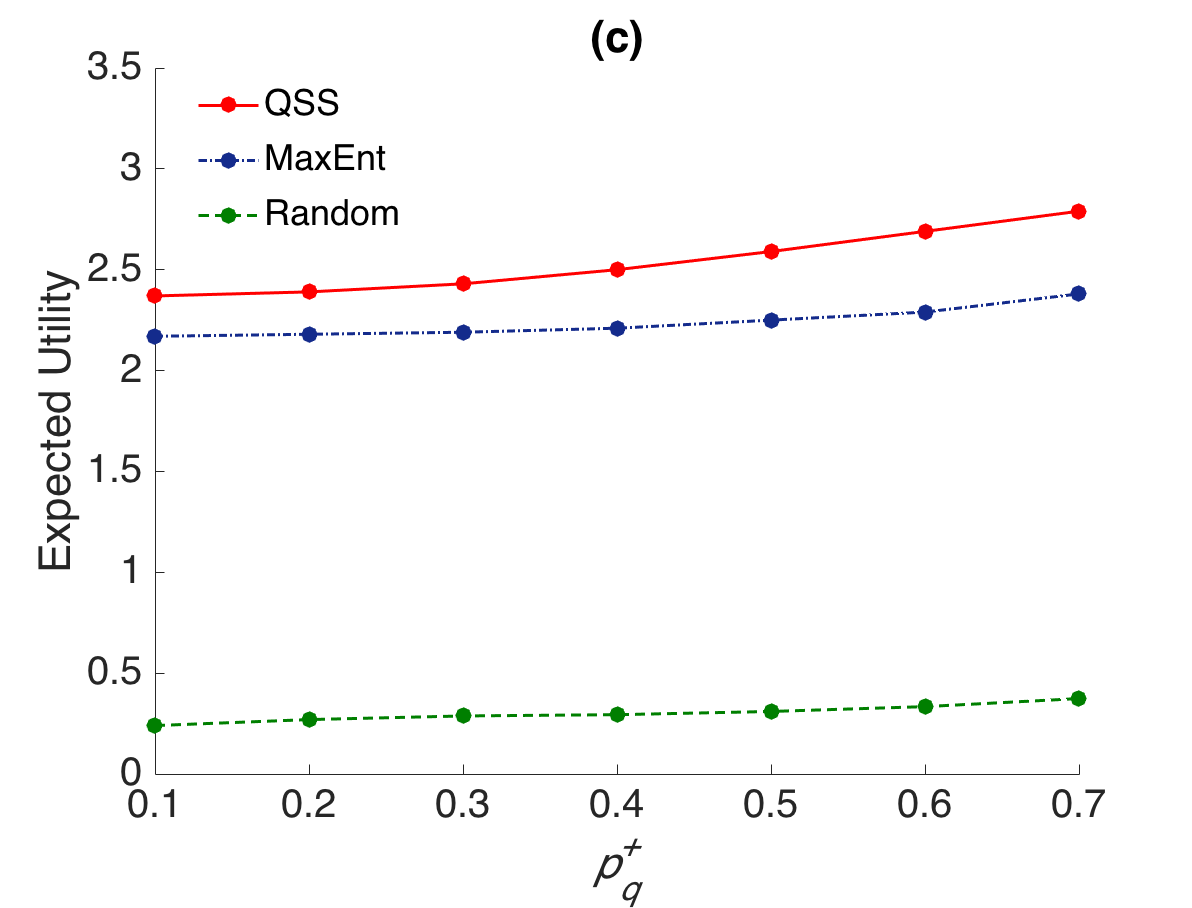}
%\hskip -0.1in
\includegraphics[scale=0.18]{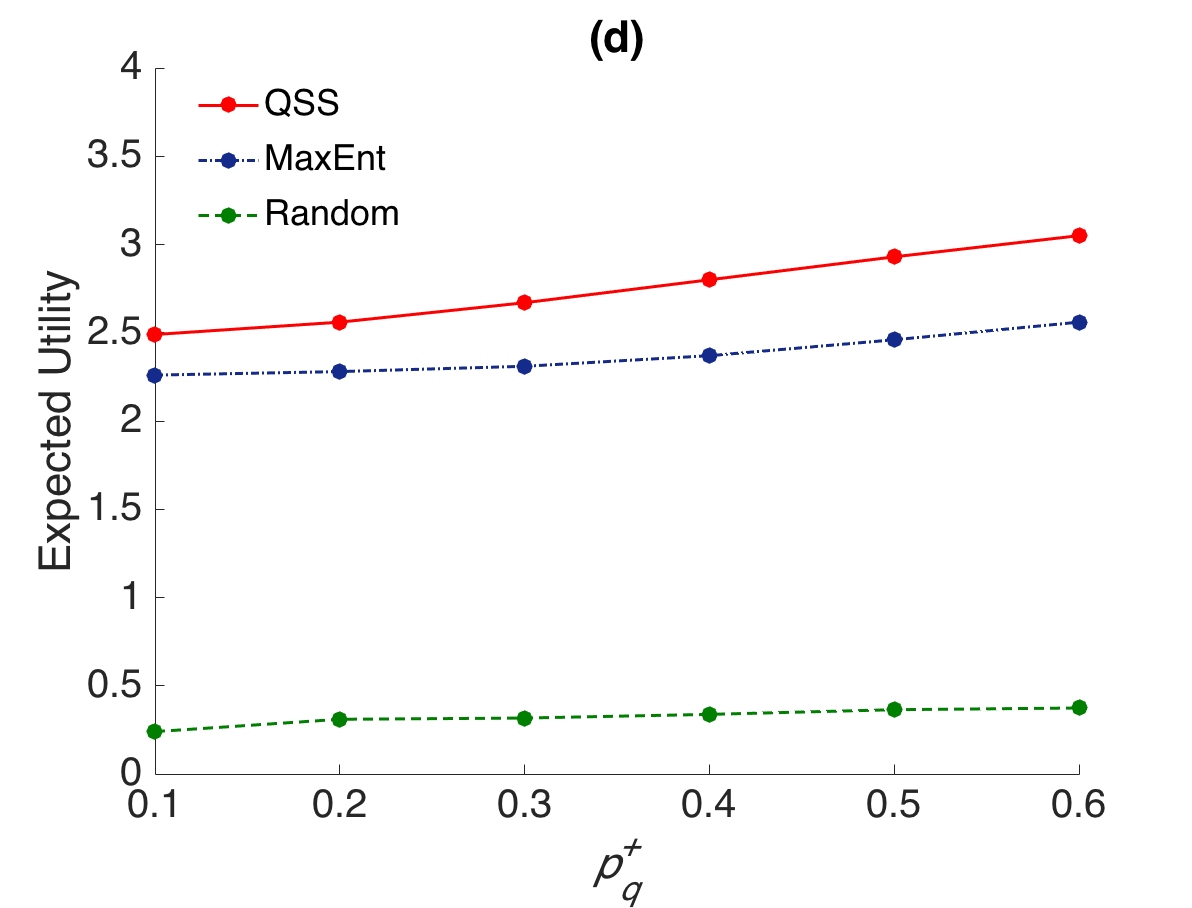}
\caption{Expected utility yielded by {\it QSS} compared with benchmark policies}
\label{fig:benchmark}
\end{center}
\end{figure*}

Next we evaluate the performance of {\it QSS} compared with two benchmark approaches. {\it Random} randomly generates a sequence of questions as an output. {\it MaxEnt} first enumerates all the subsets of questions that satisfy the cardinality constraint. Then it selects the subset that yields the maximum entropy of the covered attributes. Finally it generates a sequence from this subset by randomly arranging the questions in the subset. Note that this algorithm only considers the entropy in the best case but ignores the uncertainty in a customer's behavior.  { We first show that the worst-case performance of these two benchmarks could be arbitrarily bad. This can be proved through construction as follows: Assume the ground set $\Omega$ contains $n$ questions $\Omega=\{1, 2, \cdots, n\}$, and the utility function is a simple linear function $g(\mathcal{S}) = |\mathcal{S}|$ where $|\mathcal{S}|$ denotes the size of $\mathcal{S}$. All questions have answer-through-rate one. The (aggregated) continuation probability of question $1$ is $c_1 = 0$, the rest of the questions have continuation probabilities one, e.g, $\forall i \in \Omega\setminus \{1\}, c_i=1$. Moreover, the capacity constraint is $b = n$. For the {\it Random} policy, there is a positive probability that it displays question $1$ first. In this case, the utility of {\it Random} policy is one, this is because question $1$ has zero continuation probability. As the optimal solution  places question $1$ at the last slot, it gains utility $n$, e.g., all questions will be answered under the optimal solution. Thus, the worst-case approximation ratio of the random policy is $1/n$ which approaches zero as $n$ becomes large. In fact, we can show that the worst-case approximation ratio of the second benchmark policy ({\it MaxEnt}) is also zero using the same example. Observe that when $g(\mathcal{S}) = |\mathcal{S}|$, all questions are homogeneous in terms of their marginal contribution, it follows that {\it MaxEnt} policy randomly selects a sequence of questions to display to the customer. Thus, there is a positive probability that it places question $1$ at the first slot, which leads to the utility of one.}

In this set of experiments, we use the same parameter setting as in Section \ref{subsec:imp1}. The only difference is that here we set $c^+_q=0.5$. Figure \ref{fig:benchmark} plots the expected utility generated by {\it QSS} and the two benchmark algorithms with respect to changes in the answer-through-rate. Again, the $x$-axis represents the value of $p^+_q$ and the $y$-axis represents the expected utility generated by the algorithms. Each data point is the average value, over $1000$ problem instances, of the expected utility generated by the corresponding algorithm.

%As illustrated in the figure, {\it QSS} outperforms the benchmark algorithms under all test settings. As expected, {\it Random} yields the lowest expected utility amongst all three. The performance of {\it QSS} is substantially better -- between about $10\%$ to $20\%$ -- over the benchmark approach {\it MaxEnt}.
%We also observe that as $p^+_q$ increases, the gap between the expected utility of the {\it QSS} solution and that of the {\it MaxEnt} solution becomes larger. For instance, as shown in Figure \ref{fig:benchmark}(d), when $p^+_q=0.1$, {\it QSS} produces an expected utility of $2.49$ and {\it MaxEnt} yields $2.25$, with a gap of $0.24$. When $p^+_q=0.6$, {\it QSS} produces an expected utility of $3.05$ and {\it MaxEnt} yields $2.56$, the gap increases to $0.49$.

{As illustrated in the figure, {\it QSS} outperforms the benchmark algorithms under all test settings. As expected, {\it Random} yields the lowest expected utility amongst all three. We observe that {\it QSS} gains around $10$-fold over {\it Random}. For example, as shown in Figure \ref{fig:benchmark}(d), the expected utility of {\it QSS} lies between $2.49$ and $3.05$, and that of {\it Random} lies between $0.24$ and $0.31$. This demonstrates that our proposed algorithm is superior in performance to the randomized algorithm. We also observe that the performance of {\it QSS} is substantially better -- between about $10\%$ to $20\%$ -- over the benchmark approach {\it MaxEnt}.
We further observe that as $p^+_q$ increases, the gap between the expected utility of {\it QSS} and that of {\it MaxEnt} becomes larger. For instance, as shown in Figure \ref{fig:benchmark}(d), when $p^+_q=0.1$, {\it QSS} produces an expected utility of $2.49$ and {\it MaxEnt} yields $2.25$, with a gap of $0.24$. When $p^+_q=0.6$, {\it QSS} produces an expected utility of $3.05$ and {\it MaxEnt} yields $2.56$, the gap increases to $0.49$.}

Note that {\it QSS} takes into account the uncertainty of the customer's behaviour, selects and orders the questions in a way that each selected question is added into the sequence if it maximizes the expected marginal entropy. In the contrast, instead of maximizing the expected utility, {\it MaxEnt} is designed with the objective of finding the subset of questions with the maximum entropy, ignoring the uncertainty in a customer's behaviour. Taking the uncertainty into account, the sequence of questions generated by {\it MaxEnt} can yield the maximum entropy only when all the questions are answered. More specifically, {\it MaxEnt} assumes that the customer would not skip any question in the sequence or quit prematurely. This further indicates that the order matters when it comes to maximizing the expected utility of the sequence of selected questions.

\subsection{Comparison with the Optimal}
\label{subsec:imp3}
\subsubsection{Impact of the Answer-through-Rate and the Continue Probability}

{\begin{table*}[ht]\centering
\caption{Performance of {\it QSS} on a Wider Range of Parameters: $p^+_q$ and $c^+_q$}
\begin{tabular}{cccccccccc}
\hline
 & \multicolumn{9}{c}{$c_{q}^+$}\\
\cline{2-10} $p_{q}^+$ & $0.1$ & $0.2$ & $0.3$ & $0.4$ & $0.5$ & $0.6$ & $0.7$ & $0.8$ & $0.9$\\
\hline
\multirow{2}{*}{$0.1$} & $0.9972$ & $0.9969$ & $0.9964$ & $0.9959$ & $0.9948$ & $0.9937$ & $0.9924$ & $0.9912$ & $0.9905$\\
 & $0.9993$ & $0.9991$ & $0.9987$ & $0.9982$ & $0.9976$ & $0.9961$ & $0.9952$ & $0.9937$ & $0.9926$\\
 \hline
\multirow{2}{*}{$0.2$} & $0.9879$ & $0.9875$ & $0.9864$ & $0.9856$ & $0.9847$ & $0.9839$ & $0.9824$ & $0.9812$ & $0.9798$\\
 & $0.9907$ & $0.9905$ & $0.9899$ & $0.9889$ & $0.9878$ & $0.9867$ & $0.9856$ & $0.9845$ & $0.9837$\\
 \hline
\multirow{2}{*}{$0.3$} & $0.9769$ & $0.9757$ & $0.9742$ & $0.9735$ & $0.9724$ & $0.9712$ & $0.9693$ & $0.9675$ & $0.9652$\\
 & $0.9824$ & $0.9819$ & $0.9808$ & $0.9796$ & $0.9791$ & $0.9779$ & $0.9776$ & $0.9763$ & $0.9757$\\
 \hline
\multirow{2}{*}{$0.4$} & $0.9645$ & $0.9637$ & $0.9624$ & $0.9616$ & $0.9598$ & $0.9588$ & $0.9571$ & $0.9559$ & $0.9543$\\
 & $0.9701$ & $0.9692$ & $0.9687$ & $0.9669$ & $0.9656$ & $0.9643$ & $0.9628$ & $0.9612$ & $0.9594$\\
 \hline
\multirow{2}{*}{$0.5$} & $0.9514$ & $0.9509$ & $0.9498$ & $0.9492$ & $0.9475$ & $0.9463$ & $0.9448$ & $0.9426$ & $0.9405$\\
 & $0.9564$ & $0.9559$ & $0.9547$ & $0.9536$ & $0.9521$ & $0.9509$ & $0.9482$ & $0.9467$ & $0.9451$\\
 \hline
\multirow{2}{*}{$0.6$} & $0.9385$ & $0.9378$ & $0.9365$ & $0.9351$ & $0.9335$ & $0.9324$ & $0.9306$ & $0.9289$ & $0.9268$\\
 & $0.9416$ & $0.9411$ & $0.9398$ & $0.9385$ & $0.9367$ & $0.9352$ & $0.9338$ & $0.9316$ & $0.9304$\\
 \hline
\multirow{2}{*}{$0.7$} & $0.9244$ & $0.9229$ & $0.9217$ & $0.9205$ & $0.9186$ & $0.9175$ & $0.9154$ & $0.9138$ & $0.9107$\\
 & $0.9272$ & $0.9263$ & $0.9251$ & $0.9235$ & $0.9219$ & $0.9201$ & $0.9189$ & $0.9172$ & $0.9159$\\
 \hline
\multirow{2}{*}{$0.8$} & $0.9085$ & $0.9076$ & $0.9063$ & $0.9049$ & $0.9038$ & $0.9025$ & $0.9005$ & $0.8983$ & $0.8957$\\
 & $0.9117$ & $0.9104$ & $0.9095$ & $0.9082$ & $0.9069$ & $0.9054$ & $0.9038$ & $0.9016$ & $0.8992$\\
 \hline
\multirow{2}{*}{$0.9$} & $0.8927$ & $0.8915$ & $0.8889$ & $0.8864$ & $0.8835$ & $0.8806$ & $0.8768$ & $0.8732$ & $0.8691$\\
 & $0.8956$ & $0.8942$ & $0.8919$ & $0.8893$ & $0.8865$ & $0.8837$ & $0.8801$ & $0.8769$ & $0.8725$\\
\hline
\end{tabular}
\label{tab:approx_ratio_1}
\end{table*}}

To further assess the performance of our proposed algorithm, we next examine how {\it QSS} performs relative to the optimal policy. We implement the optimal policy by enumerating all the ordered sequences of size $b$, and picking the one with the maximum expected utility. The test bed is generated as follows. We consider nine possible values each for $p^+_q$ and $c^+_q$ ranging from $0.1$ to $0.9$ with a step of $0.1$. For each combination of values of $p^+_q$ and $c^+_q$, $1000$ problem instances are generated. We consider nine possible values each for $p^-_q$ and $c^-_q$ ranging from $0.1$ to $0.9$ with a step of $0.1$. Note that $p^-_q$ only takes values that are feasible with the corresponding value of $p^+_q$, as these two parameters must add up to a value no more than $1$. For example, when $p^+_q=0.8$, we only consider two values $0.1$ and $0.2$ for $p^-_q$ correspondingly. When $p^+_q=0.9$, we only consider $p^-_q=0.1$. Thus, for each combination of $p^+_q$ and $c^+_q$, we have at least $1000$ instances, for a total of over $1000\times81=81000$ instances in the test bed.

%Table \ref{tab:approx_ratio_1} summarizes the performance of {\it QSS} on this test bed. The value in each cell of the table is the average ratio, over the corresponding thousands of instances, of the expected utility generated by {\it QSS} to the optimal expected utility. As can be seen from the table, the performance of {\it QSS} is excellent, with the actual expected utility exceeding $87\%$ of the optimum. This demonstrates that the actual performance of {\it QSS} is much better than the worst case performance bound $\frac{1}{4}(1-1/e-\epsilon)$ as shown in Corollary \ref{col:1}.
%
%To explain the efficiency of {\it QSS} in compare with the optimal policy, we record the computation time of the two on the above test bed. All experiments were run on a machine with Intel Xeon 2.40GHz CPU and 64GB memory, running 64-bit RedHat Linux server. For {\it QSS}, it generates a solution within 15 seconds on every instance in our test bed. For the optimal policy, it takes over $2$ hours to generate a solution with larger $p^+_q$ and $c^+_q$. For instance, when $p^+_q=0.8$ and $c^+_q=0.6$, it takes the optimal policy $128$ minutes to finish; when $p^+_q=0.9$ and $c^+_q=0.4$, it takes the optimal policy $136$ minutes to finish. Therefore, {\it QSS} is much more efficient than the optimal policy. The reason is that the latter has to enumerate all the ordered sequences of questions of size $b$, and for each sequence, run thousands of Monte Carlo simulations to simulate the customer's behaviour to estimate its expected utility.

{Table \ref{tab:approx_ratio_1} summarizes the performance of {\it QSS} on this test bed. We measure the performance in terms of the ratio of the expected utility generated by {\it QSS} to the optimal expected utility. Each cell of the table corresponds to a combination of values of $p^+_q$ and $c^+_q$. In each cell we report the range of the ratio observed across all feasible settings of $p^-_q$ and $c^-_q$. For example, in the cell for $p^+_q=0.1$ and $c^+_q=0.1$, the observed ratio lies between $99.72\%$ and $99.93\%$. As can be seen from the table, the performance of {\it QSS} is excellent, with the actual expected utility exceeding $86.9\%$ of the optimum. This demonstrates that the actual performance of {\it QSS} is much better than the worst case performance bound $\frac{1}{4}(1-1/e-\epsilon)$ as shown in Corollary \ref{col:1}.}

{To explain the efficiency of {\it QSS} in compare with the enumeration based approach used to find the optimal solution, we record the computation time of the two on the above test bed. All experiments were run on a machine with Intel Xeon 2.40GHz CPU and 64GB memory, running 64-bit RedHat Linux server. For {\it QSS}, it generates a solution within $15$ seconds on every instance in our test bed. For the enumeration based approach, it takes over $2$ hours on average to generate a solution in our test bed.  To further demonstrate the superiority of {\it QSS} in efficiency, we test the case of selecting $9$ out of $12$ quiz questions. Again, {\it QSS} returns a solution within $15$ seconds. The enumeration based approach, however, does not return a solution after one week. We then move on to test the case of selecting $50$ out of $200$ quiz questions. We observe that {\it QSS} is able to return a solution within one minute. Therefore, {\it QSS} is superior in computational efficiency to the enumeration based approach. The reason is that the latter has to enumerate all the ordered sequences of questions of size $b$, and for each sequence, run thousands of Monte Carlo simulations to simulate the customer's behaviour to estimate its expected utility.}

\begin{figure*}[hptb]
\begin{center}
\includegraphics[scale=0.18]{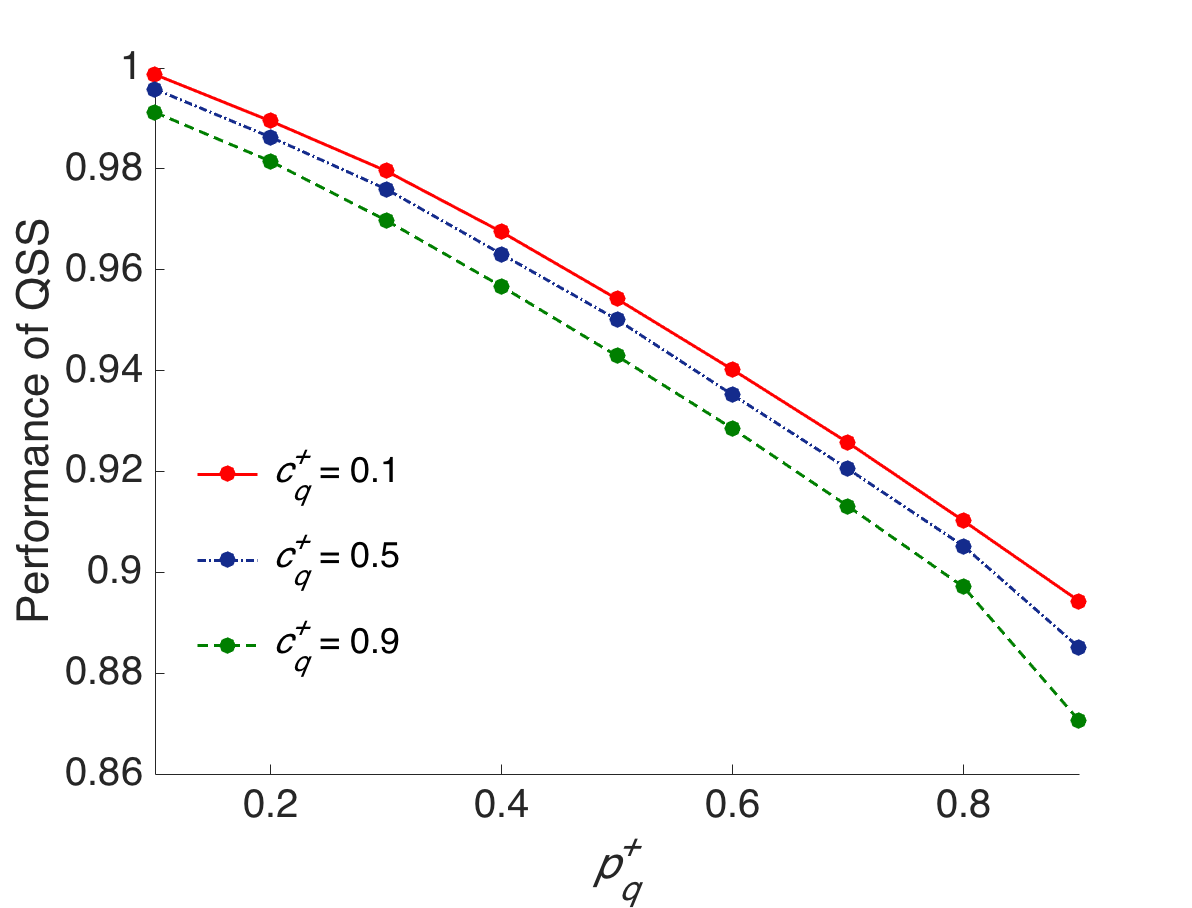}
%\hskip -0.1in
\includegraphics[scale=0.18]{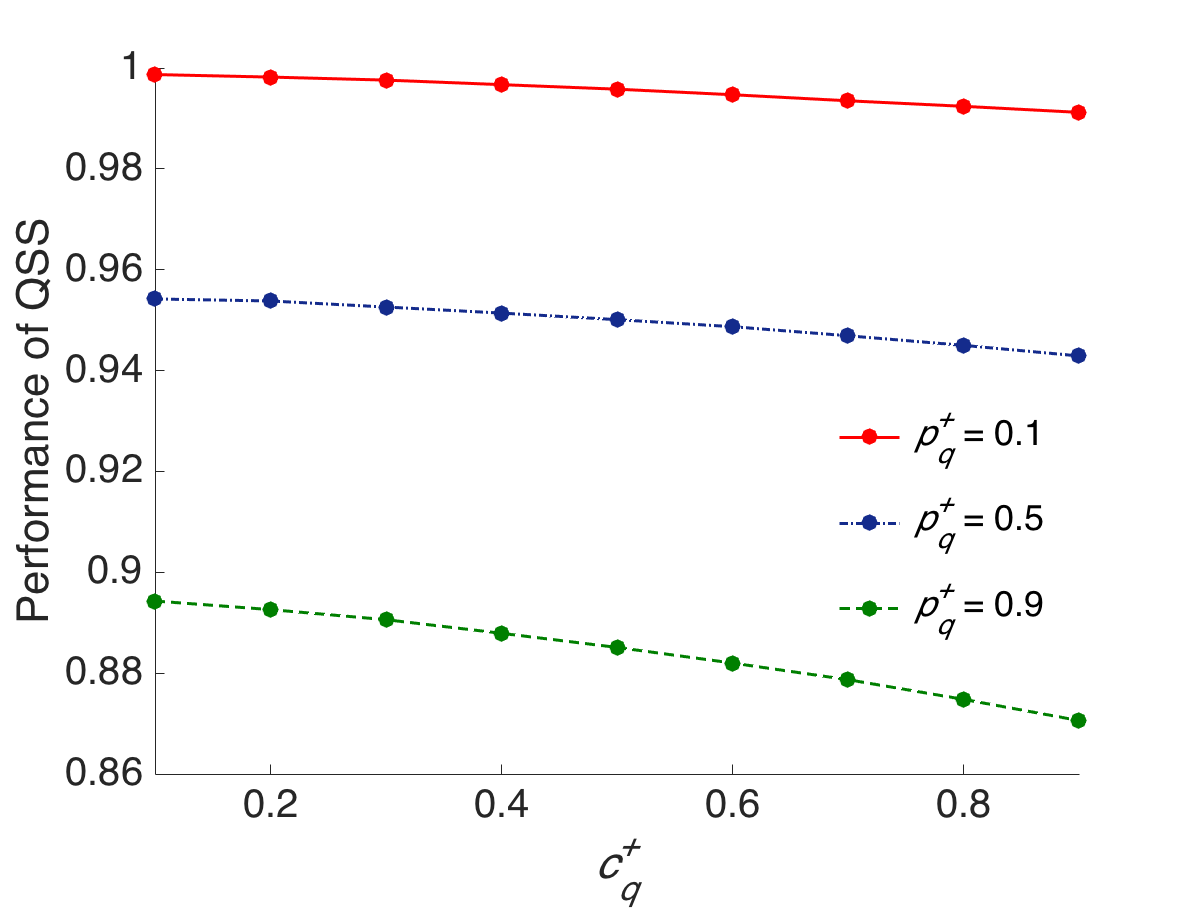}
\caption{Change in the Performance of {\it QSS} with Respect to Changes in the Answer-through-rate $p^+_q$ and the Continue Probability $c^+_q$}
\label{fig:approx_ratio_1}
\end{center}
\end{figure*}

The two plots in Figure \ref{fig:approx_ratio_1} allow us to make two finer observations on the performance of {\it QSS} with respect to changes in the answer-through-rate $p^+_q$ and continue probability $c^+_q$, respectively: (1) For a given value of $c^+_q$, the performance of {\it QSS} improves as $p^+_q$ decreases. (2) For a given value of $p^+_q$, the performance of {\it QSS} improves as $c^+_q$ decreases. Note that in both plots the y-axes shows the average ratio of the expected utility generated by {\it QSS} to the optimal expected utility.

\subsubsection{Impact of the Click-Rate and Continue Probability of the PNA Option}

In this set of experiments, our test bed considers nine possible values each for $p^-_q$ and $c^-_q$ ranging from $0.1$ to $0.9$ with a step of $0.1$. For each combination of values of $p^-_q$ and $c^-_q$, $1000$ problem instances are generated. We consider nine possible values each for $p^+_q$ and $c^+_q$ ranging from $0.1$ to $0.9$ with a step of $0.1$. Note that $p^+_q$ only takes values that are feasible with the corresponding value of $p^-_q$, as these two parameters must add up to a value no more than $1$. For instance, when $p^-_q=0.9$, we only consider $p^+_q=0.1$. Thus, for each combination of $p^-_q$ and $c^-_q$, we have at least $1000$ instances, for a total of over $1000\times81=81000$ instances in the test bed.

%Table \ref{tab:approx_ratio_2} summarizes the performance of {\it QSS} on this test bed. The value in each cell of the table is the average ratio, over the corresponding thousands of instances, of the expected utility generated by {\it QSS} to the optimal expected utility. As can be seen from the table, the performance of {\it QSS} is excellent, with the actual expected utility exceeding $93\%$ of the optimum. This again demonstrates that {\it QSS} performs very well in practice.

{Table \ref{tab:approx_ratio_2} (moved to online appendix)  summarizes the performance of {\it QSS} on this test bed. As aforementioned, We measure the performance in terms of the ratio of the expected utility generated by {\it QSS} to the optimal expected utility. Each cell of the table corresponds to a combination of values of $p^-_q$ and $c^-_q$. In each cell we report the range of the ratio observed across all feasible settings of $p^+_q$ and $c^+_q$. For example, in the cell for $p^-_q=0.1$ and $c^-_q=0.1$, the observed ratio lies between $87.25\%$ and $99.78\%$. As can be seen from the table, the performance of {\it QSS} is excellent, with the actual expected utility exceeding $86.9\%$ of the optimum. This again demonstrates that {\it QSS} performs very well in practice.}

\begin{figure*}[hptb]
\begin{center}
\includegraphics[scale=0.18]{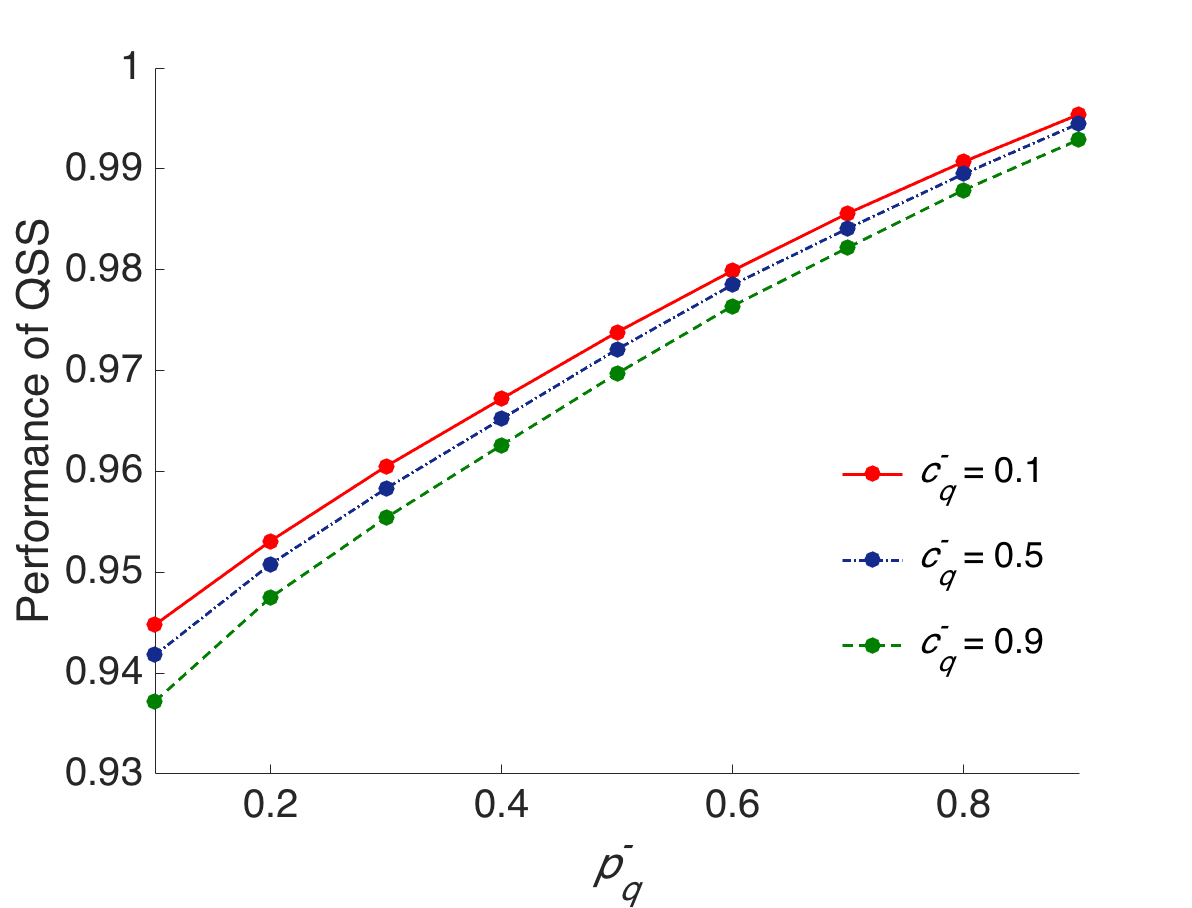}
%\hskip -0.1in
\includegraphics[scale=0.18]{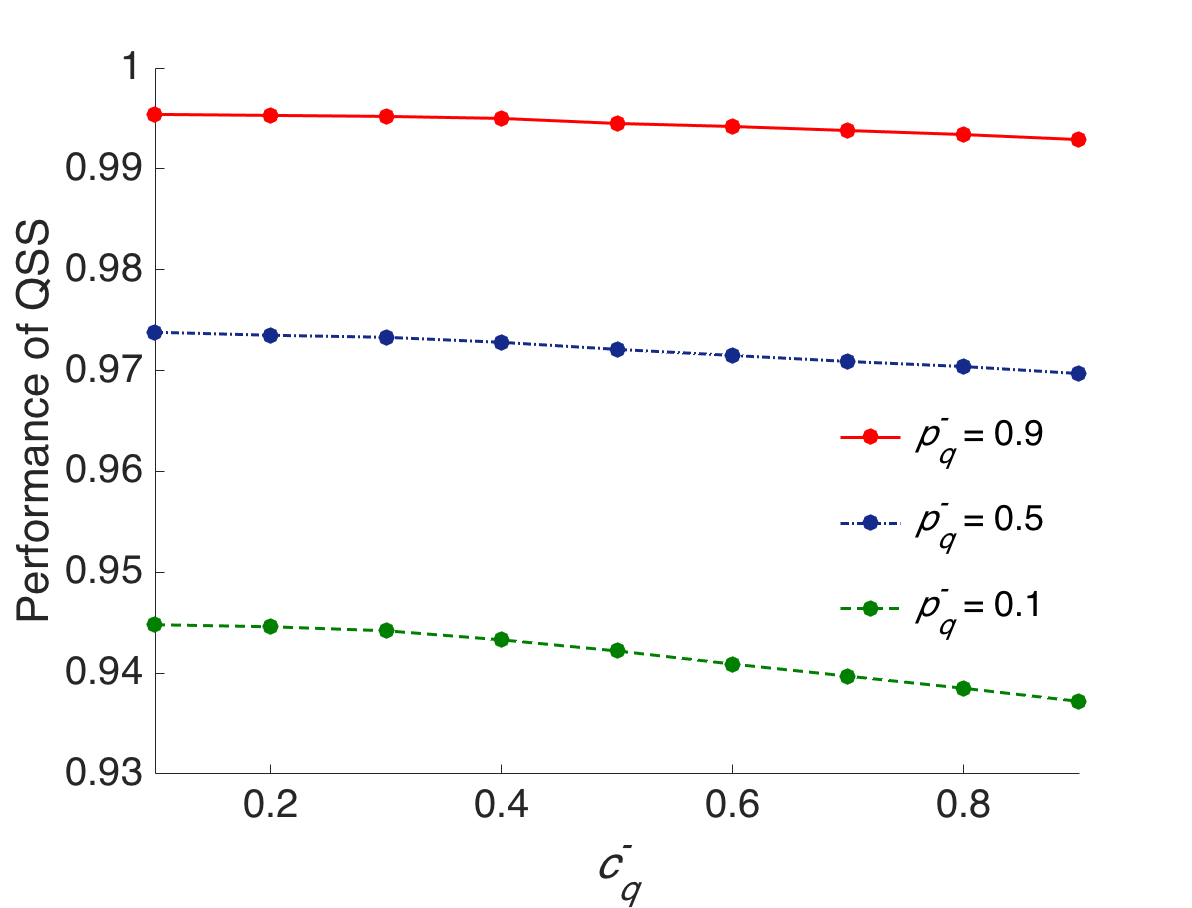}
\caption{Change in the Performance of {\it QSS} with Respect to Changes in the Probability of Choosing the PNA Option $p^-_q$ and the Continue Probability $c^-_q$}
\label{fig:approx_ratio_2}
\end{center}
\end{figure*}

The two plots in Figure \ref{fig:approx_ratio_2} allow us to make two finer observations on the performance of {\it QSS} with respect to changes in the click-rate of PNA $p^-_q$ and continue probability of PNA $c^-_q$, respectively: (1) For a given value of $c^-_q$, the performance of {\it QSS} improves as $p^-_q$ increases. (2) For a given value of $p^-_q$, the performance of {\it QSS} improves as $c^-_q$ decreases.

\subsection{Impact of the PNA Option}
\label{subsec:imp4}

Next we explore the impact of the PNA option on the quality of the solutions. In particular, we aim to figure out  how the expected utility changes before and after PNA is being offered. When PNA is not being offered, customers cannot skip a question before answering the next. Such a forced answering setting offers the potential benefit of virtually eliminating item non-response error, but at the cost of frustrating customers when they are not able to find a response option that reflects the view they want to express. For some customers, without a PNA option, they tend to take each question more seriously and provide a response that they may not do otherwise. For some other customers, not offering PNA can lead to frustration and higher chance of quitting the quiz prematurely. To evaluate both cases, we exploit a parameter, $\kappa \in [-1, +1]$, to measure the extent to which not adding the PNA option affects the answer-through-rate. Specifically, assume the answer-through-rate of question $Q[i]$ is $p^+_{Q[i]}$ when PNA option is being offered, we define the answer-through-rate of question $Q[i]$ when PNA option is not being offered as follows:
\begin{equation*}\mbox{$Q[i]$'s answer-through-rate when PNA is not being offered =}
\begin{cases}
%$\forall \theta_{\mathcal{S}}> 0: |\mathcal{S}|\leq 1 $\\
(1-p^+_{Q[i]})\kappa+p^+_{Q[i]}  \quad\kappa\in(0,1],\\
% \sum_{v\in V}\sum_{d\in D} y_{vd}\leq K \quad(C2)\\
% |\mathcal{S}|\leq b-1 \quad\mbox{(C2)}\\
(1+\kappa)p^+_{Q[i]}  \quad\kappa\in[-1,0]\\
\end{cases}
\end{equation*}

Now we summarize the customer's scanning process under different values of $\kappa$. In the first case, when not using PNA leads to a higher answer-through-rate, i.e., $\kappa>0$, the customer chooses one of the following three actions to take after reading $Q[i]$: (1) answer $Q[i]$ and continue to read the next question with probability $((1-p^+_{Q[i]})\kappa+p^+_{Q[i]})c^+_{Q[i]}$; (2) answer $Q[i]$ and exit the quiz with probability $((1-p^+_{Q[i]})\kappa+p^+_{Q[i]})(1-c^+_{Q[i]})$; (3) exit the quiz with probability $1-((1-p^+_{Q[i]})\kappa+p^+_{Q[i]})$ without answering $Q[i]$. Note that here the value of $\kappa$ indicates how close the updated answer-through-rate $(1-p^+_{Q[i]})\kappa+p^+_{Q[i]}$ is to $1$. In the second case, when not using PNA leads to a lower answer-through-rate, i.e., $\kappa \leq 0$, the customer chooses one of the following three actions to take after reading $Q[i]$: (1) answer $Q[i]$ and continue to read the next question with probability $(1+\kappa)p^+_{Q[i]}c^+_{Q[i]}$; (2) answer $Q[i]$ and exit the quiz with probability $(1+\kappa)p^+_{Q[i]}(1-c^+_{Q[i]})$; (3) exit the quiz with probability $1-(1+\kappa)p^+_{Q[i]}$ without answering $Q[i]$. Note that here the value of $\kappa$ indicates how close the updated answer-through-rate $(1+\kappa)p^+_{Q[i]}$ is to $0$. In this set of experiments, we consider $11$ values of $\kappa$ ranging from $-0.9$ to $0.9$.

Table \ref{tab:pna_no_pna_1}-\ref{tab:pna_no_pna_4} (moved to online appendix) summarizes the comparison of the optimal expected utility with and without PNA, with respect to changes in the parameter $\kappa$. The optimal expected utility is obtained using the enumeration approach as described in Section \ref{subsec:imp3}. We present the comparison under four different settings of $p^+_q$, $c^+_q$, $p^-_q$ and $c^-_q$ in four tables, respectively. For each setting, we measure the impact of removing the PNA option by providing the optimal expected utility with and without PNA and the gaps between the two. In particular, we report the reduction of the optimal expected utility and the percentage of the reduction with respect to $\kappa$. Note that the value of reduction can be negative, in which case the optimal expected utility is improved instead of reduced.

Surprisingly, we find out that the impact of removing PNA on the quality of the solution may be different from its impact on the answer-through-rate. For instance, in Table \ref{tab:pna_no_pna_2}, when $\kappa=0.1$ the optimal expected utility is reduced by $0.982\%$ after removing PNA. Since $\kappa>0$, we know that the answer-through-rate is improved in this case. Thus, when removing PNA improves the answer-through-rate ($\kappa>0$), it does not necessarily mean that it has a positive impact on the quality of the solution. The underlying reason is that when $\kappa$ is a small positive number, while the answer-through-rate ($p_1$) slightly increases as PNA is removed, the probability that the customer directly exit the quiz ($p_2$) may also increases. Here we rename the probabilities for ease of explanation. Denote the original probability that the customer chooses the PNA option by $p_3$. Intuitively, it can be viewed as once PNA is removed, $p_1$ and $p_2$ each shares a portion of $p_3$ that is reduced to $0$. It depends on the value of $p_3$ and how the value of $p_3$ is distributed between $p_1$ and $p_2$, the optimal expected utility may be improved or reduced. For example, when $p^+_q=0.4$, $p^-_q=0.5$ and $\kappa=0.1$, the updated answer-through-rate is $(1-0.4)\times 0.1+0.4=0.46$. With PNA, $p_2=1-0.4-0.5=0.1$; without PNA, it is updated to $1-0.46=0.54$. Note that in this case, the increment of $p_2$ is more significant, resulting in a reduction of the quality of the solution. In this case, we should keep the PNA option, although removing it will improve the answer-through-rate.

We also observe that when PNA is removed, the optimal expected utility is first reduced, and then improved as $\kappa$ increases. Moreover, the percentage of the reduction decreases as $\kappa$ increases. As $\kappa$ continues to increase, the percentage of the increment also increases. Therefore, the decision on whether to offer PNA option or not  is crucially depending on the value of $\kappa$. When $\kappa$ is small, removing the PNA option reduces the optimal expected utility, thus it has a negative impact on the quality of the solution. In this case, the PNA option should be kept. When $\kappa$ is large, removing the PNA option leads to an increment of the optimal expected utility. Thus, it has a positive impact on the quality of the solution. In this case, the PNA option should be removed.

%We are now ready to summarize the insights gained from our computational study. 1. We evaluate the quality of our solution by examining the expected utility generated under a wide range of parameters. 2. We validate the performance of our solution through a comparison to the benchmark solutions and to the optimal solution. Our solution outperforms the benchmarks under all test settings. Corollary \ref{col:1} guarantees a worst case lower bound on the approximation ratio of our solution. In our experiment, we show that the actual performance of our solution is excellent, with the expected utility exceeding $87\%$ of the optimum. This demonstrates that the bound is simply a worst case bound. Moreover, our solution can solve the problem in a few seconds instead of a few hours that the optimal policy has to take in certain cases. 3. We also evaluate how the expected utility would change before and after PNA is being offered. We introduce a parameter $\kappa$ to measure the extent to which removing the PNA option affects the answer-through-rate. We find out that when removing PNA improves the answer-through-rate ($\kappa>0$), it does not necessarily mean that removing PNA has a positive impact on the quality of the solution. For example, it is beneficial to offer the PNA option when $\kappa$ is positive but small. When $\kappa$ is large enough, removing PNA is more beneficial. Therefore, the decision on whether to offer the PNA option or not is crucially depending on the value of $\kappa$.

\section{Extensions}
\label{sec:5}
We now discuss three extensions. Our first extension captures the situation that the answer-through-rate of a question is position dependent, the second extension incorporates the PNA option as a decision variable, and the last extension extends our study to the scrolling design. Due to the space limitation, we move the latter two extensions to the online appendix.
\subsection{Extension 1: Incorporating Slot-Dependent Decay Factor}
\label{sec:extended}
In this extension, we take into account the slot-dependent decay factor, e.g., the answer-through-rate of a question could be influenced by its position.  In the extended model, each slot $i$ has slot-dependent decay factor $\lambda_i \leq 1$. Given  a sequence of questions $Q$, the probability that the $i$-th question  $Q[i]$ is answered, conditioned on it has been read, is $\lambda_i p^+_{Q[i]}$. We assume that $\forall i\leq j: \lambda_i\geq \lambda_j$, i.e.,
one would typically expect the answer-through-rate to decrease with slot. For ease of presentation, we assume $\lambda_1=1$, e.g., slot-dependent decay effect does not apply to the first slot.
  \begin{center}
\framebox[0.9\textwidth][c]{
\enspace
\begin{minipage}[t]{0.85\textwidth}
\small
\begin{itemize}
\item Starting with the first slot $i=1$.
\item After reading $Q[i]$, the customer chooses one of the following five actions to take:
 \begin{enumerate}
 \item Answer $Q[i]$  and continue to read the next question (resp. exit the quiz) with probability   $\lambda_i p^+_{Q[i]}c_{Q[i]}^+$ (resp. $\lambda_i p^+_{Q[i]}(1- c_{Q[i]}^+)$).
  \item PNA $Q[i]$ and continue to read the next question (resp. exit the quiz) with probability   $p^-_{Q[i]} c_{Q[i]}^-$ (resp. $p^-_{Q[i]} (1-c_{Q[i]}^-)$).
  \item Exit the quiz with probability $1-(\lambda_i p^+_{Q[i]}+p^-_{Q[i]})$.
  \end{enumerate}
% \item After answering (resp. skipping) the $Q[i]$, the customer continues to read the next question with probability  $c_{Q[i]}^+$ (resp.  $c_{Q[i]}^-$ ); otherwise, exit the quiz.
 \item The above process repeats until the customer exits the quiz or no more questions remain.
\end{itemize}
\end{minipage}
}
\end{center}
\vspace{0.1in}

We revise the definition of reachability to incorporate slot-dependent decay effect.

\begin{definition}[Reachability of a Question]
Given  a sequence of questions $Q$, for each $i\in\{1, 2, \cdots, |Q|\}$, the reachability $C_{Q[i]}$ of  the $i$-th question $Q[i]$ can be written as:
%\[C_{q}= \prod_{q'\in Q_{<i}}c_{q'}\]
%  For any given sequence of questions $Q$, the reachability of $Q[i]$ is:
$C_{Q[i]}= \prod_{j\in \{1, 2, \cdots, i-1\}} (\lambda_j p^+_{Q[j]}c_{Q[j]}^++ p^-_{Q[j]} c_{Q[j]}^-)$.
%where $Q_{\overleftarrow{q}}$ denotes the subsequence of questions in $Q$ which is displayed before $q$.
\end{definition}

\subsubsection{Question Selection and Sequencing with No PNA option}
We first study the case when PNA is not an option. A simplified question scanning process is presented as follows.

  \begin{center}
\framebox[0.9\textwidth][c]{
\enspace
\begin{minipage}[t]{0.85\textwidth}
\small

\begin{itemize}
\item Starting with the first slot $i=1$.
\item After reading $Q[i]$, the customer chooses one of the following five actions to take:
 \begin{enumerate}
 \item Answer $Q[i]$  and continue to read the next question (resp. exit the quiz) with probability   $\lambda_i p^+_{Q[i]}c_{Q[i]}^+$ (resp. $\lambda_i p^+_{Q[i]}(1- c_{Q[i]}^+)$).
 % \item PNA $Q[i]$ and continue to read the next question (resp. exit the quiz) with probability   $p^-_{Q[i]} c_{Q[i]}^-$ (resp. $p^-_{Q[i]} (1-c_{Q[i]}^-)$).
  \item Exit the quiz with probability $1-\lambda_i p^+_{Q[i]}$.
  \end{enumerate}
% \item After answering (resp. skipping) the $Q[i]$, the customer continues to read the next question with probability  $c_{Q[i]}^+$ (resp.  $c_{Q[i]}^-$ ); otherwise, exit the quiz.
 \item The above process repeats until the customer exits the quiz or no more questions remain.
\end{itemize}
\end{minipage}
}
\end{center}
\vspace{0.1in}

  \begin{center}
\framebox[0.46\textwidth][c]{
\enspace
\begin{minipage}[t]{0.46\textwidth}
\small
$\textbf{P2.1}$
\emph{Maximize$_{t, q, \mathcal{S}}$ $u(t, q, \mathcal{S})$}\\
\textbf{subject to:}
\begin{equation*}
\begin{cases}
%$\forall \theta_{\mathcal{S}}> 0: |\mathcal{S}|\leq 1 $\\
-(\log \Lambda_t +\sum_{l\in \mathcal{S}} \log (p^+_{l}c_{l}^+)) \leq -\log \rho\quad\mbox{(C1.1)}\\
% \sum_{v\in V}\sum_{d\in D} y_{vd}\leq K \quad(C2)\\
% |\mathcal{S}|\leq b-1 \quad\mbox{(C2)}\\
|\mathcal{S}| < t \quad \mbox{(C2.1)}\\
\mathcal{S} \subseteq \Omega \setminus \{q\}\\
0\leq t \leq b
\end{cases}
\end{equation*}
\end{minipage}
}
\end{center}
\vspace{0.1in}
Define $\Lambda_i=\prod_{1\leq j \leq i } \lambda_j$. By setting $p^-_{q}=0$ for every $q\in \Omega$, we derive a simplified form of $C_{Q[i]}$ as follows: $C_{Q[i]}= \prod_{1\leq j< i}(\lambda_j p^+_{Q[j]}c_{Q[j]}^+)=\Lambda_i \prod_{1\leq j< i}p^+_{Q[j]}c_{Q[j]}^+$. We first introduce a new problem \textbf{P2.1}. The formulation of \textbf{P2.1} is similar to \textbf{P2} except that there is one additional decision variable $t$, which specifies the index of the last slot occupied by our solution. The reason why we introduce this additional decision variable is because $\lambda_i$ is slot-dependent, by fixing the index of the last slot enables us to separate the slot-dependent decay effect from other question-dependent factors such as answer-through-rate and continuation probability. The basic idea of our solution is similar to Algorithm \ref{alg:greedy-peak}, after solving \textbf{P2.1} and obtain a solution $(t', q', \mathcal{S}')$, we build the final solution to the original problem based on $(t', q', \mathcal{S}')$.

We next give a detailed description of \textbf{P2.1}. The objective function of  \textbf{P2.1} is
\[u(t, q, \mathcal{S})=  \lambda_t p_q^+g(\mathcal{S}\cup\{q\})+(1-\lambda_t p_q^+)g(\mathcal{S})\]

Constraint (C1.1) ensures that the reachability of every question, after taking into account the slot-dependent decay effect $\Lambda_i$, is no smaller than $\rho$. Constraint (C2.1) ensures that our solution occupies  up to $t$ slots. Note that for any fixed $t$ and $q$, $\textbf{P2.1}$ is a monotone submodular maximization problem subject to two linear constraints. In order to solve \textbf{P2.1}, we exhaustedly try all possibilities of  $t$ and $q$. For each $t$ and $q$, we run the $(1-1/e-\epsilon)$ approximate algorithm to obtain a candidate solution $(t, q, \mathcal{S})$.   Among all candidate solutions, assume  $(t', q', \mathcal{S}')$ has the largest utility, $Q^{\mathrm{Alg3}}= S' \oplus q'$  is returned as the final solution to the original problem.
%Because the answer-through-rate of each question is dependent on its position, we use $f(i, Q)$ to denote the \emph{conditional} utility of $Q$,  given that it has been reached, with its first question scheduled at slot $i$. We use $f(Q)$ to denote $f(1, Q)$ for short.
We present the detailed description of our solution in Algorithm \ref{alg:greedy-peak3}.
\begin{algorithm}[h]
{\small
\caption{Question Selection and Sequencing with No PNA option}
\label{alg:greedy-peak3}
\textbf{Input:} $\rho, b, \Omega$.\\
\textbf{Output:} $Q^{\mathrm{Alg3}}$.
\begin{algorithmic}[1]
\STATE Set $\mathcal{S}'=\emptyset, q'=\emptyset, t'=0$.
\FOR{$t\in [1, b]$}
\FOR{$q\in \Omega$}
\STATE Fix $t$ and $q$, apply a $(1-1/e-\epsilon)$ approximate algorithm \citep{kulik2009maximizing} to solve $\textbf{P2.1}$  and obtain $\mathcal{S}$ \label{line:2}
\IF {$u(t, q, \mathcal{\mathcal{S}})> u(t', q', \mathcal{S}')$}
\STATE $t'\leftarrow t, \mathcal{S}' \leftarrow \mathcal{S}, q'\leftarrow q$
\ENDIF
\ENDFOR
\ENDFOR
%\STATE $\mathcal{Q}_{\mathrm{Alg1}}\leftarrow \mathcal{S}'\cup\{q'\}$
\STATE $Q^{\mathrm{Alg3}}\leftarrow S' \oplus \{q'\}$ where $ S'$ is an arbitrary sequence of $\mathcal{S}'$ \label{line:1}
\RETURN $Q^{\mathrm{Alg3}}$
\end{algorithmic}
}
\end{algorithm}
To provide a performance bound to our solution, we first present four preparatory lemmas as follows.

\begin{lemma}
\label{lem:767676}
Given any sequence $Q$, consider any question $Q[i]\in Q$, $f(Q)$ is a non-decreasing function of $\lambda_i$ and $C_{Q[i]}$.
\end{lemma}

Similar to Lemma \ref{lem:1}, we can show that ignoring those questions with small reachability does not affect the utility much.
\begin{lemma}
\label{lem:22}
For any $\rho\in[0,1]$, there is a solution $Q$ of value at least $(1-\rho)f(Q^*)$ such that %$Q_\rho \overset{\mathrm{subsequence}}{\preceq} Q^*$ and
$|Q|\leq b$ and $\forall i\in \{1, 2, \cdots, |Q|\}: C_{Q[i]} \geq \rho$.
\end{lemma}

\begin{lemma}
\label{lem:333}
$(t', q', \mathcal{S}')$ is a $(1-1/e-\epsilon)$ approximate solution to $\textbf{P2.1}$.
\end{lemma}

\begin{lemma}
\label{lem:44444}
For any $\rho\in[0,1]$, $f(Q^{\mathrm{Alg3}})\geq \rho u(t', q', \mathcal{S}')$.
\end{lemma}

Lemma \ref{lem:22}, Lemma \ref{lem:333}, and Lemma \ref{lem:44444} together imply the following main theorem.
\begin{theorem}
For any $\rho\in[0,1]$, $f(Q^{\mathrm{Alg3}})\geq \rho (1-\rho)(1-1/e-\epsilon) f(Q^*)$.
\end{theorem}

\begin{corollary}
By choosing $\rho=1/2$, we have $f(Q^{\mathrm{Alg3}})\geq \frac{1}{4}(1-1/e-\epsilon) f(Q^*)$.
\end{corollary}
\subsubsection{Question Selection and Sequencing under General Model}
We next study this extended problem under general model where PNA is an option. The basic idea of our approach is to covert the original joint  selection and sequencing problem to a simplified selection problem. For each question $q$, we create $b$ copies of virtual questions $\Omega^\nu_q=\{q^1, \cdots, q^b\}$. Let $\Omega^\nu= \bigcup_{q\in \Omega} \Omega^\nu_q$ denote the expanded ground set that is composed of virtual questions. We next focus on selecting a group of virtual questions. Intuitively, selecting a virtual question $q^i$ translates to placing $q$ at slot $i$.

 We next introduce some important notations. %The answer-through-rate of $q^i$ is $p_{q^i}=p_q\lambda_i$, the continuation probability of $p_{q^i}$ is $c^+_{q^i}=c^+_{q}$ and $c^-_{q^i}=c^-_{q}$.
 Define $\Omega^\nu_i=\{q^i| q\in \Omega\}$.  For every $q^i$, let $c_{q^i}=\lambda_i p^+_q c^+_{q}+p^-_q c^-_{q}$. Given a set of virtual questions $\mathcal{S}^\nu \subseteq \Omega^\nu$, %define $\mathcal{S}^\nu_{\min}=\{q^i| q^i\in \Omega^\nu_q \cap \mathcal{S}^\nu; i=\min\{j| q^j\in \Omega^\nu_q \cap \mathcal{S}^\nu\}\}$. For example, given  $\mathcal{S}^\nu=\{q_a^2, q_a^4, q_b^3\}$, we have $\mathcal{S}^\nu_{\min} =\{q_a^2, q_b^3\}$.
 let $\mathcal{S}=\{q|\Omega^\nu_q \cap \mathcal{S}^\nu\neq \emptyset\}$, we use $\mathcal{R}(\mathcal{S}^\nu)$ to denote a random set obtained by including each question $q\in \mathcal{S}$  with probability $\lambda_i p^+_q$ where $i=\min\{j| q^j\in \Omega^\nu_q \cap \mathcal{S}^\nu\}$. %Then we are ready to define the utility function $g$ over $\Omega^\nu$: $g(\mathcal{S}^\nu)=g(\mathcal{S})$.

We next introduce problem $\mathbf{P3.1}$ whose objective function is
$v(t, q, \mathcal{S}^\nu)= \mathbb{E}[g(\mathcal{R} (\mathcal{S}^\nu\cup\{q^t\}))]$. % where $\mathcal{R} (\mathcal{S}^\nu\cup\{q^t\})$ is (redefined as) a random set obtained by including each virtual question $q^i\in \mathcal{S}^\nu\cup\{q^t\}$ with probability $\lambda_i p^+_q$.
The goal of $\mathbf{P3.1}$ is to find a solution that maximizes function $v$. After solving $\mathbf{P3.1}$ approximately and obtain $(t', {q}', {\mathcal{S}^\nu}')$, we build the final solution based on $(t', {q}', {\mathcal{S}^\nu}')$.

 \begin{center}
\framebox[0.46\textwidth][c]{
\enspace
\begin{minipage}[t]{0.46\textwidth}
\small
$\textbf{P3.1}$
\emph{Maximize$_{t, q, \mathcal{S}^\nu}$ $v(t, q, \mathcal{S}^\nu)$}\\
\textbf{subject to:}
\begin{equation*}
\begin{cases}
%$\forall \theta_{\mathcal{S}}> 0: |\mathcal{S}|\leq 1 $\\
-\sum_{q^i\in \mathcal{S}^\nu} \log c_{q^i} \leq -\log \rho\\%\quad\mbox{(C3.1)}\\
% \sum_{v\in V}\sum_{d\in D} y_{vd}\leq K \quad(C2)\\
% |\mathcal{S}|\leq b-1 \quad\mbox{(C2)}\\
|\mathcal{S}^\nu| < t \quad \\
\forall 1 \leq i \leq b: |\mathcal{S}^\nu \cap \Omega^\nu_i|\leq 1 \mbox{ (C4.1)}\\
\mathcal{S}^\nu \subseteq \cup_{1\leq i<t}\Omega^\nu_i\\
0 \leq t < b
\end{cases}
\end{equation*}
\end{minipage}
}
\end{center}
\vspace{0.1in}

The formulation of $\textbf{P3.1}$ is similar to $\textbf{P3}$, except that now we are dealing with virtual questions. We next explain how to covert a solution to $\textbf{P3.1}$ to a solution to the original problem: given a solution $(t, q, \mathcal{S}^\nu)$ to  $\textbf{P3.1}$, we place $q$ at slot $i$ if and only if $q^i\in \mathcal{S}^\nu\cup q^t$. For example, $(4, q_d^4, \{q_a^1, q_b^3, q_c^2\})$ translates to placing $q_a$ (resp. $q_b, q_c, q_d$) at the first (resp. third, second, forth) slot.   To ensure the feasibility of the solution, we employ condition (C4.1) to avoid assigning multiple questions to the same slot. Similar to Lemma \ref{lem:67}, we can prove that for any fixed $t$ and $q$, $v(t, q, \mathcal{S}^\nu)$ is monotone and submodular as a function of $\mathcal{S}^\nu$. Together with the fact that (C4.1) is a (partition) matroid constraint, we have that for any fixed $t$ and $q$, $\textbf{P3.1}$ is a monotone submodular maximization problem subject to two linear and one matroid constraints. There exists a $0.38$ approximate solution   \citep{vondrak2011submodular} to this problem.

\emph{Remark:} Notice that a feasible solution to $\textbf{P3.1}$ may include multiple copies from the same question, this  redundancy issue can be easily resolved by keeping the one which has the largest answer-through-rate in the solution. This will not affect the utility of our solution due to the definition of $v(t, q, \mathcal{S}^\nu)$. Another potential issue is that our solution may contain some ``gap'', i.e., there is at least one empty slot between two scheduled questions. This gap issue can also be easily resolved by simply removing those gaps from the final solution, e.g., this can be done by moving all questions to its earliest possible slot while respecting their original ordering. Due to Lemma \ref{lem:767676}, moving questions to some earlier slot will not decrease its utility.

We present the detailed description of our solution in Algorithm \ref{alg:greedy-peak4}.
\begin{algorithm}[h]
{\small
\caption{Question Selection and Sequencing with PNA option}
\label{alg:greedy-peak4}
\textbf{Input:} $\rho, b, \Omega$.\\
\textbf{Output:} $Q^{\mathrm{Alg4}}$.
\begin{algorithmic}[1]
\STATE Set ${\mathcal{S}^\nu}'=\emptyset, {q}'=\emptyset, t'=0$.
\FOR{$t\in [1, b]$}
\FOR{$q\in \Omega$}
\STATE Fix $t$ and $q$, apply a $0.38$ approximate algorithm \citep{vondrak2011submodular} to solve $\textbf{P3.1}$  and obtain $\mathcal{S}^\nu$ \label{line:77}
\IF {$u(t, q, {\mathcal{S}^\nu})> u(t', {q}', {\mathcal{S}^\nu}')$}
\STATE $t'\leftarrow t, {\mathcal{S}^\nu}' \leftarrow {\mathcal{S}^\nu}, {q}'\leftarrow q$
\ENDIF
\ENDFOR
\ENDFOR
\FOR{$q^i \in {\mathcal{S}^\nu}'$}
\STATE place $q$ at slot $i$ of $Q^{\mathrm{Alg4}}$
\ENDFOR
\STATE Place $q'$ at slot $t'$ of $Q^{\mathrm{Alg4}}$
%\STATE $Q^{\mathrm{Alg3}}\leftarrow S' \oplus \{{q^t}'\}$ where $ S'$ is the sequence of  actual questions decided by $\mathcal{S}'$ \label{line:1}
\RETURN $Q^{\mathrm{Alg4}}$  \COMMENT{we may need to refine $Q^{\mathrm{Alg4}}$ by removing any redundant questions and gaps.}
\end{algorithmic}
}
\end{algorithm}
In the rest of this paper, we redefine $\mathcal{R}(Q)$ as a random set obtained by including each question $Q[i]\in Q$ with probability $\lambda_ip_{Q[i]}^+$, then the proof of Lemma \ref{lem:777} (resp. Lemma \ref{lem:88888}) is similar to the proof of Lemma \ref{lem:7} (resp. Lemma \ref{lem:9999999}).

{\begin{lemma}
\label{lem:777}
$f(Q^*_{\leq k}) \leq \mathbb{E}[g(\mathcal{R}(Q^*_{\leq k}))]$.
\end{lemma}}
%\proof
%The proof is similar to the proof of Lemma \ref{lem:7}.
%\endproof

\begin{lemma}
\label{lem:88888}
For any $\rho\in(0,1]$, $f(Q^{\mathrm{Alg4}})\geq \rho \mathbb{E}[g(\mathcal{R}(Q^{\mathrm{Alg4}}))]$.
\end{lemma}
We next present the main theorem.

{\begin{theorem}
\label{thm:extend1}For any $\rho\in[0,1]$, $f(Q^{\mathrm{Alg4}})\geq 0.38 \rho(1-\rho)f(Q^*)$.
\end{theorem}

\begin{corollary} By choosing $\rho=1/2$, we have $f(Q^{\mathrm{Alg4}})\geq \frac{0.38}{4}f(Q^*)$.
\end{corollary}
}

{\paragraph{Contribution to assortment optimization.} As one natural extension to the ``consider-then-choose'' model introduced in Section \ref{sec:connection}, we may assume a slot-dependent consider-through-rate for each product in the stage of forming a consideration set. This is because it is very likely that the probability of a product  being added to the consideration set decreases as its ranking drops. In this case, we can define the consider-through-rate of the product $Q[i]$ placed at slot $i$ as $\lambda_i p^+_{Q[i]}$, where $p^+_{Q[i]}$ is the initial consider-through-rate, $\lambda_i$ is the slot-dependent decay factor of slot $i$ and  $\forall i\leq j: \lambda_i\geq \lambda_j$. To solve this extended assortment optimization problem, we can  use Algorithm \ref{alg:greedy-peak4} to find a $\frac{0.38}{4}$ approximation when the revenue function (2) is monotone and submodular.}

\section{Conclusion}
In this paper, we study the optimal quiz design problem. We assume the utility function of a group of answered questions is submodular and our objective is to select and sequence of a group of questions so as to maximize the expected utility. We model the customer behavior as a Markov process. Then we develop a series of question allocation strategies with provable performance bound. Although we restrict our attention to the quiz design problem in this paper, our results apply to a broad range of applications which can be formulated as a submodular maximization problem under the cascade browse model.
\bibliographystyle{ormsv080}
\bibliography{social-advertising-1}

\begin{thebibliography}{65}
\expandafter\ifx\csname natexlab\endcsname\relax\def\natexlab#1{#1}\fi
\expandafter\ifx\csname url\endcsname\relax
  \def\url#1{{\tt #1}}\fi
\expandafter\ifx\csname urlprefix\endcsname\relax\def\urlprefix{URL }\fi
\expandafter\ifx\csname urlstyle\endcsname\relax
  \expandafter\ifx\csname doi\endcsname\relax
  \def\doi#1{doi:\discretionary{}{}{}#1}\fi \else
  \expandafter\ifx\csname doi\endcsname\relax
  \def\doi{doi:\discretionary{}{}{}\begingroup \urlstyle{rm}\Url}\fi \fi

\bibitem[{Abeliuk et~al.(2016)Abeliuk, Berbeglia, Cebrian, and
  Van~Hentenryck}]{abeliuk2016assortment}
Abeliuk, Andr{\'e}s, Gerardo Berbeglia, Manuel Cebrian, Pascal Van~Hentenryck.
  2016.
\newblock Assortment optimization under a multinomial logit model with position
  bias and social influence.
\newblock {\it 4OR\/} {\bf 14}(1) 57--75.

\bibitem[{Alaei et~al.(2010)Alaei, Makhdoumi, and
  Malekian}]{alaei2010maximizing}
Alaei, Saeed, Ali Makhdoumi, Azarakhsh Malekian. 2010.
\newblock Maximizing sequence-submodular functions and its application to
  online advertising.
\newblock {\it arXiv preprint arXiv:1009.4153\/} .

\bibitem[{Ansari and Mela(2003)}]{ansari2003customization}
Ansari, Asim, Carl~F Mela. 2003.
\newblock E-customization.
\newblock {\it Journal of marketing research\/} {\bf 40}(2) 131--145.

\bibitem[{Aouad et~al.(2019)Aouad, Feldman, Segev, and Zhang}]{aouad2019click}
Aouad, Ali, Jacob Feldman, Danny Segev, Dennis Zhang. 2019.
\newblock Click-based mnl: Algorithmic frameworks for modeling click data in
  assortment optimization.
\newblock {\it Available at SSRN 3340620\/} .

\bibitem[{Aouad and Segev(2015)}]{aouad2015display}
Aouad, Ali, Danny Segev. 2015.
\newblock Display optimization for vertically differentiated locations under
  multinomial logit choice preferences.
\newblock {\it Available at SSRN 2709652\/} .

\bibitem[{Asadpour et~al.(2020)Asadpour, Niazadeh, Saberi, and
  Shameli}]{asadpour2020ranking}
Asadpour, Arash, Rad Niazadeh, Amin Saberi, Ali Shameli. 2020.
\newblock Ranking an assortment of products via sequential submodular
  optimization.
\newblock {\it arXiv preprint arXiv:2002.09458\/} .

\bibitem[{Atahan and Sarkar(2011)}]{atahan2011accelerated}
Atahan, Pelin, Sumit Sarkar. 2011.
\newblock Accelerated learning of user profiles.
\newblock {\it Management Science\/} {\bf 57}(2) 215--239.

\bibitem[{Bishop(2006)}]{bishop2006pattern}
Bishop, Christopher~M. 2006.
\newblock Pattern recognition and machine learning (information science and
  statistics) springer-verlag new york.
\newblock {\it Inc. Secaucus, NJ, USA\/} .

\bibitem[{Blanchet et~al.(2016)Blanchet, Gallego, and
  Goyal}]{blanchet2016markov}
Blanchet, Jose, Guillermo Gallego, Vineet Goyal. 2016.
\newblock A markov chain approximation to choice modeling.
\newblock {\it Operations Research\/} {\bf 64}(4) 886--905.

\bibitem[{Boutilier et~al.(2002)Boutilier, Zemel, and
  Marlin}]{boutilier2002active}
Boutilier, Craig, Richard~S Zemel, Benjamin Marlin. 2002.
\newblock Active collaborative filtering.
\newblock {\it Proceedings of the Nineteenth conference on Uncertainty in
  Artificial Intelligence\/}. Morgan Kaufmann Publishers Inc., 98--106.

\bibitem[{Cachon et~al.(2005)Cachon, Terwiesch, and
  Xu}]{doi:10.1287/msom.1050.0088}
Cachon, G{\'e}rard~P., Christian Terwiesch, Yi~Xu. 2005.
\newblock Retail assortment planning in the presence of consumer search.
\newblock {\it Manufacturing \& Service Operations Management\/} {\bf 7}(4)
  330--346.
\newblock \doi{10.1287/msom.1050.0088}.
\newblock \urlprefix\url{https://doi.org/10.1287/msom.1050.0088}.

\bibitem[{Calinescu et~al.(2011)Calinescu, Chekuri, P{\'a}l, and
  Vondr{\'a}k}]{calinescu2011maximizing}
Calinescu, Gruia, Chandra Chekuri, Martin P{\'a}l, Jan Vondr{\'a}k. 2011.
\newblock Maximizing a monotone submodular function subject to a matroid
  constraint.
\newblock {\it SIAM Journal on Computing\/} {\bf 40}(6) 1740--1766.

\bibitem[{Chang et~al.(2015)Chang, Harper, and Terveen}]{chang2015using}
Chang, Shuo, F~Maxwell Harper, Loren Terveen. 2015.
\newblock Using groups of items for preference elicitation in recommender
  systems.
\newblock {\it Proceedings of the 18th ACM Conference on Computer Supported
  Cooperative Work \& Social Computing\/}. ACM, 1258--1269.

\bibitem[{Couper et~al.(2001)Couper, Traugott, and Lamias}]{couper2001web}
Couper, Mick~P, Michael~W Traugott, Mark~J Lamias. 2001.
\newblock Web survey design and administration.
\newblock {\it Public opinion quarterly\/} {\bf 65}(2) 230--253.

\bibitem[{Craswell et~al.(2008)Craswell, Zoeter, Taylor, and
  Ramsey}]{craswell2008experimental}
Craswell, Nick, Onno Zoeter, Michael Taylor, Bill Ramsey. 2008.
\newblock An experimental comparison of click position-bias models.
\newblock {\it Proceedings of the 2008 International Conference on Web Search
  and Data Mining\/}. ACM, 87--94.

\bibitem[{Cremonesi et~al.(2010)Cremonesi, Koren, and
  Turrin}]{cremonesi2010performance}
Cremonesi, Paolo, Yehuda Koren, Roberto Turrin. 2010.
\newblock Performance of recommender algorithms on top-n recommendation tasks.
\newblock {\it Proceedings of the fourth ACM conference on Recommender
  systems\/}. ACM, 39--46.

\bibitem[{Culnan(2001)}]{culnan2001culnan}
Culnan, Mary~J. 2001.
\newblock The culnan-milne survey on consumers \& online privacy notices:
  Summary of responses.
\newblock {\it Interagency public workshop: Get noticed: Effective financial
  privacy notices\/}. 47--54.

\bibitem[{Davis et~al.(2013)Davis, Gallego, and
  Topaloglu}]{davis2013assortment}
Davis, James, Guillermo Gallego, Huseyin Topaloglu. 2013.
\newblock Assortment planning under the multinomial logit model with totally
  unimodular constraint structures.
\newblock {\it Work in Progress\/} .

\bibitem[{Davis et~al.(2014)Davis, Gallego, and
  Topaloglu}]{davis2014assortment}
Davis, James~M, Guillermo Gallego, Huseyin Topaloglu. 2014.
\newblock Assortment optimization under variants of the nested logit model.
\newblock {\it Operations Research\/} {\bf 62}(2) 250--273.

\bibitem[{Davis et~al.(2015)Davis, Topaloglu, and
  Williamson}]{davis2015assortment}
Davis, James~M, Huseyin Topaloglu, David~P Williamson. 2015.
\newblock Assortment optimization over time.
\newblock {\it Operations Research Letters\/} {\bf 43}(6) 608--611.

\bibitem[{Farias et~al.(2013)Farias, Jagabathula, and
  Shah}]{farias2013nonparametric}
Farias, Vivek~F, Srikanth Jagabathula, Devavrat Shah. 2013.
\newblock A nonparametric approach to modeling choice with limited data.
\newblock {\it Management science\/} {\bf 59}(2) 305--322.

\bibitem[{Feige(1998)}]{feige1998threshold}
Feige, Uriel. 1998.
\newblock A threshold of ln n for approximating set cover.
\newblock {\it Journal of the ACM (JACM)\/} {\bf 45}(4) 634--652.

\bibitem[{Feldman and Topaloglu(2015)}]{feldman2015bounding}
Feldman, Jacob, Huseyin Topaloglu. 2015.
\newblock Bounding optimal expected revenues for assortment optimization under
  mixtures of multinomial logits.
\newblock {\it Production and Operations Management\/} {\bf 24}(10) 1598--1620.

\bibitem[{Ferreira et~al.(2019)Ferreira, Parthasarathy, and
  Sekar}]{ferreira2019learning}
Ferreira, Kris, Sunanda Parthasarathy, Shreyas Sekar. 2019.
\newblock Learning to rank an assortment of products.
\newblock {\it Available at SSRN 3395992\/} .

\bibitem[{Gallego et~al.(2015)Gallego, Ratliff, and
  Shebalov}]{gallego2015general}
Gallego, Guillermo, Richard Ratliff, Sergey Shebalov. 2015.
\newblock A general attraction model and sales-based linear program for network
  revenue management under customer choice.
\newblock {\it Operations Research\/} {\bf 63}(1) 212--232.

\bibitem[{Golbandi et~al.(2010)Golbandi, Koren, and
  Lempel}]{golbandi2010bootstrapping}
Golbandi, Nadav, Yehuda Koren, Ronny Lempel. 2010.
\newblock On bootstrapping recommender systems.
\newblock {\it Proceedings of the 19th ACM international conference on
  Information and knowledge management\/}. ACM, 1805--1808.

\bibitem[{Golbandi et~al.(2011)Golbandi, Koren, and
  Lempel}]{golbandi2011adaptive}
Golbandi, Nadav, Yehuda Koren, Ronny Lempel. 2011.
\newblock Adaptive bootstrapping of recommender systems using decision trees.
\newblock {\it Proceedings of the fourth ACM international conference on Web
  search and data mining\/}. ACM, 595--604.

\bibitem[{Han et~al.(2019)Han, G{\'o}mez, and Prokopyev}]{han2019assortment}
Han, Shaoning, Aandr{\'e}s G{\'o}mez, Oleg Prokopyev. 2019.
\newblock Assortment optimization and submodularity .

\bibitem[{Hawkins and Coney(1981)}]{hawkins1981uninformed}
Hawkins, Del~I, Kenneth~A Coney. 1981.
\newblock Uninformed response error in survey research.
\newblock {\it Journal of Marketing Research\/}  370--374.

\bibitem[{Hu and Pu(2009{\natexlab{a}})}]{hu2009acceptance}
Hu, Rong, Pearl Pu. 2009{\natexlab{a}}.
\newblock Acceptance issues of personality-based recommender systems.
\newblock {\it Proceedings of the third ACM conference on Recommender
  systems\/}. ACM, 221--224.

\bibitem[{Hu and Pu(2009{\natexlab{b}})}]{hu2009comparative}
Hu, Rong, Pearl Pu. 2009{\natexlab{b}}.
\newblock A comparative user study on rating vs. personality quiz based
  preference elicitation methods.
\newblock {\it Proceedings of the 14th international conference on Intelligent
  user interfaces\/}. ACM, 367--372.

\bibitem[{H{\"u}bner et~al.(2020)H{\"u}bner, Sch{\"a}fer, and
  Schaal}]{doi:10.1111/poms.13111}
H{\"u}bner, Alexander, Fabian Sch{\"a}fer, Kai~N. Schaal. 2020.
\newblock Maximizing profit via assortment and shelf-space optimization for
  two-dimensional shelves.
\newblock {\it Production and Operations Management\/} {\bf 29}(3) 547--570.
\newblock \doi{10.1111/poms.13111}.
\newblock
  \urlprefix\url{https://onlinelibrary.wiley.com/doi/abs/10.1111/poms.13111}.

\bibitem[{Kawahara et~al.(2009)Kawahara, Nagano, Tsuda, and
  Bilmes}]{kawahara2009submodularity}
Kawahara, Yoshinobu, Kiyohito Nagano, Koji Tsuda, Jeff~A Bilmes. 2009.
\newblock Submodularity cuts and applications.
\newblock {\it Advances in Neural Information Processing Systems\/}. 916--924.

\bibitem[{Kempe and Mahdian(2008)}]{kempe2008cascade}
Kempe, David, Mohammad Mahdian. 2008.
\newblock A cascade model for externalities in sponsored search.
\newblock {\it International Workshop on Internet and Network Economics\/}.
  Springer, 585--596.

\bibitem[{Kohrs(2001)}]{kohrs2001improving}
Kohrs, Arnd. 2001.
\newblock Improving collaborative filtering for new-users by smart object
  selection.
\newblock {\it Proc. Int'l Conf. on Media Features, 2001\/}.

\bibitem[{Krause and Guestrin(2005)}]{krause2005note}
Krause, Andreas, Carlos Guestrin. 2005.
\newblock {\it A note on the budgeted maximization of submodular functions\/}.
\newblock Carnegie Mellon University. Center for Automated Learning and
  Discovery.

\bibitem[{Kulik et~al.(2009)Kulik, Shachnai, and Tamir}]{kulik2009maximizing}
Kulik, Ariel, Hadas Shachnai, Tami Tamir. 2009.
\newblock Maximizing submodular set functions subject to multiple linear
  constraints.
\newblock {\it Proceedings of the twentieth annual ACM-SIAM symposium on
  Discrete algorithms\/}. Society for Industrial and Applied Mathematics,
  545--554.

\bibitem[{Li et~al.(2015)Li, Rusmevichientong, and Topaloglu}]{li2015d}
Li, Guang, Paat Rusmevichientong, Huseyin Topaloglu. 2015.
\newblock The d-level nested logit model: Assortment and price optimization
  problems.
\newblock {\it Operations Research\/} {\bf 63}(2) 325--342.

\bibitem[{Li(2007)}]{doi:10.1111/j.1937-5956.2007.tb00265.x}
Li, Zhaolin. 2007.
\newblock A single-period assortment optimization model.
\newblock {\it Production and Operations Management\/} {\bf 16}(3) 369--380.
\newblock \doi{10.1111/j.1937-5956.2007.tb00265.x}.
\newblock
  \urlprefix\url{https://onlinelibrary.wiley.com/doi/abs/10.1111/j.1937-5956.2007.tb00265.x}.

\bibitem[{Liu et~al.(2011)Liu, Meng, Liu, and Yang}]{liu2011wisdom}
Liu, Nathan~N, Xiangrui Meng, Chao Liu, Qiang Yang. 2011.
\newblock Wisdom of the better few: cold start recommendation via
  representative based rating elicitation.
\newblock {\it Proceedings of the fifth ACM conference on Recommender
  systems\/}. ACM, 37--44.

\bibitem[{Mahajan and Van~Ryzin(2001)}]{mahajan2001stocking}
Mahajan, Siddharth, Garrett Van~Ryzin. 2001.
\newblock Stocking retail assortments under dynamic consumer substitution.
\newblock {\it Operations Research\/} {\bf 49}(3) 334--351.

\bibitem[{Najafi et~al.(2019)Najafi, Duenyas, Jasin, and
  Uichanco}]{najafi2019multi}
Najafi, Sajjad, Izak Duenyas, Stefanus Jasin, Joline Uichanco. 2019.
\newblock Multi-product dynamic pricing with limited inventories under cascade
  click model.
\newblock {\it Available at SSRN 3362921\/} .

\bibitem[{Nemhauser and Wolsey(1978)}]{nemhauser1978best}
Nemhauser, George~L, Laurence~A Wolsey. 1978.
\newblock Best algorithms for approximating the maximum of a submodular set
  function.
\newblock {\it Mathematics of operations research\/} {\bf 3}(3) 177--188.

\bibitem[{Nemhauser et~al.(1978)Nemhauser, Wolsey, and
  Fisher}]{nemhauser1978analysis}
Nemhauser, George~L, Laurence~A Wolsey, Marshall~L Fisher. 1978.
\newblock An analysis of approximations for maximizing submodular set
  functions-i.
\newblock {\it Mathematical Programming\/} {\bf 14}(1) 265--294.

\bibitem[{Poe et~al.(1988)Poe, Seeman, McLaughlin, Mehl, and
  Dietz}]{poe1988don}
Poe, Gail~S, Isadore Seeman, Joseph McLaughlin, Eric Mehl, Michael Dietz. 1988.
\newblock ``don't know'' boxes in factual questions in a mail questionnaire:
  Effects on level and quality of response.
\newblock {\it Public Opinion Quarterly\/} {\bf 52}(2) 212--222.

\bibitem[{Rashid et~al.(2002)Rashid, Albert, Cosley, Lam, McNee, Konstan, and
  Riedl}]{rashid2002getting}
Rashid, Al~Mamunur, Istvan Albert, Dan Cosley, Shyong~K Lam, Sean~M McNee,
  Joseph~A Konstan, John Riedl. 2002.
\newblock Getting to know you: learning new user preferences in recommender
  systems.
\newblock {\it Proceedings of the 7th international conference on Intelligent
  user interfaces\/}. ACM, 127--134.

\bibitem[{Rashid et~al.(2008)Rashid, Karypis, and Riedl}]{rashid2008learning}
Rashid, Al~Mamunur, George Karypis, John Riedl. 2008.
\newblock Learning preferences of new users in recommender systems: an
  information theoretic approach.
\newblock {\it Acm Sigkdd Explorations Newsletter\/} {\bf 10}(2) 90--100.

\bibitem[{Roberts and Lattin(1991)}]{roberts1991development}
Roberts, John~H, James~M Lattin. 1991.
\newblock Development and testing of a model of consideration set composition.
\newblock {\it Journal of Marketing Research\/}  429--440.

\bibitem[{Rubens and Sugiyama(2007)}]{rubens2007influence}
Rubens, Neil, Masashi Sugiyama. 2007.
\newblock Influence-based collaborative active learning.
\newblock {\it Proceedings of the 2007 ACM conference on Recommender
  systems\/}. ACM, 145--148.

\bibitem[{Rusmevichientong et~al.(2010)Rusmevichientong, Shen, and
  Shmoys}]{rusmevichientong2010dynamic}
Rusmevichientong, Paat, Zuo-Jun~Max Shen, David~B Shmoys. 2010.
\newblock Dynamic assortment optimization with a multinomial logit choice model
  and capacity constraint.
\newblock {\it Operations research\/} {\bf 58}(6) 1666--1680.

\bibitem[{Rusmevichientong et~al.(2014)Rusmevichientong, Shmoys, Tong, and
  Topaloglu}]{rusmevichientong2014assortment}
Rusmevichientong, Paat, David Shmoys, Chaoxu Tong, Huseyin Topaloglu. 2014.
\newblock Assortment optimization under the multinomial logit model with random
  choice parameters.
\newblock {\it Production and Operations Management\/} {\bf 23}(11) 2023--2039.

\bibitem[{Sanchez and Morchio(1992)}]{sanchez1992probing}
Sanchez, Maria~Elena, Giovanna Morchio. 1992.
\newblock Probing ``dont know'' answers: Effects on survey estimates and
  variable relationships.
\newblock {\it Public Opinion Quarterly\/} {\bf 56}(4) 454--474.

\bibitem[{Saur{\'e} and Zeevi(2013)}]{saure2013optimal}
Saur{\'e}, Denis, Assaf Zeevi. 2013.
\newblock Optimal dynamic assortment planning with demand learning.
\newblock {\it Manufacturing \& Service Operations Management\/} {\bf 15}(3)
  387--404.

\bibitem[{Schuman and Presser(1996)}]{schuman1996questions}
Schuman, Howard, Stanley Presser. 1996.
\newblock {\it Questions and answers in attitude surveys: Experiments on
  question form, wording, and context\/}.
\newblock Sage.

\bibitem[{Strauss and Talluri(2017)}]{doi:10.1111/poms.12685}
Strauss, Arne~K., Kalyan Talluri. 2017.
\newblock Tractable consideration set structures for assortment optimization
  and network revenue management.
\newblock {\it Production and Operations Management\/} {\bf 26}(7) 1359--1368.
\newblock \doi{10.1111/poms.12685}.
\newblock
  \urlprefix\url{https://onlinelibrary.wiley.com/doi/abs/10.1111/poms.12685}.

\bibitem[{Tang(2020)}]{tang2020assortment}
Tang, Shaojie. 2020.
\newblock Assortment optimization with repeated exposures and product-dependent
  patience cost.
\newblock {\it arXiv preprint arXiv:2002.05321\/} .

\bibitem[{Tang and Yuan(2020)}]{tang2020product}
Tang, Shaojie, Jing Yuan. 2020.
\newblock Product sequencing and pricing under cascade browse model.
\newblock {\it Operations Research Letters\/} {\bf 48}(6) 687--692.

\bibitem[{Tang and Yuan(2021)}]{tang2021aamas}
Tang, Shaojie, Jing Yuan. 2021.
\newblock Adaptive cascade submodular maximization.
\newblock {\it Proceedings of the 2021 International Conference on Autonomous
  Agents \& Multiagent Systems\/}.

\bibitem[{Tayur et~al.(2012)Tayur, Ganeshan, and
  Magazine}]{tayur2012quantitative}
Tayur, Sridhar, Ram Ganeshan, Michael Magazine. 2012.
\newblock {\it Quantitative models for supply chain management\/}, vol.~17.
\newblock Springer Science \& Business Media.

\bibitem[{Teixeira et~al.(2002)Teixeira, de~Carvalho, Ramalho, and
  Corruble}]{teixeira2002activecp}
Teixeira, Ivan~R, Francisco de~AT de~Carvalho, Geber~L Ramalho, Vincent
  Corruble. 2002.
\newblock Activecp: A method for speeding up user preferences acquisition in
  collaborative filtering systems.
\newblock {\it Brazilian Symposium on Artificial Intelligence\/}. Springer,
  237--247.

\bibitem[{Tschiatschek et~al.(2017)Tschiatschek, Singla, and
  Krause}]{tschiatschek2017selecting}
Tschiatschek, Sebastian, Adish Singla, Andreas Krause. 2017.
\newblock Selecting sequences of items via submodular maximization.
\newblock {\it AAAI\/}. 2667--2673.

\bibitem[{Vondr{\'a}k et~al.(2011)Vondr{\'a}k, Chekuri, and
  Zenklusen}]{vondrak2011submodular}
Vondr{\'a}k, Jan, Chandra Chekuri, Rico Zenklusen. 2011.
\newblock Submodular function maximization via the multilinear relaxation and
  contention resolution schemes.
\newblock {\it Proceedings of the forty-third annual ACM symposium on Theory of
  computing\/}. ACM, 783--792.

\bibitem[{Wang(2012)}]{wang2012capacitated}
Wang, Ruxian. 2012.
\newblock Capacitated assortment and price optimization under the multinomial
  logit model.
\newblock {\it Operations Research Letters\/} {\bf 40}(6) 492--497.

\bibitem[{Yang and Padmanabhan(2005)}]{yang2005evaluation}
Yang, Yinghui, Balaji Padmanabhan. 2005.
\newblock Evaluation of online personalization systems: A survey of evaluation
  schemes and a knowledge-based approach.
\newblock {\it Journal of Electronic Commerce Research\/} {\bf 6}(2) 112.

\bibitem[{Zhang et~al.(2015)Zhang, Chong, Pezeshki, and
  Moran}]{zhang2015string}
Zhang, Zhenliang, Edwin~KP Chong, Ali Pezeshki, William Moran. 2015.
\newblock String submodular functions with curvature constraints.
\newblock {\it IEEE Transactions on Automatic Control\/} {\bf 61}(3) 601--616.

\end{thebibliography}

\clearpage
\section*{Online Appendix}

\section*{Proof of Lemma \ref{lem:submodular}:} \proof $g(\mathcal{S})$ is clearly non-decreasing according to \emph{information never hurts principle} \citep{krause2005note}.
Moreover, consider any $\mathcal{S}_1 \subseteq \mathcal{S}_2\subseteq \Omega$, we have $\mathcal{A}(\mathcal{S}_1) \subseteq \mathcal{A}(\mathcal{S}_2)$, it follows that $g(\mathcal{S}_1\cup\{q\})-g(\mathcal{S}_1\cup\{q\})=H(X_{\mathcal{A}(\mathcal{S}_1\cup\{q\})})-H(X_{\mathcal{A}(\mathcal{S}_1)})=H(X_{\mathcal{A}(\mathcal{S}_1)\cup\mathcal{Z}(\{q\})})-H(X_{\mathcal{Z}(\mathcal{A})})\geq H(X_{\mathcal{A}(\mathcal{S}_2)\cup\mathcal{A}(\{q\})})-H(X_{\mathcal{A}(\mathcal{S}_2)})=g(\mathcal{S}_2\cup\{q\})-g(\mathcal{S}_2\cup\{q\})$. The inequality is due to $H(X_{\mathcal{A}})$ as a function of $\mathcal{A}$ is submodular \citep{krause2005note}. Therefore, $g(\mathcal{S})$ is also non-decreasing and submodular. \endproof

\section*{Proof of Theorem \ref{thm:hard}:}
\proof
Consider a special case of problem $\textbf{P1}$ where (1) $g(\mathcal{S})=H(X_{\mathcal{A}(\mathcal{S})})$, i.e., we assume an entropy-like utility function as defined in Section \ref{sec:utility}, (2) $p^+_{Q[i]}=1$ and $c_{Q[i]}^+=1$, i.e., the customer is guaranteed to answer all questions, and (3) $\forall a\in \Phi: X_a\in\{0,1\}$ and $X_{\Phi}$ follows uniform distribution i.e., each attribute has binary value and $\forall \mathbf{x}\in \{0,1\}^{|\Phi|}, \Pr[X_{\Phi}=\mathbf{x}]=\frac{1}{2^{|\Phi|}}$. It is easy to see that finding a solution to this special case is reduced to selecting a subset of questions that covers the largest number of attributes subject to a cardinality constraint $b$. Next, using a reduction to the \emph{maximum coverage problem} \citep{feige1998threshold}, a known NP-hard problem, we show that $\textbf{P1}$ is NP-hard. Given sets $\{\mathcal{Y}_1, \cdots , \mathcal{Y}_n\}$ and a ground set $\{w_1, \cdots , w_m\}$ of elements to cover, the goal of the maximum coverage problem is to find a group of at most of $h$ sets so as the cover the largest number of elements. We next construct an equivalent instance of $\textbf{P1}$. We first set $b=h$.  There is an attribute $a_i$ for each element $w_i$, and there is a question $Q[j]$ for each set $\mathcal{Y}_j$, and we define $Q[j]$ covers $a_i$ if and only if $\mathcal{Y}_j$ covers $w_i$. Then finding an optimal solution to the maximum coverage problem is equivalent to solving the special case of $\textbf{P1}$ optimally. This finishes the proof of this theorem.
\endproof

{\section*{Proof of Theorem \ref{thm:harder}:}
\proof
We give an approximation-preserving reduction from the \emph{maximum coverage problem} \citep{feige1998threshold} to $\textbf{P1}$. It is well known that the \emph{maximum coverage problem} cannot be approximated in polynomial time within a ratio of $(1-1/e+\epsilon)$, unless $P=NP$. Consider a special case of problem $\textbf{P1}$ by setting (1) $g(\mathcal{S})=|\mathcal{A}(\mathcal{S})|$, i.e., the utility of $\mathcal{S}$ is defined as the total number of attributes covered by $\mathcal{S}$, (2) $p^+_{Q[i]}=1$ and $c_{Q[i]}^+=1$, i.e., the customer is guaranteed to answer all questions. First of all, $g(\mathcal{S})=|\mathcal{A}(\mathcal{S})|$ is monotone and submodular, thus the above setting is indeed a special case of $\textbf{P1}$. It is easy to see that finding a solution to this special case is reduced to selecting a subset of questions that covers the largest number of attributes subject to a cardinality constraint $b$. Next, using an approximation-preserving reduction from the \emph{maximum coverage problem} to the special case of $\textbf{P1}$, we show that $\textbf{P1}$ cannot be approximated in polynomial time within a ratio of $(1-1/e+\epsilon)$, unless $P=NP$. Given sets $\{\mathcal{Y}_1, \cdots , \mathcal{Y}_n\}$ and a ground set $\{w_1, \cdots , w_m\}$ of elements to cover, the goal of the maximum coverage problem is to find a group of at most of $h$ sets so as to cover the largest number of elements. We next construct an equivalent instance of $\textbf{P1}$. We first set $b=h$.  There is an attribute $a_i$ for each element $w_i$, and there is a question $Q[j]$ for each set $\mathcal{Y}_j$, and we define $Q[j]$ covers $a_i$ if and only if $\mathcal{Y}_j$ covers $w_i$. Firstly, give any valid solution $\{\mathcal{Y}_i\mid i\in L\}$ to the maximum coverage problem where $L\subseteq \{1, 2, \cdots, n\}$, assume $|\cup_{ i\in L}\mathcal{Y}_i| = c$, i.e., it covers $c$ elements, we immediately obtain a valid solution $\mathcal{S}=\{Q[i]\mid i\in L\}$ to the special case of $\textbf{P1}$ with $g(\mathcal{S})=c$. Conversely, if there exists a valid solution  $\mathcal{S}=\{Q[i]\mid i\in L\}$ to the special case of $\textbf{P1}$ which covers $c$ attributes, then we can find a valid solution $\{\mathcal{Y}_i\mid i\in L\}$ to the maximum coverage problem with $|\cup_{ i\in L}\mathcal{Y}_i| = c$. This finishes the proof of this theorem.
\endproof}

\section*{Proof of Lemma \ref{lem:1}:}
\proof Let $Q^*[i]$ denote the $i$-th question in $Q^*$. Assume $Q^*[k]$ is the last question in $Q^*$ whose reachability is no smaller than $\rho$, e.g.,  $k=\arg\max_{i} (C_{Q^*[i]}\geq \rho)$. Recall that we use $Q^*_{> k}$ (resp. $Q^*_{\leq k}$) to denote the sequence of questions scheduled after (resp. no later than) slot $k$. Therefore, the reachability of every question in $Q^*_{\leq k}$ is no smaller than $\rho$.

We first show that $C_{Q^*[k+1]} f(Q^*_{> k})\geq  f(Q^*)-f(Q^*_{\leq k})$. Let $\mathbf{1}_{\mathcal{S}}$ be indicator variable  that $\mathcal{S}\subseteq \mathcal{Q}^*_{\leq k}$ is answered by the customer and let $\mathbf{1}_{\mathcal{A}}$ be indicator variable that $\mathcal{A}\subseteq \mathcal{Q}^*_{> k}$ is answered by the customer.
\begin{eqnarray}
f(Q^*)&=& \sum_{\substack{\mathcal{S}\subseteq \mathcal{Q}^*_{\leq k}\\ \mathcal{A}\subseteq \mathcal{Q}^*_{> k}}} \Pr[\mathbf{1}_{\mathcal{S}}=1] (g(\mathcal{S})+ \Pr[\mathbf{1}_{\mathcal{A}}=1|\mathbf{1}_{\mathcal{S}}=1](g(\mathcal{S}\cup \mathcal{A})-g(\mathcal{S}))~\nonumber\\
&\leq& \sum_{\substack{\mathcal{S}\subseteq \mathcal{Q}^*_{\leq k}\\ \mathcal{A}\subseteq \mathcal{Q}^*_{> k}}}  \Pr[\mathbf{1}_{\mathcal{S}}=1] (g(\mathcal{S})+ \Pr[\mathbf{1}_{\mathcal{A}}=1|\mathbf{1}_{\mathcal{S}}=1]g(\mathcal{A})) \label{eq:6565}\\
&=&  \sum_{\substack{\mathcal{S}\subseteq \mathcal{Q}^*_{\leq k}\\ \mathcal{A}\subseteq \mathcal{Q}^*_{> k}}}  \Pr[\mathbf{1}_{\mathcal{S}}=1]g(\mathcal{S}) + \sum_{\substack{\mathcal{S}\subseteq \mathcal{Q}^*_{\leq k}\\ \mathcal{A}\subseteq \mathcal{Q}^*_{> k}}} \Pr[\mathbf{1}_{\mathcal{S}}=1] \Pr[\mathbf{1}_{\mathcal{A}}=1|\mathbf{1}_{\mathcal{S}}=1]g(\mathcal{A})~\nonumber\\
&=&  f(Q^*_{\leq k}) + \sum_{\substack{\mathcal{S}\subseteq \mathcal{Q}^*_{\leq k}\\ \mathcal{A}\subseteq \mathcal{Q}^*_{> k}}} \Pr[\mathbf{1}_{\mathcal{S}}=1] \Pr[\mathbf{1}_{\mathcal{A}}=1|\mathbf{1}_{\mathcal{S}}=1]g(\mathcal{A})\label{eq:added}\\
&=&  f(Q^*_{\leq k}) + \sum_{\substack{\mathcal{A}\subseteq \mathcal{Q}^*_{> k}}} \sum_{\substack{\mathcal{S}\subseteq \mathcal{Q}^*_{\leq k}}} \Pr[\mathbf{1}_{\mathcal{S}}=1] \Pr[\mathbf{1}_{\mathcal{A}}=1|\mathbf{1}_{\mathcal{S}}=1]g(\mathcal{A})~\nonumber\\
&=&  f(Q^*_{\leq k}) + \sum_{\substack{\mathcal{A}\subseteq \mathcal{Q}^*_{> k}}}   \Pr[\mathbf{1}_{\mathcal{A}}=1]g(\mathcal{A})~\nonumber\\
&=&  f(Q^*_{\leq k})+C_{Q^*[{k+1}]} f(Q^*_{> k}) \label{eq:9090}
\end{eqnarray}
Inequality (\ref{eq:6565}) is due to $g$ is a submodular function, Eq. (\ref{eq:added}) is due to $f(Q^*_{\leq k}) = \sum_{\substack{\mathcal{S}\subseteq \mathcal{Q}^*_{\leq k}}}  \Pr[\mathbf{1}_{\mathcal{S}}=1]g(\mathcal{S}) $, and Eq. (\ref{eq:9090}) is due to given $Q^*$, the first question of $Q^*_{> k}$ is reached with probability $C_{Q^*[{k+1}]}$.

It follows that $f(Q^*_{\leq k})\geq f(Q^*)-C_{Q^*[{k+1}]} f(Q^*_{> k})$. Since $C_{Q^*[{k+1}]} < \rho$ due to the definition of $k$, and $f(Q^*_{> k})\leq f(Q^*)$ due to $Q^*$ is the optimal solution, we have $f(Q^*_{\leq k})\geq (1-\rho)f(Q^*)$. Since every question in $Q^*_{\leq k}$ can be reached with probability at least $\rho$ and  $|Q^*_{\leq k}|\leq b$, $Q^*_{\leq k}$ satisfies all conditions specified in this lemma. \endproof

\section*{Proof of Lemma \ref{lem:9}:}
\proof First, $u(q, \mathcal{S})= p_q^+ g(\mathcal{S}\cup\{q\})+(1-p_q^+)g(\mathcal{S})$, as a function of $\mathcal{S}$, is clearly monotone due to $g$ is a monotone function. We next show that for any fixed $q$,  $g(\mathcal{S}\cup\{q\})$ is submodular as a  function of $\mathcal{S}$. For any  $\mathcal{S}_1 \subseteq \mathcal{S}_2 \subseteq \Omega$ and $q'\notin \mathcal{S}_2$, we have $g(\mathcal{S}_1\cup \{q'\}\cup\{q\})-g(\mathcal{S}_1\cup\{q\})\geq  g(\mathcal{S}_2\cup \{q'\}\cup\{q\})-g(\mathcal{S}_2\cup\{q\})$ due to $g(\mathcal{S})$ is submodular and $\mathcal{S}_1\cup \{q\}\subseteq \mathcal{S}_2\cup \{q\}$. Thus, $g(\mathcal{S}\cup\{q\})$ is a submodular function of $\mathcal{S}$. It follows that for any fixed $q$, $u(q, \mathcal{S})$ is submodular due to the linear combination of two submodular functions is submodular. \endproof

\section*{Proof of Lemma \ref{lem:3}:}
\proof Due to $\mathcal{S}'$ satisfies constraint (C2) in problem $\textbf{P2}$, we have $\prod_{q\in \mathcal{S}'} p^+_{q}c_{q}^+\geq \rho$. It follows that with probability at least $\rho$, all questions in $\mathcal{S}'$ will be answered  and $q'$ will be read. Moreover, the probability of $q'$ being answered is $p_{q'}^+$ conditioned on all questions in $\mathcal{S}'$ are answered and $q'$ is read. It follows that $f(Q^{\mathrm{Alg1}})\geq \rho p_{q'}^+ g(\mathcal{S}'\cup\{q'\})+\rho (1-p_{q'}^+) g(\mathcal{S}')=\rho u(q', \mathcal{S}')$. \endproof

\section*{Proof of Theorem \ref{thm:1}:}
\proof For any $\rho\in[0,1]$, let $Q^\diamond$  denote the optimal solution to the following problem:
\begin{eqnarray}
\max_Q f(Q) \mbox{ subject to } |Q|\leq b; \forall i\in\{1,2,\cdots, |Q|\}: C_{Q[i]}\geq\rho\label{question}
\end{eqnarray}
Lemma \ref{lem:1} implies that $f(Q^\diamond) \geq (1-\rho)f(Q^*)$. Therefore, in order to prove this theorem, it suffice to prove that  $f(Q^{\mathrm{Alg1}})\geq \rho (1-1/e-\epsilon) f(Q^\diamond)$.

Assume $|Q^\diamond|=z$, let  $Q^\diamond_{<z}$ denote the subsequence of  $Q^\diamond$ by excluding the last question $Q^\diamond[z]$. Because $Q^\diamond$ is a valid solution to problem (\ref{question}), $(Q^\diamond[z],\mathcal{Q}^\diamond_{<z}\})$ is a valid solution to problem $\textbf{P2}$. Therefore, $u(q', \mathcal{S}')\geq (1-1/e-\epsilon)u(Q^\diamond[z], \mathcal{Q}^\diamond_{<z})$ due to  Algorithm \ref{alg:greedy-peak} finds a $(1-1/e-\epsilon)$ approximate solution to $\textbf{P2}$ (Lemma \ref{lem:90}). We next prove that $f(Q^\diamond)\leq u(Q^\diamond[z], \mathcal{Q}^\diamond_{<z})$.
 \begin{eqnarray}f(Q^\diamond) &=&f(Q^\diamond_{<z})+ (f(Q^\diamond)-f(Q^\diamond_{<z}))~\nonumber\\
 &=& f(Q^\diamond_{<z})+ C_{Q^\diamond[z]}p^+_{Q^\diamond[z]}(g(\mathcal{Q}^\diamond)- g(\mathcal{Q}^\diamond_{<z}))~\nonumber\\
 &\leq& g(\mathcal{Q}^\diamond_{<z})+ p^+_{Q^\diamond[z]}(g(\mathcal{Q}^\diamond)- g(\mathcal{Q}^\diamond_{<z}))\label{eq:7}\\
 &=& u(Q^\diamond[z], \mathcal{Q}^\diamond_{<z}) \label{eq:81}
 \end{eqnarray}
 Inequality (\ref{eq:7}) is due to $f(Q^\diamond_{<z})\leq g(\mathcal{Q}^\diamond_{<z})$ and $C_{Q^\diamond[z]}\leq1$. Eq. (\ref{eq:81}) is due to the definition of $u(Q^\diamond[z], \mathcal{Q}^\diamond_{<z})$. Together with Lemma \ref{lem:3}, we have $f(Q^{\mathrm{Alg1}})\geq \rho u(q', \mathcal{S}') \geq \rho (1-1/e-\epsilon)u(Q^\diamond[z], \mathcal{Q}^\diamond_{<z})\geq \rho (1-1/e-\epsilon) f(Q^\diamond)$. \endproof

\section*{Proof of Lemma \ref{lem:67}:}
 \proof Assume $\mathbf{r}$ is a (random) realization of  $\mathcal{R}(\Omega\setminus\{q\})$, let $\Pr[\mathbf{r}]$ denote the probability that $\mathbf{r}$ is realized, we have
 \begin{eqnarray}
 v(q, \mathcal{S})&=& \mathbb{E}[g(\mathcal{R} (\mathcal{S}\cup\{q\}))]\\
 &=&p^+_q\sum_{\mathbf{r}\subseteq \Omega\setminus\{q\}} \Pr[\mathbf{r}]g(\mathbf{r}\cap \mathcal{S} \cup \{q\})+(1-p^+_q)\sum_{\mathbf{r}\subseteq \Omega\setminus\{q\}} \Pr[\mathbf{r}]g(\mathbf{r}\cap \mathcal{S})
 \end{eqnarray}

We next prove that for any fixed $\mathbf{r}$ and $q$, $g(\mathbf{r}\cap \mathcal{S} \cup \{q\})$ as a function of $\mathcal{S}$ is monotone and submodular. First, $g(\mathbf{r}\cap \mathcal{S} \cup \{q\})$, as a function of $\mathcal{S}$,  is clearly monotone due to $g$ is monotone. We focus on proving $g(\mathbf{r}\cap \mathcal{S} \cup \{q\})$ is  a sumodular function of $\mathcal{S}$. For any  $\mathcal{S}_1 \subseteq \mathcal{S}_2 \subseteq \Omega$ and $q'\notin \mathcal{S}_2$, we have  $g(\mathbf{r}\cap (\mathcal{S}_1\cup\{q'\}) \cup \{q\})- g(\mathbf{r}\cap \mathcal{S}_1 \cup \{q\})=g((\mathbf{r}\cap \mathcal{S}_1)\cup (\mathbf{r}\cap \{q'\}) \cup \{q\})- g(\mathbf{r}\cap \mathcal{S}_1 \cup \{q\}) \geq g((\mathbf{r}\cap \mathcal{S}_2)\cup (\mathbf{r}\cap \{q'\}) \cup \{q\})- g(\mathbf{r}\cap \mathcal{S}_2 \cup \{q\}) $. The inequality is due to $(\mathbf{r}\cap \mathcal{S}_1) \subseteq (\mathbf{r}\cap \mathcal{S}_2)$ and $g$ is submodular. Thus $g(\mathbf{r}\cap \mathcal{S} \cup \{q\})$ is a submodular function of $\mathcal{S}$. By a similar proof, we can show that $g(\mathbf{r}\cap \mathcal{S})$ is also a sumodular function of $\mathcal{S}$. It follows that $ v(q, \mathcal{S})$ is  a monotone and submodular function of $\mathcal{S}$ due to linear combination of monotone and submodular functions is monotone and submodular. \endproof

{\section*{Proof of Lemma \ref{lem:7}:}\proof
Consider the case when all questions in $Q^*_{\leq k}$ are guaranteed to be read by the customer, which clearly does not decrease the expected utility of $Q^*_{\leq k}$ due to $g$ is monotone, %e.g., this can be achieved by setting $p^-_q=1-p^+_q$ and $c_q^+=c_q^-=1$ for all $q\in Q^*_{\leq k}$,
the set of questions answered by the customer can be obtained by including each question $q\in Q^*_{\leq k}$ with probability $p^+_q$. Recall that $\mathcal{R}(Q^*_{\leq k})$ is a random set obtained by including each question $q\in Q^*_{\leq k}$ with probability $p^+_q$, thus the expected utility $\mathbb{E}[g(\mathcal{R}(Q^*_{\leq k}))]$  of $\mathcal{R}(Q^*_{\leq k})$ is lower bounded by  $f(Q^*_{\leq k})$, i.e., $\mathbb{E}[g(\mathcal{R}(Q^*_{\leq k}))] \geq f(Q^*_{\leq k}) $. %when $p^-_q=1-p^+_q$ and $c_q^+=c_q^-=1$  for all $q\in Q^*_{\leq k}$,
\endproof}

\section*{Proof of Lemma \ref{lem:9999999}:}
\proof  We first introduce some useful notations. Given the solution $Q^{\mathrm{Alg2}}$ that is returned from Algorithm \ref{alg:greedy-peak1}, let $\mathcal{J}(Q^{\mathrm{Alg2}})$ denote the (random) set of questions answered by the customer given $Q^{\mathrm{Alg2}}$. Then we have $f(Q^{\mathrm{Alg2}})=\mathbb{E}[g(\mathcal{J}(Q^{\mathrm{Alg2}}))]$.  For notational simplicity, we use $\mathcal{J}$ (resp. $\mathcal{R}$) to denote $\mathcal{J}(Q^{\mathrm{Alg2}})$ (resp. $\mathcal{R}(Q^{\mathrm{Alg2}})$) for short in the rest of this proof. Because  $f(Q^{\mathrm{Alg2}})= \mathbb{E}[g(\mathcal{J})]$, we focus on proving $\mathbb{E}[g(\mathcal{J})]\geq \rho \mathbb{E}[g(\mathcal{R})]$.

  Define $\mathcal{J}_{\geq i}= \mathcal{J} \cap \mathcal{Q}^{\mathrm{Alg2}}_{\geq i}$ and  $\mathcal{R}_{\geq i}= \mathcal{R} \cap \mathcal{Q}^{\mathrm{Alg2}}_{\geq i}$. For ease of presentation, define $g(\mathcal{J}_{\geq |{Q}^{\mathrm{Alg2}}|+1})=g(\mathcal{R}_{\geq |{Q}^{\mathrm{Alg2}}|+1})=0$.
  % with $g(Q^{\mathrm{Alg2}}[{|Q^{\mathrm{Alg2}}|+1}])=0$.
  The main result that we will prove is that for any fixed $i\in \{1,2,\cdots |Q^{\mathrm{Alg2}}|\}$,
\begin{equation}
\label{eq:1}
\mathbb{E}[g(\mathcal{J}_{\geq i})-g(\mathcal{J}_{\geq i+1})]\geq \rho \mathbb{E}[g(\mathcal{R}_{\geq i})-g(\mathcal{R}_{\geq i+1})]
\end{equation}
Then the theorem follows from Eq. (\ref{eq:1}) since
\[\mathbb{E}[g(\mathcal{J})]=\sum_{i=1}^{|Q^{\mathrm{Alg2}}|} \mathbb{E}[g(\mathcal{J}_{\geq i})-g(\mathcal{J}_{\geq i+1})]\geq \rho \sum_{i=1}^{|Q^{\mathrm{Alg2}}|} \mathbb{E}[g(\mathcal{R}_{\geq i})-g(\mathcal{R}_{\geq i+1})]= \rho \mathbb{E}[g(\mathcal{R})]\]
Based on this observation, we next prove Eq. (\ref{eq:1}). The case when $i=|{Q}^{\mathrm{Alg2}}|$ is trivial due to $\mathbb{E}[g(\mathcal{J}_{\geq |\mathcal{Q}^{\mathrm{Alg2}}|})]=C_{Q^{\mathrm{Alg2}}[|{Q}^{\mathrm{Alg2}}|]}\mathbb{E}[g(\mathcal{R}_{\geq |{Q}^{\mathrm{Alg2}}|})]\geq \rho\mathbb{E}[g(\mathcal{R}_{\geq |{Q}^{\mathrm{Alg2}}|})]$. We next focus on the case when $i<|{Q}^{\mathrm{Alg2}}|$.  %Define $Q^{\mathrm{Alg2}}_{\geq i}=\{Q[j]: Q[j]\in Q^{\mathrm{Alg2}}, j\geq i\}$.
   Notice that the distribution of $\mathcal{J}$ is determined by the cascade browse model. For ease of analysis, for any fixed slot $i$, we next introduce an alternative way to generate the distribution of $\mathcal{J}_{\geq i}$: For every $q\in Q^{\mathrm{Alg2}}$,  (1) we first determine whether $q$ will be answered or not given that $q$ has been read by the customer,  and (2) then determine whether $q$ will be read or not. In particular,  for each $i\in \{1, 2, \cdots, |{Q}^{\mathrm{Alg2}}|-1\}$, we first construct a random set $\mathcal{R}$, then %Let $Q[j]$ denote the $j$-th question in $Q^{\mathrm{Alg2}}$.
   % we
   generate two random sets $\mathcal{U}^+_{>i}$ and $\mathcal{U}^-_{>i}$ based on $\mathcal{R}$ as follows:

   \begin{center}
\framebox[1\textwidth][c]{
\enspace
\begin{minipage}[t]{0.95\textwidth}
   \begin{itemize}
   \item Initially, $\mathcal{R}$ is randomly generated, $\mathcal{U}^+_{>i}=\emptyset$, $\mathcal{U}^-_{>i}=\emptyset$.
%\item Adding $q_1$ to $\mathcal{U}$.
  \item Generating $\mathcal{U}^+_{>i}$:
  \begin{enumerate}
  \item Starting  from $j=i+1$.
  \item If $Q^{\mathrm{Alg2}}[{j-1}]\in \mathcal{R}_{> i}\cup{Q^{\mathrm{Alg2}}[i]}$, add $Q^{\mathrm{Alg2}}[{j}]$ to $\mathcal{U}^+_{> i}$ with probability $c^+_{Q^{\mathrm{Alg2}}[{j-1}]}$; if $Q^{\mathrm{Alg2}}[{j-1}]\notin \mathcal{R}_{> i}\cup{Q^{\mathrm{Alg2}}[i]}$, add $Q^{\mathrm{Alg2}}[{j}]$ to $\mathcal{U}^+_{> i}$ with probability  $\frac{p^-_{Q^{\mathrm{Alg2}}[{j-1}]}}{1-p^+_{Q^{\mathrm{Alg2}}[{j-1}]}}\cdot c^-_{Q^{\mathrm{Alg2}}[{j-1}]}$.
  \item If $Q^{\mathrm{Alg2}}[{j}]$ is not  added to $\mathcal{U}^+_{>i}$, return $\mathcal{U}^+_{> i}$, otherwise, goto $2$ with $j=j+1$. Return $\mathcal{U}^+_{> i}$ also once no more questions remain.
  \end{enumerate}
  \item Generating $\mathcal{U}^-_{>i}$:
  \begin{enumerate}
  \item Starting  from $j=i+1$.
  \item If $Q^{\mathrm{Alg2}}[{j-1}]\in \mathcal{R}_{> i}$, add $Q^{\mathrm{Alg2}}[{j}]$ to $\mathcal{U}^-_{> i}$ with probability $c^+_{Q^{\mathrm{Alg2}}[{j-1}]}$; if $Q^{\mathrm{Alg2}}[{j-1}]\notin \mathcal{R}_{> i}$, add $Q^{\mathrm{Alg2}}[{j}]$ to $\mathcal{U}^-_{> i}$ with probability  $\frac{p^-_{Q^{\mathrm{Alg2}}[{j-1}]}}{1-p^+_{Q^{\mathrm{Alg2}}[{j-1}]}}\cdot c^-_{Q^{\mathrm{Alg2}}[{j-1}]}$.
  \item If $Q^{\mathrm{Alg2}}[{j}]$ is not   added to $\mathcal{U}^-_{>i}$, return $\mathcal{U}^-_{> i}$, otherwise, goto $2$ with $j=j+1$. Return $\mathcal{U}^-_{> i}$ also once no more questions remain.
    \end{enumerate}
  \end{itemize}
\end{minipage}
}
\end{center}
\vspace{0.1in}

 Intuitively, $\mathcal{R}_{> i}$ includes those questions in $Q^{\mathrm{Alg2}}_{>i}$ that can be answered by the customer if they have been read,  $\mathcal{U}^+_{> i}$ includes those questions in  $Q^{\mathrm{Alg2}}_{>i}$ that can be read by the customer given that $Q^{\mathrm{Alg2}}[i]$ is answered, and $\mathcal{U}^-_{> i}$ includes those questions in  $Q^{\mathrm{Alg2}}_{>i}$ that can be read by the customer given that $Q^{\mathrm{Alg2}}[i]$ is reached but not answered. Now we are able to express $\mathbb{E}[g(\mathcal{J}_{\geq i})]$ and $\mathbb{E}[g(\mathcal{J}_{\geq i+1})]$ using $\mathcal{R}$, $\mathcal{U}^+_{>i}$, and $\mathcal{U}^-_{>i}$.
    \begin{eqnarray}
  \mathbb{E}[g(\mathcal{J}_{\geq i+1})]= &&\Pr[{Q^{\mathrm{Alg2}}[i] \mbox{ is answered}}]\cdot \mathbb{E}_{\mathcal{R}, \mathcal{U}^+_{> i}} [g(\mathcal{R}_{>i}\cap \mathcal{U}^+_{>i})]~\nonumber\\
  &&+\Pr[{Q^{\mathrm{Alg2}}[i] \mbox{ is reached but not answered}}]\cdot  \mathbb{E}_{\mathcal{R}, \mathcal{U}^-_{> i}} [g(\mathcal{R}_{>i}\cap \mathcal{U}^-_{>i})]\label{eq:ii}
  \end{eqnarray}
      \begin{eqnarray}
  \mathbb{E}[g(\mathcal{J}_{\geq i})]= &&\Pr[{Q^{\mathrm{Alg2}}[i] \mbox{ is answered}}]\cdot \mathbb{E}_{\mathcal{R}, \mathcal{U}^+_{> i}} [g(Q^{\mathrm{Alg2}}[i]\cup (\mathcal{R}_{>i}\cap \mathcal{U}^+_{>i}))]~\nonumber\\
  &&+\Pr[{Q^{\mathrm{Alg2}}[i] \mbox{ is reached but not answered}}]\cdot  \mathbb{E}_{\mathcal{R}, \mathcal{U}^-_{> i}} [g(\mathcal{R}_{>i}\cap \mathcal{U}^-_{>i})]\label{eq:iii}
  \end{eqnarray}%It is easy to verify that $\mathcal{J}_{>i}$ has the same distribution of  $\mathcal{R}\cap \mathcal{U}$.   We use $\mathcal{R}_{\geq i}$ (resp. $\mathcal{U}_{\geq i}$) to denote $\mathcal{R} \cap Q^{\mathrm{Alg2}}_{\geq i}$ (resp. $\mathcal{U} \cap Q^{\mathrm{Alg2}}_{\geq i}$), it follows that $\mathcal{J}_{\geq i}$ has the same distribution of  $\mathcal{R}_{\geq i}\cap \mathcal{U}_{\geq i}$.
%  \begin{eqnarray}
%  &&E[f(J\cap Q^{\mathrm{Alg2}}_{\geq i})-f(J\cap Q^{\mathrm{Alg2}}_{\geq i-1})]~\nonumber\\
%  &=& \Pr[i\in R]E_R[\Pr[i\in U]E_{U}[f_{R\cap U\cap Q^{\mathrm{Alg2}}_{\geq i-1}}(i)|R]|i\in R] \label{eq:0}\\
% &\geq& \Pr[i\in R]E_R[\Pr[i\in U] f_{R\cap Q^{\mathrm{Alg2}}_{\geq i-1}}(i)|i\in R]\label{eq:2}\\
% &\geq& \Pr[i\in R]E_R[\Pr[i\in U|i\in R]E_R[f_{R\cap Q^{\mathrm{Alg2}}_{\geq i-1}}(i)|i\in R] \label{eq:3}\\
%  &\geq& \rho \Pr[i\in R] E_R[f_{R\cap Q^{\mathrm{Alg2}}_{\geq i-1}}(i)|i\in R]~\nonumber\\
%  &=&\rho E[f_{R\cap Q^{\mathrm{Alg2}}_{\geq i-1}}(i)]=\rho E(f(R\cap Q^{\mathrm{Alg2}}_{\geq i})-f(R\cap Q^{\mathrm{Alg2}}_{\geq i-1}))~\nonumber\\
%      \end{eqnarray}

We next focus on bounding the value of $\mathbb{E}[g(\mathcal{J}_{\geq i})-g(\mathcal{J}_{\geq i+1})]$. %Let $\mathbf{1}_{Q^{\mathrm{Alg2}}[i]\in \mathcal{R}_{\geq i} }$ (resp. $\mathbf{1}_{Q^{\mathrm{Alg2}}[i]\in \mathcal{U}_{\geq i}}$) denote the event that $Q^{\mathrm{Alg2}}[i]$ is included in $\mathcal{R}_{\geq i}$ (resp. $\mathcal{U}_{\geq i}$).
For notational simplicity, define $g_\mathcal{S}(q)=g(\mathcal{S}\cup\{q\})-g(\mathcal{S})$ as the marginal benefit of $q$ given $\mathcal{S}$. According to equations  (\ref{eq:ii}) and (\ref{eq:iii}), for any fixed $i$, we have
  \begin{eqnarray}
  &&\mathbb{E}[g(\mathcal{J}_{\geq i})-g(\mathcal{J}_{\geq i+1})]~\nonumber\\
  &=& \Pr[{Q^{\mathrm{Alg2}}[i] \mbox{ is answered}}]\cdot \mathbb{E}_{\mathcal{R}, \mathcal{U}^+_{> i}}[g_{\mathcal{R}_{>i}\cap \mathcal{U}^+_{>i}}(Q^{\mathrm{Alg2}}[i])] \label{eq:0}\\
 &\geq& \Pr[{Q^{\mathrm{Alg2}}[i] \mbox{ is answered}}]\cdot \mathbb{E}_{\mathcal{R}, \mathcal{U}^+_{> i}}[g_{\mathcal{R}_{>i}}(Q^{\mathrm{Alg2}}[i])]\label{eq:2}\\
 &=& \Pr[{Q^{\mathrm{Alg2}}[i] \mbox{ is answered}}]\cdot \mathbb{E}_{\mathcal{R}}[g_{\mathcal{R}_{>i}}(Q^{\mathrm{Alg2}}[i])]\label{eq:3}\\
 &=& \Pr[{Q^{\mathrm{Alg2}}[i] \mbox{ is reached}}]\cdot \Pr[{Q^{\mathrm{Alg2}}[i] \mbox{ is answered after being reached}}]\cdot \mathbb{E}_{\mathcal{R}}[g_{\mathcal{R}_{>i}}(Q^{\mathrm{Alg2}}[i])]\\
 &\geq& \rho \cdot p^+_{Q^{\mathrm{Alg2}}[i]} \cdot \mathbb{E}_{\mathcal{R}}[g_{\mathcal{R}_{>i}}(Q^{\mathrm{Alg2}}[i])]\label{eq:4}\\
 % &\geq& \rho \Pr[{Q^{\mathrm{Alg2}}[i]\in \mathcal{R}_{\geq i} }] \mathbb{E}_{\mathcal{R}_{\geq i}}[g_{\mathcal{R}_{\geq i+1} }(Q^{\mathrm{Alg2}}[i])|{Q^{\mathrm{Alg2}}[i]\in \mathcal{R}_{\geq i} }]\label{eq:111}\\
&=& \rho  \cdot \mathbb{E}_{\mathcal{R}}[g(\mathcal{R}_{>i-1})-g(\mathcal{R}_{>i})]~\nonumber
      \end{eqnarray}

Eq. (\ref{eq:2}) is due to   $g$ is a submodular function. Inequality (\ref{eq:4}) is is due to the fact that every question in $Q^{\mathrm{Alg2}}$ has reachability no less than $\rho$, e.g., $\Pr[{Q^{\mathrm{Alg2}}[i] \mbox{ is reached}}]\geq\rho$.
\endproof

{\section*{Proof of Theorem \ref{thm:@}:}\proof Lemma \ref{lem:1}, which holds under the general model, together with Lemma \ref{lem:7} implies that $\mathbb{E}[g(\mathcal{R}(Q^*_{\leq k}))]\geq f(Q^*_{\leq k})\geq  (1-\rho)  f(Q^*)$. We next focus on proving \begin{equation}\label{eq:88}\mathbb{E}[g(\mathcal{R}(Q^{\mathrm{Alg2}}))]\geq (1-1/e-\epsilon)\mathbb{E}[g(\mathcal{R}(Q^*_{\leq k}))]\end{equation} Then this theorem follows from Inequality (\ref{eq:88}) and Lemma \ref{lem:9999999}.

Note that for a fixed $q$, Line \ref{line:3} of Algorithm \ref{alg:greedy-peak1} finds a $(1-1/e-\epsilon)$ approximate solution to \textbf{P.3}. Since we enumerate  all possibilities of $q$ and return the best solution $(q', \mathcal{S}')$, it is easy to verify that $v(q', \mathcal{S}')\geq (1-1/e-\epsilon)v^*$ where $v^*$ denotes the optimal solution to \textbf{P3}. In addition, $(Q^*[k], \mathcal{Q}^*_{\leq k-1})$ is a valid solution to \textbf{P3}. Thus $(1-1/e-\epsilon)v(Q^*[k], \mathcal{Q}^*_{\leq k-1})\leq (1-1/e-\epsilon)v^* \leq v(q', \mathcal{S}')$. According to the definition of $v$, we have $\mathbb{E}[g(\mathcal{R}(Q^*_{\leq k}))]=v(Q^*[k], \mathcal{Q}^*_{\leq k-1})$ and $\mathbb{E}[g(\mathcal{R}(Q^{\mathrm{Alg2}}))]=v(q', \mathcal{S}')$, it follows that $\mathbb{E}[g(\mathcal{R}(Q^{\mathrm{Alg2}}))]\geq (1-1/e-\epsilon)\mathbb{E}[g(\mathcal{R}(Q^*_{\leq k}))]$. \endproof}

\section*{Proof of Lemma \ref{lem:767676}:}\proof
Given $Q$ and any $Q[i]\in Q$, let $\Delta^+$ (resp. $\Delta^-$) denote the marginal benefit of all questions scheduled at and after slot $i$ conditioned on $Q[i]$ has been answered (resp. PNA). Hence, the expected utility of $Q$ can be written as $f(Q)=f(Q_{<i})+ C_{Q[i]} (\lambda_i p^+_{Q[i]} \Delta^+ +  p^-_{Q[i]}\Delta^{-})$. Since both $ p^+_{Q[i]} \Delta^+$ and $\lambda_i p^+_{Q[i]} \Delta^+ +  p^-_{Q[i]}\Delta^{-}$ are non-negative, $f(Q)$ is a non-decreasing function of $\lambda_i$ and $C_{Q[i]}$.
\endproof

\section*{Proof of Lemma \ref{lem:22}:}
\proof
Recall that $Q^*[i]$ denotes the $i$-th question in $Q^*$. Assume that $Q^*[k]$ is the last question in $Q^*$ whose reachability is no smaller than $\rho$. The main result that we will prove is that
\begin{equation}
\label{eq:99}
C_{Q^*[{k+1}]} f(Q^*_{> k})\geq  f(Q^*)-f(Q^*_{\leq k})
 \end{equation}

Then this lemma follows from $C_{Q^*[{k+1}]} < \rho$ and $f(Q^*_{> k}) \leq f(Q^*)$. We next prove (\ref{eq:99}). We use $f(Q|Q'\oplus Q)$ to denote the \emph{conditional} utility of $Q$ given that $Q'$ is scheduled ahead of $Q$ and the first question of $Q$ has been read. We use $f(Q)$ to denote  $f(Q|\emptyset\oplus Q)$ for short. It follows that  $C_{Q^*[{k+1}]} f(Q^*_{> k}|Q^*_{\leq k}\oplus Q^*_{> k})\geq  f(Q^*)-f(Q^*_{\leq k})$ due to for any given $Q^*$, the first question of $Q^*_{> k}$ can be reached with probability $C_{Q^*[{k+1}]} $ and $g$ is a submodular function. In order to prove Inequality (\ref{eq:99}), it remains to prove that $f(Q^*_{> k}) \geq f(Q^*_{> k}|Q^*_{\leq k}\oplus Q^*_{> k})$, i.e.,  we need to show that moving $Q^*_{> k}$ $k$ slots earlier does not decrease its utility. Because $\forall i\geq j: \lambda_i\leq \lambda_j$, it implies that moving a question to some earlier slot does not decrease its reachability and answer-through-rate, then we have  $f(Q^*_{> k}) \geq f(Q^*_{> k}|Q^*_{\leq k}\oplus Q^*_{> k})$ due to Lemma \ref{lem:767676}. %Here we give the main intuition behind the proof, moving $Q^*_{> k}$ $k$ slots earlier does not decrease the answer-through-rate and reachability of any question in $Q^*_{> k}$.

\endproof

\section*{Proof of Lemma \ref{lem:333}:}
\proof
According to Algorithm \ref{alg:greedy-peak3}, for any fixed $t$ and $q$, we are able to find a $(1-1/e-\epsilon)$ approximate solution to $\textbf{P2.1}$. Then this lemma follows from the fact that $(t', q', \mathcal{S}')$ is returned as the best solutions after exhaustively trying all possibilities of $t$ and $q$.
\endproof

\section*{Proof of Lemma \ref{lem:44444}:}
\proof Due to $\mathcal{S}'$ satisfies constraint (C2.1) in problem $\textbf{P2.1}$, we have $\Lambda_{t'-1}\prod_{q\in \mathcal{S}'} p^+_{q}c_{q}^+\geq \rho$. It follows that with probability at least $\rho$, all questions in $\mathcal{S}'$ will be answered and $q'$ will be read. Moreover, the probability that $q'$ will be answered by the customer is $\lambda_{t'}p_{q'}^+$ conditioned on all questions in $\mathcal{S}'$ are answered and $q'$ is read. It follows that $f(Q^{\mathrm{Alg3}})\geq \rho \lambda_{t'}p_{q'}^+ g(\mathcal{S}'\cup\{q'\})+\rho (1-\lambda_{t'}p_{q'}^+) g(\mathcal{S}')=\rho u(t', q', \mathcal{S}')$. \endproof

\section*{Proof of Theorem \ref{thm:extend1}:}
\proof Lemma \ref{lem:22}, which holds under the general model, together with Lemma \ref{lem:777} implies that
 \begin{equation}
\label{eq:222}
\mathbb{E}[g(\mathcal{R}(Q^*_{\leq k}))]\geq f(Q^*_{\leq k})\geq  (1-\rho)  f(Q^*)
 \end{equation}

 We next focus on proving \begin{equation}\label{eq:8}\mathbb{E}[g(\mathcal{R}(Q^{\mathrm{Alg4}}))]\geq 0.38\mathbb{E}[g(\mathcal{R}(Q^*_{\leq k}))]\end{equation} Then this theorem follows from Inequality (\ref{eq:8}), (\ref{eq:222}) and Lemma \ref{lem:88888}.

Note that for a fixed pair of $t$ and $q$, Line \ref{line:77} of Algorithm \ref{alg:greedy-peak4} finds a $0.38$ approximate solution to problem $\textbf{P3.1}$. Since $(t', {q}', {\mathcal{S}^\nu}')$ is the best candidate solution after enumerating  all possibilities of $t$ and $q$, it is easy to verify that $v(t', {q}', {\mathcal{S}^\nu}')\geq 0.38 v^*$ where $v^*$ denotes the utility of the optimal solution to \textbf{P3.1}. In addition, because $(k, Q^*[k], \mathcal{Q}^*_{\leq k-1})$ is a valid solution to \textbf{P3.1}, we have $0.38 v(k, Q^*[k], \mathcal{Q}^*_{\leq k-1})\leq 0.38v^* \leq v(t', {q}', {\mathcal{S}^\nu}')$. According to the definition of $v$, we have $\mathbb{E}[g(\mathcal{R}(Q^*_{\leq k}))]=v(k, Q^*[k], \mathcal{Q}^*_{\leq k-1})$ and $\mathbb{E}[g(\mathcal{R}(Q^{\mathrm{Alg4}}))]=v(t', {q}', {\mathcal{S}^\nu}')$, it follows that $\mathbb{E}[g(\mathcal{R}(Q^{\mathrm{Alg4}}))]\geq 0.38 \mathbb{E}[g(\mathcal{R}(Q^*_{\leq k}))]$. \endproof

{\section*{Extension 2: Incorporating the PNA Option as a Decision Variable}
Until now we assume that the decision on whether or not to offer PNA option is pre-given. As mentioned earlier, there exists two contradicting arguments regarding the impact of the PNA option on the performance of a quiz, it is not immediately clear whether including the PNA option in a question is beneficial or not. One natural extension is to treat the PNA option as a decision variable for each question. In this extended model, in addition to selecting and sequencing a group of questions, we must decide whether or not to offer the PNA option for each question so as to maximize the expected utility. To capture this situation, we extend the basic model by creating two versions $\{q^{w},  q^{w/o}\}$ of each question $q \in \Omega$, where $q^{w}$ denotes the version with PNA option and $q^{w/o}$ denotes the version that does not provide PNA option. Let $\bigcup_{q\in \Omega} \{q^{w},  q^{w/o}\}$ denote the expanded ground set. $q^{w}$ and $q^{w/o}$ have their own answer-through-rate and continuation probabilities.  Intuitively, selecting $q^{w}$ (resp. $q^{w/o}$) translates to displaying question $q$ with (resp. without including) the PNA option.

We follow a similar procedure as introduced in Section \ref{sec:4} to solve this problem except that we  are now dealing with the expanded set of questions. In particular, we first introduce a new problem $\textbf{P3.2}$ as follows.

 \begin{center}
\framebox[0.46\textwidth][c]{
\enspace
\begin{minipage}[t]{0.46\textwidth}
\small
$\textbf{P3.2}$
\emph{Maximize$_{a, \mathcal{S}}$ $v(a, \mathcal{S})$}\\
\textbf{subject to:}
\begin{equation*}
\begin{cases}
%$\forall \theta_{\mathcal{S}}> 0: |\mathcal{S}|\leq 1 $\\
-\sum_{a'\in \mathcal{S}} \log c_{a'} \leq -\log \rho\\
% \sum_{v\in V}\sum_{d\in D} y_{vd}\leq K \quad(C2)\\
% |\mathcal{S}|\leq b-1 \quad\mbox{(C2)}\\
|\mathcal{S}| \leq t \quad \\
\forall 1 \leq i \leq b: |\mathcal{S} \cap \{q^{w},  q^{w/o}\}|\leq 1 \mbox{ (C4.2)}\\
\mathcal{S} \subseteq \bigcup_{q\in \Omega} \{q^{w},  q^{w/o}\}\setminus \{a\}\\
%0 \leq t < b
\end{cases}
\end{equation*}
\end{minipage}
}
\end{center}
\vspace{0.1in}

As compared with $\textbf{P3}$, problem $\textbf{P3.2}$ imposes one additional constraint (C4.2). Condition (C4.2), which is a partition matroid constraint, ensures that only one version of a question can be selected in a feasible solution. Similar to the solution proposed in Section \ref{sec:4}, we enumerate all $a \in \bigcup_{q\in \Omega} \{q^{w},  q^{w/o}\}$, and solve $\textbf{P3.2}$ approximately to obtain a group of candidate solutions. Assume the best solution is $(a', \mathcal{S}')$, we build the final solution by first displaying $\mathcal{S}'$ in an arbitrary sequence and then displaying $a'$. Let $Q^{\mathrm{Alg5}}$ denote the sequence of questions returned from our algorithm.

Notice that $\textbf{P3.2}$ is a monotone submodular maximization problem subject to one partition matroid constraint (C4.2) and two linear constraints, we can obtain a $0.38$ approximation solution to this problem  \citep{vondrak2011submodular}. Then based on a similar proof of Theorem \ref{thm:extend1}, we have the following performance bound of our algorithm.
\begin{theorem}
\label{thm:extend2}For any $\rho\in[0,1]$, $f(Q^{\mathrm{Alg5}})\geq 0.38 \rho(1-\rho)f(Q^*)$.
\end{theorem}

\begin{corollary} By choosing $\rho=1/2$, we have $f(Q^{\mathrm{Alg5}})\geq \frac{0.38}{4}f(Q^*)$.
\end{corollary}

{\paragraph{Contribution to assortment optimization.} A similar extension applies to the assortment optimization problem by adding one more decision variable: \emph{product configuration}. Product configuration refers to the process of customizing the design of a product.  Most of existing studies in assortment optimization assume that the configuration of a product is fixed. In practise, the platform may face multiple ways of configuring a product before offering it to the customer. For the same product, different configurations may lead to different click-through-rate and continuation probabilities.  In this case, the platform needs to jointly optimize the product configuration and assortment selection to maximize the expected revenue. Our proposed framework is able to handle this joint optimization problem.  In particular, to incorporate the product configuration as decision variable, we  create multiple versions of the same product, then adopt a partition matroid constraint similar to  Condition (C4.2) to ensure the feasibility of the solution, e.g., the returned solution contains at most  one configuration of a product. It was worth noting that product pricing, as a special case of product configuration, has been well studied in the literature of assortment optimization \citep{wang2012capacitated,najafi2019multi}. Our proposed solution contributes to the literature by supporting a general form of product configuration.}

\section*{Extension 3: Extension to Scrolling Design}
 We next discuss how to extend this study to the scrolling design.  First of all, since scrolling design allows to display multiple questions on the same page, the customer may not strictly follow the sequence to answer questions, i.e., the customer may jump among questions. Therefore, the current markovian model is not well suited to the scrolling design. One possible way to handle this case is to introduce a \emph{position-dependent answer-through-rate} $p_{qi}$ for each slot $i$ and question $q$. $p_{qi}$ captures the probability that question $q$ is answered by the customer given $q$ is positioned in slot $i$. This value can be calculated on the basis of a simple formula:
\[p_{qi}=\frac{\mbox{the number of times $q$ is answered given it is displayed in slot $i$}}{\mbox{the number of times $q$ is displayed in slot $i$}}\]
As the above calculation does not involve the actual reading sequence of the customer, it is general enough to capture the scrolling design. However, as a tradeoff, this new model fails to capture the externality of different questions, i.e.,  the probability of answering a question does not depend on the events involving other questions. Therefore, when dealing with paging design, it is still preferable to use markovian model as it captures the fine-grained customer behavior, i.e., the externality of questions is well captured by the markovian model.

We first introduce the problem formulation under the scrolling design. The basic idea of our solution is similar to Algorithm \ref{alg:greedy-peak3}, i.e., convert the  joint  selection and sequencing problem to a selection problem. We recall some notations used in Section \ref{sec:extended} and introduce some new notations. For each question $q$, we create $b$ copies of virtual questions $\Omega^\nu_q=\{q^1, \cdots, q^b\}$. Let $\Omega^\nu= \bigcup_{q\in \Omega} \Omega^\nu_q$ denote the expanded ground set that is composed of virtual questions.  Define $\Omega^\nu_i=\{q^i| q\in \Omega\}$. %Given a set of virtual questions $\mathcal{S}^\nu \subseteq \Omega^\nu$, define $\mathcal{S}^\nu_{\min}=\{q^i| q^i\in \Omega^\nu_q \cap \mathcal{S}^\nu; i=\arg\max_j \{p_{q^j}| q^j\in \mathcal{S}^\nu\}\}$.
Let $\mathcal{S}=\{q|\Omega^\nu_q \cap \mathcal{S}^\nu\neq \emptyset\}$, we use $\mathcal{R}(\mathcal{S}^\nu)$ to denote a random set obtained by including each question $q\in \mathcal{S}$  with probability $p_{q^i}$ where  $p_{q^i}=\max \{p_{q^j}| q^j\in \mathcal{S}^\nu\}$. %Given a set of virtual questions $\mathcal{S}^\nu \subseteq \Omega^\nu$, let $\mathcal{S}\subseteq \Omega$ to denote its actual copies and define $g(\mathcal{S}^\nu)=g(\mathcal{S})$.
% Recall that selecting a virtual question $q^i$ translates to placing $q$ at slot $i$, thus
The expected utility $v(\mathcal{S}^\nu)$ of selecting $\mathcal{S}^\nu$ is
\[v(\mathcal{S}^\nu)= \mathbb{E}[g(\mathcal{R} (\mathcal{S}^\nu))]\] %where $\mathcal{R} (\mathcal{S}^\nu)$ is (redefined as) a random set obtained by including each virtual question $q^i\in \mathcal{S}^\nu$ with probability $p_{qi}$.

 The objective of our problem $\textbf{P4}$ is to select a group of virtual questions that maximizes the expected utility.

 \begin{center}
\framebox[0.46\textwidth][c]{
\enspace
\begin{minipage}[t]{0.46\textwidth}
\small
$\textbf{P4}$
\emph{Maximize$_{\mathcal{S}^\nu \subseteq \Omega^\nu}$ $v(\mathcal{S}^\nu)$}\\
\textbf{subject to:} \begin{equation*}
\forall 1 \leq i \leq b: |\mathcal{S}^\nu \cap \Omega^\nu_i|\leq 1 \mbox{ (C5.1)}\\
%\begin{cases}
%%$\forall \theta_{\mathcal{S}}> 0: |\mathcal{S}|\leq 1 $\\
%% \sum_{v\in V}\sum_{d\in D} y_{vd}\leq K \quad(C2)\\
%% |\mathcal{S}|\leq b-1 \quad\mbox{(C2)}\\
%%|\mathcal{S}^\nu| \leq t \quad \\
%\forall 1 \leq i \leq b: |\mathcal{S}^\nu \cap \Omega^\nu_i|\leq 1 \mbox{ (C5.1)}\\
%\forall q \in \Omega: |\mathcal{S}^\nu \cap \Omega^\nu_q|\leq 1 \mbox{ (C5.2)}\\
%%0 \leq t < b
%\end{cases}
\end{equation*}
\end{minipage}
}
\end{center}
\vspace{0.1in}

In the above formulation, constraint (C5.1) ensures that we assign at most one question to each slot. Because $g(\mathcal{R} (\mathcal{S}^\nu))$ is monotone and submodular, we have the following lemma. %, and constraint (C5.2) ensures that the same question will not appear multiple times.
\begin{lemma}
\label{lem:123445}
$v(\mathcal{S}^\nu)$ is monotone and submodular.
\end{lemma}
%\begin{proof}The proof of this lemma is straightforward. First of all, it is easy to verify that $g(\mathcal{S}^\nu)$ is monotone and submodular. According to the definition of $v(\mathcal{S}^\nu)$, we have $v(\mathcal{S}^\nu)=\mathbb{E}[g(\mathcal{R} (\mathcal{S}^\nu))]=\sum_{\mathcal{U}^\nu\subseteq \Omega^\nu}\Pr[\mathcal{R} (\mathcal{S}^\nu)=\mathcal{U}^\nu]g(\mathcal{U}^\nu)$. Thus, $v(\mathcal{S}^\nu)$, as a linear combination of $g(\mathcal{S}^\nu)$, is also monotone and submodular. \end{proof}

Due to Lemma \ref{lem:123445}, $\textbf{P4}$ is a monotone submodular maximization problem subject to a matroid constraint. According to \citep{nemhauser1978analysis}, a simple greedy algorithm (Algorithm \ref{alg:greedy-peak5}), which starts with an empty set, and at every step assigns a question to some slot which maximizes the marginal benefit subject to the matroid constraint, provides a sequence that achieves a $1/2$-approximation of the optimum. Notice that a feasible solution to $\textbf{P4}$ may include multiple copies from the same question, this  redundancy issue can be easily resolved by keeping the one which has the largest answer-through-rate in the solution. This will not affect the utility of our solution due to the definition of $v(\mathcal{S}^\nu)$.

\begin{algorithm}[h]
{\small
\caption{Question Selection and Sequencing under Scrolling Design}
\label{alg:greedy-peak5}
\textbf{Input:} $\Omega$.\\
\textbf{Output:} $Q^{\mathrm{Alg6}}$.
\begin{algorithmic}[1]
\STATE Set ${\mathcal{S}^\nu}'=\emptyset$.
\FOR{$t\in [1, b]$}
%\FOR{$q\in \Omega$}
\STATE add to ${\mathcal{S}^\nu}'$ an virtual question $q^i$ that maximizes $v(\mathcal{S}^\nu\cup \{q^i\})-v(\mathcal{S}^\nu)$  subject to constraints (C5.1)\label{line:757}
%\IF {$u(t, q, {\mathcal{S}^\nu})> u(t', {q}', {\mathcal{S}^\nu}')$}
%\STATE $t'\leftarrow t, {\mathcal{S}^\nu}' \leftarrow {\mathcal{S}^\nu}, {q}'\leftarrow q$
%\ENDIF
%\ENDFOR
\ENDFOR
\FOR{$q^i \in {\mathcal{S}^\nu}'$}
\STATE place $q$ at slot $i$ of $Q^{\mathrm{Alg6}}$
\ENDFOR
%\STATE Place $q'$ at slot $t'$ of $Q^{\mathrm{Alg4}}$
%\STATE $Q^{\mathrm{Alg3}}\leftarrow S' \oplus \{{q^t}'\}$ where $ S'$ is the sequence of  actual questions decided by $\mathcal{S}'$ \label{line:1}
\RETURN $Q^{\mathrm{Alg6}}$ \COMMENT{we may need to refine $Q^{\mathrm{Alg6}}$ by removing any redundant questions.}  %\COMMENT{we may need to refine $Q^{\mathrm{Alg4}}$ by removing any redundant questions and gaps.}
\end{algorithmic}
}
\end{algorithm}

\begin{theorem}
A simple greedy algorithm (Algorithm \ref{alg:greedy-peak5}) achieves a $1/2$ approximation ratio.
\end{theorem}}

{\paragraph{Contribution to assortment optimization.}  In the context of assortment optimization, \citeauthor{aouad2019click} considers a click-based model that is similar to our customer browse model presented in this section. They assume that each product is associated with a consideration probability. In the stage of forming a consideration set, each product is added to the consideration set independently with its consideration probability. Our study is different from theirs in two ways: Our customer browse model is more general than theirs in that under our model, the probability of a product being considered is not only dependent on its index but also dependent on its position. Moreover, they choose the multinomial logit model as their choice model, where the revenue function is not always monotone and submodular. \cite{han2019assortment} identify  the conditions under which the revenue function $r$  is submodular under the  multinomial logit model.  Our results are not restricted to any particular underlying choice model. Rather, for any assortment optimization problem that involves position-dependent consider-through-rate, and monotone and submodular revenue function,  our greedy algorithm (Algorithm 5) achieves a $1/2$ approximation ratio.}

\section*{Missing tables from the experiment}

{\begin{table*}[ht]\centering
\caption{Performance of {\it QSS} on a Wider Range of Parameters: $p^-_q$ and $c^-_q$}
\begin{tabular}{cccccccccc}
\hline
 & \multicolumn{9}{c}{$c_{q}^-$}\\
\cline{2-10} $p_{q}^-$ & $0.1$ & $0.2$ & $0.3$ & $0.4$ & $0.5$ & $0.6$ & $0.7$ & $0.8$ & $0.9$\\
\hline
\multirow{2}{*}{$0.1$} & $0.8725$ & $0.8724$ & $0.8721$ & $0.8717$ & $0.8712$ & $0.8706$ & $0.8702$ & $0.8696$ & $0.8691$\\
 & $0.9978$ & $0.9978$ & $0.9977$ & $0.9977$ & $0.9976$ & $0.9975$ & $0.9975$ & $0.9974$ & $0.9972$\\
 \hline
\multirow{2}{*}{$0.2$} & $0.8992$ & $0.8990$ & $0.8987$ & $0.8983$ & $0.8979$ & $0.8975$ & $0.8971$ & $0.8965$ & $0.8960$\\
 & $0.9980$ & $0.9980$ & $0.9980$ & $0.9979$ & $0.9979$ & $0.9978$ & $0.9978$ & $0.9977$ & $0.9977$ \\
\hline
\multirow{2}{*}{$0.3$} & $0.9159$ & $0.9158$ & $0.9156$ & $0.9152$ & $0.9149$ & $0.9144$ & $0.9139$ & $0.9135$ & $0.9128$\\
 & $0.9983$ & $0.9983$ & $0.9982$ & $0.9982$ & $0.9982$ & $0.9981$ & $0.9981$ & $0.9981$ & $0.9980$\\
\hline
\multirow{2}{*}{$0.4$} & $0.9304$ & $0.9303$ & $0.9301$ & $0.9297$ & $0.9292$ & $0.9286$ & $0.9281$ & $0.9276$ & $0.9271$\\
 & $0.9985$ & $0.9985$ & $0.9985$ & $0.9984$ & $0.9983$ & $0.9983$ & $0.9983$ & $0.9982$ & $0.9982$\\
\hline
\multirow{2}{*}{$0.5$} & $0.9451$ & $0.9449$ & $0.9445$ & $0.9440$ & $0.9436$ & $0.9431$ & $0.9427$ & $0.9424$ & $0.9418$\\
 & $0.9986$ & $0.9986$ & $0.9986$ & $0.9985$ & $0.9985$ & $0.9985$ & $0.9985$ & $0.9985$ & $0.9984$\\
\hline
\multirow{2}{*}{$0.6$} & $0.9594$ & $0.9592$ & $0.9589$ & $0.9585$ & $0.9581$ & $0.9576$ & $0.9572$ & $0.9568$ & $0.9562$\\
 & $0.9987$ & $0.9987$ & $0.9987$ & $0.9986$ & $0.9986$ & $0.9986$ & $0.9986$ & $0.9986$ & $0.9985$\\
\hline
\multirow{2}{*}{$0.7$} & $0.9757$ & $0.9754$ & $0.9752$ & $0.9748$ & $0.9745$ & $0.9740$ & $0.9736$ & $0.9733$ & $0.9727$\\
 & $0.9989$ & $0.9989$ & $0.9988$ & $0.9988$ & $0.9988$ & $0.9988$ & $0.9987$ & $0.9987$ & $0.9987$\\
\hline
\multirow{2}{*}{$0.8$} & $0.9837$ & $0.9835$ & $0.9831$ & $0.9829$ & $0.9824$ & $0.9821$ & $0.9817$ & $0.9812$ & $0.9808$\\
 & $0.9991$ & $0.9991$ & $0.9991$ & $0.9991$ & $0.9991$ & $0.9991$ & $0.9990$ & $0.9990$ & $0.9990$\\
 \hline
\multirow{2}{*}{$0.9$} & $0.9926$ & $0.9925$ & $0.9921$ & $0.9918$ & $0.9915$ & $0.9912$ & $0.9908$ & $0.9903$ & 0.$9899$\\
 & $0.9993$ & $0.9993$ & $0.9992$ & $0.9992$ & $0.9992$ & $0.9992$ & $0.9992$ & $0.9991$ & $0.9991$\\
\hline
\end{tabular}
\label{tab:approx_ratio_2}
\end{table*}}

\begin{table*}[ht]\centering
\caption{Impact of the PNA Option on the Optimal Expected Utility}
\begin{tabular}{|c|c|c|c|c|c|c|c|c|}
\hline
$p_{q}^+$ & $c_{q}^+$ & $p_{q}^-$ & $c_{q}^-$ & $\kappa$ & \tabincell{c}{Expected Utility\\(with PNA)} & \tabincell{c}{Expected Utility\\(without PNA)} & \tabincell{c}{Utility\\Reduction} & \tabincell{c}{Reduction\\Percentage} \\
\hline
\hline
$0.3$ & $0.3$ & $0.1$ & $0.1$ & $-0.9$ & $2.2652$ & $2.0019$ & $0.2633$ & $11.624\%$ \\
\hline
$0.3$ & $0.3$ & $0.1$ & $0.1$ & $-0.7$ & $2.2652$ & $2.0201$ & $0.2451$ & $10.820\%$ \\
\hline
$0.3$ & $0.3$ & $0.1$ & $0.1$ & $-0.5$ & $2.2652$ & $2.0408$ & $0.2246$ & $9.915\%$ \\
\hline
$0.3$ & $0.3$ & $0.1$ & $0.1$ & $-0.3$ & $2.2652$ & $2.1003$ & $0.1649$ & $7.280\%$ \\
\hline
$0.3$ & $0.3$ & $0.1$ & $0.1$ & $-0.1$ & $2.2652$ & $2.1617$ & $0.1035$ & $4.569\%$ \\
\hline
$0.3$ & $0.3$ & $0.1$ & $0.1$ & $0$ & $2.2652$ & $2.2252$ & $0.0400$ & $1.766\%$ \\
\hline
$0.3$ & $0.3$ & $0.1$ & $0.1$ & $0.1$ & $2.2652$ & $2.2633$ & $0.0019$ & $0.085\%$ \\
\hline
$0.3$ & $0.3$ & $0.1$ & $0.1$ & $0.3$ & $2.2652$ & $2.4568$ & $-0.1916$ & $-8.458\%$ \\
\hline
$0.3$ & $0.3$ & $0.1$ & $0.1$ & $0.5$ & $2.2652$ & $2.5242$ & $-0.2590$ & $-11.434\%$ \\
\hline
$0.3$ & $0.3$ & $0.1$ & $0.1$ & $0.7$ & $2.2652$ & $2.7174$ & $-0.4522$ & $-19.963\%$ \\
\hline
$0.3$ & $0.3$ & $0.1$ & $0.1$ & $0.9$ & $2.2652$ & $2.8621$ & $-0.5969$ & $-26.351\%$ \\
\hline
\end{tabular}
\label{tab:pna_no_pna_1}
\end{table*}

\begin{table*}[ht]\centering
\caption{Impact of the PNA Option on the Optimal Expected Utility}
\begin{tabular}{|c|c|c|c|c|c|c|c|c|}
\hline
$p_{q}^+$ & $c_{q}^+$ & $p_{q}^-$ & $c_{q}^-$ & $\kappa$ & \tabincell{c}{Expected Utility\\(with PNA)} & \tabincell{c}{Expected Utility\\(without PNA)} & \tabincell{c}{Utility\\Reduction} & \tabincell{c}{Reduction\\Percentage} \\
\hline
\hline
$0.35$ & $0.35$ & $0.3$ & $0.3$ & $-0.9$ & $2.4654$ & $2.1454$ & $0.3200$ & $12.980\%$ \\
\hline
$0.35$ & $0.35$ & $0.3$ & $0.3$ & $-0.7$ & $2.4654$ & $2.2216$ & $0.2438$ & $9.889\%$ \\
\hline
$0.35$ & $0.35$ & $0.3$ & $0.3$ & $-0.5$ & $2.4654$ & $2.2466$ & $0.2188$ & $8.875\%$ \\
\hline
$0.35$ & $0.35$ & $0.3$ & $0.3$ & $-0.3$ & $2.4654$ & $2.3091$ & $0.1563$ & $6.340\%$ \\
\hline
$0.35$ & $0.35$ & $0.3$ & $0.3$ & $-0.1$ & $2.4654$ & $2.3575$ & $0.1079$ & $4.377\%$ \\
\hline
$0.35$ & $0.35$ & $0.3$ & $0.3$ & $0$ & $2.4654$ & $2.4035$ & $0.0619$ & $2.511\%$ \\
\hline
$0.35$ & $0.35$ & $0.3$ & $0.3$ & $0.1$ & $2.4654$ & $2.4412$ & $0.0242$ & $0.982\%$ \\
\hline
$0.35$ & $0.35$ & $0.3$ & $0.3$ & $0.3$ & $2.4654$ & $2.5075$ & $-0.0421$ & $-1.708\%$ \\
\hline
$0.35$ & $0.35$ & $0.3$ & $0.3$ & $0.5$ & $2.4654$ & $2.6858$ & $-0.2204$ & $-8.940\%$ \\
\hline
$0.35$ & $0.35$ & $0.3$ & $0.3$ & $0.7$ & $2.4654$ & $2.8889$ & $-0.4235$ & $-17.178\%$ \\
\hline
$0.35$ & $0.35$ & $0.3$ & $0.3$ & $0.9$ & $2.4654$ & $2.9878$ & $-0.5224$ & $-21.189\%$ \\
\hline
\end{tabular}
\label{tab:pna_no_pna_2}
\end{table*}

\begin{table*}[ht]\centering
\caption{Impact of the PNA Option on the Optimal Expected Utility}
\begin{tabular}{|c|c|c|c|c|c|c|c|c|}
\hline
$p_{q}^+$ & $c_{q}^+$ & $p_{q}^-$ & $c_{q}^-$ & $\kappa$ & \tabincell{c}{Expected Utility\\(with PNA)} & \tabincell{c}{Expected Utility\\(without PNA)} & \tabincell{c}{Utility\\Reduction} & \tabincell{c}{Reduction\\Percentage} \\
\hline
\hline
$0.4$ & $0.4$ & $0.5$ & $0.5$ & $-0.9$ & $2.5608$ & $2.2490$ & $0.3118$ & $12.176\%$ \\
\hline
$0.4$ & $0.4$ & $0.5$ & $0.5$ & $-0.7$ & $2.5608$ & $2.3022$ & $0.2586$ & $10.098\%$ \\
\hline
$0.4$ & $0.4$ & $0.5$ & $0.5$ & $-0.5$ & $2.5608$ & $2.3554$ & $0.2054$ & $8.021\%$ \\
\hline
$0.4$ & $0.4$ & $0.5$ & $0.5$ & $-0.3$ & $2.5608$ & $2.3987$ & $0.1621$ & $6.330\%$ \\
\hline
$0.4$ & $0.4$ & $0.5$ & $0.5$ & $-0.1$ & $2.5608$ & $2.4347$ & $0.1261$ & $4.924\%$ \\
\hline
$0.4$ & $0.4$ & $0.5$ & $0.5$ & $0$ & $2.5608$ & $2.4670$ & $0.0938$ & $3.663\%$ \\
\hline
$0.4$ & $0.4$ & $0.5$ & $0.5$ & $0.1$ & $2.5608$ & $2.4959$ & $0.0649$ & $2.534\%$ \\
\hline
$0.4$ & $0.4$ & $0.5$ & $0.5$ & $0.3$ & $2.5608$ & $2.6742$ & $-0.1134$ & $-4.428\%$ \\
\hline
$0.4$ & $0.4$ & $0.5$ & $0.5$ & $0.5$ & $2.5608$ & $2.9149$ & $-0.3541$ & $-13.828\%$ \\
\hline
$0.4$ & $0.4$ & $0.5$ & $0.5$ & $0.7$ & $2.5608$ & $3.0763$ & $-0.5155$ & $-20.130\%$ \\
\hline
$0.4$ & $0.4$ & $0.5$ & $0.5$ & $0.9$ & $2.5608$ & $3.1450$ & $-0.5842$ & $-22.813\%$ \\
\hline
\end{tabular}
\label{tab:pna_no_pna_3}
\end{table*}

\begin{table*}[ht]\centering
\caption{Impact of the PNA Option on the Optimal Expected Utility}
\begin{tabular}{|c|c|c|c|c|c|c|c|c|}
\hline
$p_{q}^+$ & $c_{q}^+$ & $p_{q}^-$ & $c_{q}^-$ & $\kappa$ & \tabincell{c}{Expected Utility\\(with PNA)} & \tabincell{c}{Expected Utility\\(without PNA)} & \tabincell{c}{Utility\\Reduction} & \tabincell{c}{Reduction\\Percentage} \\
\hline
\hline
$0.5$ & $0.5$ & $0.3$ & $0.3$ & $-0.9$ & $2.5976$ & $2.2739$ & $0.3237$ & $12.462\%$ \\
\hline
$0.5$ & $0.5$ & $0.3$ & $0.3$ & $-0.7$ & $2.5976$ & $2.3524$ & $0.2452$ & $9.439\%$ \\
\hline
$0.5$ & $0.5$ & $0.3$ & $0.3$ & $-0.5$ & $2.5976$ & $2.4174$ & $0.1802$ & $6.937\%$ \\
\hline
$0.5$ & $0.5$ & $0.3$ & $0.3$ & $-0.3$ & $2.5976$ & $2.4653$ & $0.1323$ & $5.093\%$ \\
\hline
$0.5$ & $0.5$ & $0.3$ & $0.3$ & $-0.1$ & $2.5976$ & $2.4992$ & $0.0984$ & $3.788\%$ \\
\hline
$0.5$ & $0.5$ & $0.3$ & $0.3$ & $0$ & $2.5976$ & $2.5287$ & $0.0689$ & $2.652\%$ \\
\hline
$0.5$ & $0.5$ & $0.3$ & $0.3$ & $0.1$ & $2.5976$ & $2.5658$ & $0.0318$ & $1.224\%$ \\
\hline
$0.5$ & $0.5$ & $0.3$ & $0.3$ & $0.3$ & $2.5976$ & $2.7245$ & $-0.1269$ & $-4.885\%$ \\
\hline
$0.5$ & $0.5$ & $0.3$ & $0.3$ & $0.5$ & $2.5976$ & $2.9624$ & $-0.3648$ & $-14.044\%$ \\
\hline
$0.5$ & $0.5$ & $0.3$ & $0.3$ & $0.7$ & $2.5976$ & $3.1636$ & $-0.5660$ & $-21.789\%$ \\
\hline
$0.5$ & $0.5$ & $0.3$ & $0.3$ & $0.9$ & $2.5976$ & $3.2438$ & $-0.6462$ & $-24.877\%$ \\
\hline
\end{tabular}
\label{tab:pna_no_pna_4}
\end{table*}

\end{document}